\documentclass[11pt,letterpaper]{article}
\usepackage[margin=1in]{geometry}
\usepackage[T1]{fontenc}
\usepackage[utf8]{inputenc}
\usepackage{lmodern}
\usepackage{microtype}
\usepackage{amsmath,amssymb}
\usepackage{graphicx,booktabs,tabularx,longtable,array}
\usepackage{caption}
\usepackage{placeins}
\usepackage[numbers,sort&compress]{natbib}
\usepackage{xcolor}
\usepackage{hyperref}
\hypersetup{colorlinks=true,linkcolor=blue!45!black,citecolor=blue!45!black,urlcolor=blue!45!black,
 pdftitle={When Does Cross-View Consistency Matter in Multi-View Appearance Augmentation?}}
\newcommand{\SR}{\operatorname{SR}}
\newcommand{\eff}[1]{\ensuremath{\res{#1}{value2}\,[\res{#1}{low2},\,\res{#1}{high2}]}}
\newcommand{\effone}[1]{\ensuremath{\res{#1}{value1}\,[\res{#1}{low1},\,\res{#1}{high1}]}}
\newcommand{\res}[2]{\csname result@#1@#2\endcsname}
\expandafter\def\csname result@camera_mask/E1_brms@value1\endcsname{33.3}
\expandafter\def\csname result@camera_mask/E1_brms@low1\endcsname{23.8}
\expandafter\def\csname result@camera_mask/E1_brms@high1\endcsname{41.5}
\expandafter\def\csname result@camera_mask/E1_brms@low951\endcsname{22.0}
\expandafter\def\csname result@camera_mask/E1_brms@high951\endcsname{42.8}
\expandafter\def\csname result@camera_mask/E1_brms@value2\endcsname{33.26}
\expandafter\def\csname result@camera_mask/E1_brms@low2\endcsname{23.83}
\expandafter\def\csname result@camera_mask/E1_brms@high2\endcsname{41.52}
\expandafter\def\csname result@camera_mask/E1_brms@low952\endcsname{22.02}
\expandafter\def\csname result@camera_mask/E1_brms@high952\endcsname{42.84}
\expandafter\def\csname result@camera_mask/dp_C0_agentview_brms@value1\endcsname{26.7}
\expandafter\def\csname result@camera_mask/dp_C0_agentview_brms@low1\endcsname{19.2}
\expandafter\def\csname result@camera_mask/dp_C0_agentview_brms@high1\endcsname{33.6}
\expandafter\def\csname result@camera_mask/dp_C0_agentview_brms@low951\endcsname{17.8}
\expandafter\def\csname result@camera_mask/dp_C0_agentview_brms@high951\endcsname{34.9}
\expandafter\def\csname result@camera_mask/dp_C0_agentview_brms@value2\endcsname{26.70}
\expandafter\def\csname result@camera_mask/dp_C0_agentview_brms@low2\endcsname{19.24}
\expandafter\def\csname result@camera_mask/dp_C0_agentview_brms@high2\endcsname{33.61}
\expandafter\def\csname result@camera_mask/dp_C0_agentview_brms@low952\endcsname{17.77}
\expandafter\def\csname result@camera_mask/dp_C0_agentview_brms@high952\endcsname{34.89}
\expandafter\def\csname result@camera_mask/dp_C0_wrist_brms@value1\endcsname{54.9}
\expandafter\def\csname result@camera_mask/dp_C0_wrist_brms@low1\endcsname{43.2}
\expandafter\def\csname result@camera_mask/dp_C0_wrist_brms@high1\endcsname{63.4}
\expandafter\def\csname result@camera_mask/dp_C0_wrist_brms@low951\endcsname{40.3}
\expandafter\def\csname result@camera_mask/dp_C0_wrist_brms@high951\endcsname{64.5}
\expandafter\def\csname result@camera_mask/dp_C0_wrist_brms@value2\endcsname{54.86}
\expandafter\def\csname result@camera_mask/dp_C0_wrist_brms@low2\endcsname{43.24}
\expandafter\def\csname result@camera_mask/dp_C0_wrist_brms@high2\endcsname{63.39}
\expandafter\def\csname result@camera_mask/dp_C0_wrist_brms@low952\endcsname{40.28}
\expandafter\def\csname result@camera_mask/dp_C0_wrist_brms@high952\endcsname{64.47}
\expandafter\def\csname result@camera_mask/dp_C10_agentview_brms@value1\endcsname{19.4}
\expandafter\def\csname result@camera_mask/dp_C10_agentview_brms@low1\endcsname{14.0}
\expandafter\def\csname result@camera_mask/dp_C10_agentview_brms@high1\endcsname{24.7}
\expandafter\def\csname result@camera_mask/dp_C10_agentview_brms@low951\endcsname{12.9}
\expandafter\def\csname result@camera_mask/dp_C10_agentview_brms@high951\endcsname{25.7}
\expandafter\def\csname result@camera_mask/dp_C10_agentview_brms@value2\endcsname{19.40}
\expandafter\def\csname result@camera_mask/dp_C10_agentview_brms@low2\endcsname{13.97}
\expandafter\def\csname result@camera_mask/dp_C10_agentview_brms@high2\endcsname{24.72}
\expandafter\def\csname result@camera_mask/dp_C10_agentview_brms@low952\endcsname{12.93}
\expandafter\def\csname result@camera_mask/dp_C10_agentview_brms@high952\endcsname{25.68}
\expandafter\def\csname result@camera_mask/dp_C10_wrist_brms@value1\endcsname{57.5}
\expandafter\def\csname result@camera_mask/dp_C10_wrist_brms@low1\endcsname{44.7}
\expandafter\def\csname result@camera_mask/dp_C10_wrist_brms@high1\endcsname{66.9}
\expandafter\def\csname result@camera_mask/dp_C10_wrist_brms@low951\endcsname{41.3}
\expandafter\def\csname result@camera_mask/dp_C10_wrist_brms@high951\endcsname{68.1}
\expandafter\def\csname result@camera_mask/dp_C10_wrist_brms@value2\endcsname{57.51}
\expandafter\def\csname result@camera_mask/dp_C10_wrist_brms@low2\endcsname{44.72}
\expandafter\def\csname result@camera_mask/dp_C10_wrist_brms@high2\endcsname{66.89}
\expandafter\def\csname result@camera_mask/dp_C10_wrist_brms@low952\endcsname{41.32}
\expandafter\def\csname result@camera_mask/dp_C10_wrist_brms@high952\endcsname{68.07}
\expandafter\def\csname result@camera_mask/dp_C11_agentview_brms@value1\endcsname{32.8}
\expandafter\def\csname result@camera_mask/dp_C11_agentview_brms@low1\endcsname{24.9}
\expandafter\def\csname result@camera_mask/dp_C11_agentview_brms@high1\endcsname{39.7}
\expandafter\def\csname result@camera_mask/dp_C11_agentview_brms@low951\endcsname{23.2}
\expandafter\def\csname result@camera_mask/dp_C11_agentview_brms@high951\endcsname{40.9}
\expandafter\def\csname result@camera_mask/dp_C11_agentview_brms@value2\endcsname{32.83}
\expandafter\def\csname result@camera_mask/dp_C11_agentview_brms@low2\endcsname{24.91}
\expandafter\def\csname result@camera_mask/dp_C11_agentview_brms@high2\endcsname{39.73}
\expandafter\def\csname result@camera_mask/dp_C11_agentview_brms@low952\endcsname{23.17}
\expandafter\def\csname result@camera_mask/dp_C11_agentview_brms@high952\endcsname{40.87}
\expandafter\def\csname result@camera_mask/dp_C11_wrist_brms@value1\endcsname{42.9}
\expandafter\def\csname result@camera_mask/dp_C11_wrist_brms@low1\endcsname{32.3}
\expandafter\def\csname result@camera_mask/dp_C11_wrist_brms@high1\endcsname{51.5}
\expandafter\def\csname result@camera_mask/dp_C11_wrist_brms@low951\endcsname{29.6}
\expandafter\def\csname result@camera_mask/dp_C11_wrist_brms@high951\endcsname{52.8}
\expandafter\def\csname result@camera_mask/dp_C11_wrist_brms@value2\endcsname{42.92}
\expandafter\def\csname result@camera_mask/dp_C11_wrist_brms@low2\endcsname{32.32}
\expandafter\def\csname result@camera_mask/dp_C11_wrist_brms@high2\endcsname{51.54}
\expandafter\def\csname result@camera_mask/dp_C11_wrist_brms@low952\endcsname{29.65}
\expandafter\def\csname result@camera_mask/dp_C11_wrist_brms@high952\endcsname{52.77}
\expandafter\def\csname result@camera_mask/dp_C12_agentview_brms@value1\endcsname{30.8}
\expandafter\def\csname result@camera_mask/dp_C12_agentview_brms@low1\endcsname{23.5}
\expandafter\def\csname result@camera_mask/dp_C12_agentview_brms@high1\endcsname{37.1}
\expandafter\def\csname result@camera_mask/dp_C12_agentview_brms@low951\endcsname{21.8}
\expandafter\def\csname result@camera_mask/dp_C12_agentview_brms@high951\endcsname{38.2}
\expandafter\def\csname result@camera_mask/dp_C12_agentview_brms@value2\endcsname{30.75}
\expandafter\def\csname result@camera_mask/dp_C12_agentview_brms@low2\endcsname{23.52}
\expandafter\def\csname result@camera_mask/dp_C12_agentview_brms@high2\endcsname{37.09}
\expandafter\def\csname result@camera_mask/dp_C12_agentview_brms@low952\endcsname{21.82}
\expandafter\def\csname result@camera_mask/dp_C12_agentview_brms@high952\endcsname{38.23}
\expandafter\def\csname result@camera_mask/dp_C12_wrist_brms@value1\endcsname{45.7}
\expandafter\def\csname result@camera_mask/dp_C12_wrist_brms@low1\endcsname{34.0}
\expandafter\def\csname result@camera_mask/dp_C12_wrist_brms@high1\endcsname{55.1}
\expandafter\def\csname result@camera_mask/dp_C12_wrist_brms@low951\endcsname{31.4}
\expandafter\def\csname result@camera_mask/dp_C12_wrist_brms@high951\endcsname{56.4}
\expandafter\def\csname result@camera_mask/dp_C12_wrist_brms@value2\endcsname{45.66}
\expandafter\def\csname result@camera_mask/dp_C12_wrist_brms@low2\endcsname{33.98}
\expandafter\def\csname result@camera_mask/dp_C12_wrist_brms@high2\endcsname{55.10}
\expandafter\def\csname result@camera_mask/dp_C12_wrist_brms@low952\endcsname{31.37}
\expandafter\def\csname result@camera_mask/dp_C12_wrist_brms@high952\endcsname{56.43}
\expandafter\def\csname result@camera_mask/dp_C1_agentview_brms@value1\endcsname{15.3}
\expandafter\def\csname result@camera_mask/dp_C1_agentview_brms@low1\endcsname{10.7}
\expandafter\def\csname result@camera_mask/dp_C1_agentview_brms@high1\endcsname{19.9}
\expandafter\def\csname result@camera_mask/dp_C1_agentview_brms@low951\endcsname{9.9}
\expandafter\def\csname result@camera_mask/dp_C1_agentview_brms@high951\endcsname{20.9}
\expandafter\def\csname result@camera_mask/dp_C1_agentview_brms@value2\endcsname{15.29}
\expandafter\def\csname result@camera_mask/dp_C1_agentview_brms@low2\endcsname{10.71}
\expandafter\def\csname result@camera_mask/dp_C1_agentview_brms@high2\endcsname{19.94}
\expandafter\def\csname result@camera_mask/dp_C1_agentview_brms@low952\endcsname{9.85}
\expandafter\def\csname result@camera_mask/dp_C1_agentview_brms@high952\endcsname{20.92}
\expandafter\def\csname result@camera_mask/dp_C1_wrist_brms@value1\endcsname{56.2}
\expandafter\def\csname result@camera_mask/dp_C1_wrist_brms@low1\endcsname{43.7}
\expandafter\def\csname result@camera_mask/dp_C1_wrist_brms@high1\endcsname{65.7}
\expandafter\def\csname result@camera_mask/dp_C1_wrist_brms@low951\endcsname{40.1}
\expandafter\def\csname result@camera_mask/dp_C1_wrist_brms@high951\endcsname{67.0}
\expandafter\def\csname result@camera_mask/dp_C1_wrist_brms@value2\endcsname{56.23}
\expandafter\def\csname result@camera_mask/dp_C1_wrist_brms@low2\endcsname{43.67}
\expandafter\def\csname result@camera_mask/dp_C1_wrist_brms@high2\endcsname{65.72}
\expandafter\def\csname result@camera_mask/dp_C1_wrist_brms@low952\endcsname{40.07}
\expandafter\def\csname result@camera_mask/dp_C1_wrist_brms@high952\endcsname{67.03}
\expandafter\def\csname result@camera_mask/dp_C5_agentview_brms@value1\endcsname{16.5}
\expandafter\def\csname result@camera_mask/dp_C5_agentview_brms@low1\endcsname{11.2}
\expandafter\def\csname result@camera_mask/dp_C5_agentview_brms@high1\endcsname{21.7}
\expandafter\def\csname result@camera_mask/dp_C5_agentview_brms@low951\endcsname{10.1}
\expandafter\def\csname result@camera_mask/dp_C5_agentview_brms@high951\endcsname{22.7}
\expandafter\def\csname result@camera_mask/dp_C5_agentview_brms@value2\endcsname{16.53}
\expandafter\def\csname result@camera_mask/dp_C5_agentview_brms@low2\endcsname{11.15}
\expandafter\def\csname result@camera_mask/dp_C5_agentview_brms@high2\endcsname{21.73}
\expandafter\def\csname result@camera_mask/dp_C5_agentview_brms@low952\endcsname{10.13}
\expandafter\def\csname result@camera_mask/dp_C5_agentview_brms@high952\endcsname{22.75}
\expandafter\def\csname result@camera_mask/dp_C5_wrist_brms@value1\endcsname{56.1}
\expandafter\def\csname result@camera_mask/dp_C5_wrist_brms@low1\endcsname{43.4}
\expandafter\def\csname result@camera_mask/dp_C5_wrist_brms@high1\endcsname{65.7}
\expandafter\def\csname result@camera_mask/dp_C5_wrist_brms@low951\endcsname{40.5}
\expandafter\def\csname result@camera_mask/dp_C5_wrist_brms@high951\endcsname{66.9}
\expandafter\def\csname result@camera_mask/dp_C5_wrist_brms@value2\endcsname{56.10}
\expandafter\def\csname result@camera_mask/dp_C5_wrist_brms@low2\endcsname{43.43}
\expandafter\def\csname result@camera_mask/dp_C5_wrist_brms@high2\endcsname{65.68}
\expandafter\def\csname result@camera_mask/dp_C5_wrist_brms@low952\endcsname{40.48}
\expandafter\def\csname result@camera_mask/dp_C5_wrist_brms@high952\endcsname{66.88}
\expandafter\def\csname result@camera_mask/dp_C9_agentview_brms@value1\endcsname{21.1}
\expandafter\def\csname result@camera_mask/dp_C9_agentview_brms@low1\endcsname{14.8}
\expandafter\def\csname result@camera_mask/dp_C9_agentview_brms@high1\endcsname{27.0}
\expandafter\def\csname result@camera_mask/dp_C9_agentview_brms@low951\endcsname{13.5}
\expandafter\def\csname result@camera_mask/dp_C9_agentview_brms@high951\endcsname{28.0}
\expandafter\def\csname result@camera_mask/dp_C9_agentview_brms@value2\endcsname{21.07}
\expandafter\def\csname result@camera_mask/dp_C9_agentview_brms@low2\endcsname{14.83}
\expandafter\def\csname result@camera_mask/dp_C9_agentview_brms@high2\endcsname{26.99}
\expandafter\def\csname result@camera_mask/dp_C9_agentview_brms@low952\endcsname{13.52}
\expandafter\def\csname result@camera_mask/dp_C9_agentview_brms@high952\endcsname{28.02}
\expandafter\def\csname result@camera_mask/dp_C9_wrist_brms@value1\endcsname{59.5}
\expandafter\def\csname result@camera_mask/dp_C9_wrist_brms@low1\endcsname{45.4}
\expandafter\def\csname result@camera_mask/dp_C9_wrist_brms@high1\endcsname{70.4}
\expandafter\def\csname result@camera_mask/dp_C9_wrist_brms@low951\endcsname{42.2}
\expandafter\def\csname result@camera_mask/dp_C9_wrist_brms@high951\endcsname{72.0}
\expandafter\def\csname result@camera_mask/dp_C9_wrist_brms@value2\endcsname{59.48}
\expandafter\def\csname result@camera_mask/dp_C9_wrist_brms@low2\endcsname{45.44}
\expandafter\def\csname result@camera_mask/dp_C9_wrist_brms@high2\endcsname{70.44}
\expandafter\def\csname result@camera_mask/dp_C9_wrist_brms@low952\endcsname{42.22}
\expandafter\def\csname result@camera_mask/dp_C9_wrist_brms@high952\endcsname{71.98}
\expandafter\def\csname result@fixed_task/G1_brms@value1\endcsname{-10.0}
\expandafter\def\csname result@fixed_task/G1_brms@low1\endcsname{-13.1}
\expandafter\def\csname result@fixed_task/G1_brms@high1\endcsname{-6.8}
\expandafter\def\csname result@fixed_task/G1_brms@low951\endcsname{-13.8}
\expandafter\def\csname result@fixed_task/G1_brms@high951\endcsname{-6.2}
\expandafter\def\csname result@fixed_task/G1_brms@value2\endcsname{-9.99}
\expandafter\def\csname result@fixed_task/G1_brms@low2\endcsname{-13.13}
\expandafter\def\csname result@fixed_task/G1_brms@high2\endcsname{-6.81}
\expandafter\def\csname result@fixed_task/G1_brms@low952\endcsname{-13.76}
\expandafter\def\csname result@fixed_task/G1_brms@high952\endcsname{-6.24}
\expandafter\def\csname result@fixed_task/H1_brms@value1\endcsname{-0.6}
\expandafter\def\csname result@fixed_task/H1_brms@low1\endcsname{-3.2}
\expandafter\def\csname result@fixed_task/H1_brms@high1\endcsname{1.9}
\expandafter\def\csname result@fixed_task/H1_brms@low951\endcsname{-3.7}
\expandafter\def\csname result@fixed_task/H1_brms@high951\endcsname{2.4}
\expandafter\def\csname result@fixed_task/H1_brms@value2\endcsname{-0.62}
\expandafter\def\csname result@fixed_task/H1_brms@low2\endcsname{-3.19}
\expandafter\def\csname result@fixed_task/H1_brms@high2\endcsname{1.89}
\expandafter\def\csname result@fixed_task/H1_brms@low952\endcsname{-3.69}
\expandafter\def\csname result@fixed_task/H1_brms@high952\endcsname{2.36}
\expandafter\def\csname result@fixed_task/H2_brms@value1\endcsname{-0.0}
\expandafter\def\csname result@fixed_task/H2_brms@low1\endcsname{-2.4}
\expandafter\def\csname result@fixed_task/H2_brms@high1\endcsname{2.4}
\expandafter\def\csname result@fixed_task/H2_brms@low951\endcsname{-2.9}
\expandafter\def\csname result@fixed_task/H2_brms@high951\endcsname{2.8}
\expandafter\def\csname result@fixed_task/H2_brms@value2\endcsname{-0.00}
\expandafter\def\csname result@fixed_task/H2_brms@low2\endcsname{-2.39}
\expandafter\def\csname result@fixed_task/H2_brms@high2\endcsname{2.36}
\expandafter\def\csname result@fixed_task/H2_brms@low952\endcsname{-2.86}
\expandafter\def\csname result@fixed_task/H2_brms@high952\endcsname{2.82}
\expandafter\def\csname result@fixed_task/H3_brms@value1\endcsname{-1.8}
\expandafter\def\csname result@fixed_task/H3_brms@low1\endcsname{-4.6}
\expandafter\def\csname result@fixed_task/H3_brms@high1\endcsname{0.9}
\expandafter\def\csname result@fixed_task/H3_brms@low951\endcsname{-5.1}
\expandafter\def\csname result@fixed_task/H3_brms@high951\endcsname{1.4}
\expandafter\def\csname result@fixed_task/H3_brms@value2\endcsname{-1.83}
\expandafter\def\csname result@fixed_task/H3_brms@low2\endcsname{-4.56}
\expandafter\def\csname result@fixed_task/H3_brms@high2\endcsname{0.89}
\expandafter\def\csname result@fixed_task/H3_brms@low952\endcsname{-5.07}
\expandafter\def\csname result@fixed_task/H3_brms@high952\endcsname{1.38}
\expandafter\def\csname result@fixed_task/H5_brms@value1\endcsname{-1.8}
\expandafter\def\csname result@fixed_task/H5_brms@low1\endcsname{-4.3}
\expandafter\def\csname result@fixed_task/H5_brms@high1\endcsname{0.6}
\expandafter\def\csname result@fixed_task/H5_brms@low951\endcsname{-4.8}
\expandafter\def\csname result@fixed_task/H5_brms@high951\endcsname{1.1}
\expandafter\def\csname result@fixed_task/H5_brms@value2\endcsname{-1.83}
\expandafter\def\csname result@fixed_task/H5_brms@low2\endcsname{-4.32}
\expandafter\def\csname result@fixed_task/H5_brms@high2\endcsname{0.62}
\expandafter\def\csname result@fixed_task/H5_brms@low952\endcsname{-4.77}
\expandafter\def\csname result@fixed_task/H5_brms@high952\endcsname{1.08}
\expandafter\def\csname result@geometry_coupling/R1b_L_brms@value1\endcsname{4.3}
\expandafter\def\csname result@geometry_coupling/R1b_L_brms@low1\endcsname{1.6}
\expandafter\def\csname result@geometry_coupling/R1b_L_brms@high1\endcsname{7.4}
\expandafter\def\csname result@geometry_coupling/R1b_L_brms@low951\endcsname{1.2}
\expandafter\def\csname result@geometry_coupling/R1b_L_brms@high951\endcsname{8.1}
\expandafter\def\csname result@geometry_coupling/R1b_L_brms@value2\endcsname{4.29}
\expandafter\def\csname result@geometry_coupling/R1b_L_brms@low2\endcsname{1.61}
\expandafter\def\csname result@geometry_coupling/R1b_L_brms@high2\endcsname{7.37}
\expandafter\def\csname result@geometry_coupling/R1b_L_brms@low952\endcsname{1.19}
\expandafter\def\csname result@geometry_coupling/R1b_L_brms@high952\endcsname{8.11}
\expandafter\def\csname result@geometry_coupling/R1b_c_brms@value1\endcsname{3.3}
\expandafter\def\csname result@geometry_coupling/R1b_c_brms@low1\endcsname{0.6}
\expandafter\def\csname result@geometry_coupling/R1b_c_brms@high1\endcsname{6.3}
\expandafter\def\csname result@geometry_coupling/R1b_c_brms@low951\endcsname{0.1}
\expandafter\def\csname result@geometry_coupling/R1b_c_brms@high951\endcsname{6.9}
\expandafter\def\csname result@geometry_coupling/R1b_c_brms@value2\endcsname{3.30}
\expandafter\def\csname result@geometry_coupling/R1b_c_brms@low2\endcsname{0.64}
\expandafter\def\csname result@geometry_coupling/R1b_c_brms@high2\endcsname{6.28}
\expandafter\def\csname result@geometry_coupling/R1b_c_brms@low952\endcsname{0.14}
\expandafter\def\csname result@geometry_coupling/R1b_c_brms@high952\endcsname{6.94}
\expandafter\def\csname result@leave_one_task_out/H0a__drop_libero_10_5_brms@value1\endcsname{1.6}
\expandafter\def\csname result@leave_one_task_out/H0a__drop_libero_10_5_brms@low1\endcsname{-1.7}
\expandafter\def\csname result@leave_one_task_out/H0a__drop_libero_10_5_brms@high1\endcsname{5.0}
\expandafter\def\csname result@leave_one_task_out/H0a__drop_libero_10_5_brms@low951\endcsname{-2.3}
\expandafter\def\csname result@leave_one_task_out/H0a__drop_libero_10_5_brms@high951\endcsname{5.8}
\expandafter\def\csname result@leave_one_task_out/H0a__drop_libero_10_5_brms@value2\endcsname{1.58}
\expandafter\def\csname result@leave_one_task_out/H0a__drop_libero_10_5_brms@low2\endcsname{-1.65}
\expandafter\def\csname result@leave_one_task_out/H0a__drop_libero_10_5_brms@high2\endcsname{5.05}
\expandafter\def\csname result@leave_one_task_out/H0a__drop_libero_10_5_brms@low952\endcsname{-2.34}
\expandafter\def\csname result@leave_one_task_out/H0a__drop_libero_10_5_brms@high952\endcsname{5.81}
\expandafter\def\csname result@leave_one_task_out/H0a__drop_libero_10_9_brms@value1\endcsname{9.3}
\expandafter\def\csname result@leave_one_task_out/H0a__drop_libero_10_9_brms@low1\endcsname{5.2}
\expandafter\def\csname result@leave_one_task_out/H0a__drop_libero_10_9_brms@high1\endcsname{13.8}
\expandafter\def\csname result@leave_one_task_out/H0a__drop_libero_10_9_brms@low951\endcsname{4.6}
\expandafter\def\csname result@leave_one_task_out/H0a__drop_libero_10_9_brms@high951\endcsname{14.8}
\expandafter\def\csname result@leave_one_task_out/H0a__drop_libero_10_9_brms@value2\endcsname{9.29}
\expandafter\def\csname result@leave_one_task_out/H0a__drop_libero_10_9_brms@low2\endcsname{5.24}
\expandafter\def\csname result@leave_one_task_out/H0a__drop_libero_10_9_brms@high2\endcsname{13.83}
\expandafter\def\csname result@leave_one_task_out/H0a__drop_libero_10_9_brms@low952\endcsname{4.62}
\expandafter\def\csname result@leave_one_task_out/H0a__drop_libero_10_9_brms@high952\endcsname{14.81}
\expandafter\def\csname result@leave_one_task_out/H0a__drop_libero_goal_1_brms@value1\endcsname{6.6}
\expandafter\def\csname result@leave_one_task_out/H0a__drop_libero_goal_1_brms@low1\endcsname{3.3}
\expandafter\def\csname result@leave_one_task_out/H0a__drop_libero_goal_1_brms@high1\endcsname{10.6}
\expandafter\def\csname result@leave_one_task_out/H0a__drop_libero_goal_1_brms@low951\endcsname{2.7}
\expandafter\def\csname result@leave_one_task_out/H0a__drop_libero_goal_1_brms@high951\endcsname{11.5}
\expandafter\def\csname result@leave_one_task_out/H0a__drop_libero_goal_1_brms@value2\endcsname{6.64}
\expandafter\def\csname result@leave_one_task_out/H0a__drop_libero_goal_1_brms@low2\endcsname{3.25}
\expandafter\def\csname result@leave_one_task_out/H0a__drop_libero_goal_1_brms@high2\endcsname{10.57}
\expandafter\def\csname result@leave_one_task_out/H0a__drop_libero_goal_1_brms@low952\endcsname{2.71}
\expandafter\def\csname result@leave_one_task_out/H0a__drop_libero_goal_1_brms@high952\endcsname{11.54}
\expandafter\def\csname result@leave_one_task_out/H0a__drop_libero_goal_6_brms@value1\endcsname{3.9}
\expandafter\def\csname result@leave_one_task_out/H0a__drop_libero_goal_6_brms@low1\endcsname{0.7}
\expandafter\def\csname result@leave_one_task_out/H0a__drop_libero_goal_6_brms@high1\endcsname{7.6}
\expandafter\def\csname result@leave_one_task_out/H0a__drop_libero_goal_6_brms@low951\endcsname{0.1}
\expandafter\def\csname result@leave_one_task_out/H0a__drop_libero_goal_6_brms@high951\endcsname{8.4}
\expandafter\def\csname result@leave_one_task_out/H0a__drop_libero_goal_6_brms@value2\endcsname{3.95}
\expandafter\def\csname result@leave_one_task_out/H0a__drop_libero_goal_6_brms@low2\endcsname{0.66}
\expandafter\def\csname result@leave_one_task_out/H0a__drop_libero_goal_6_brms@high2\endcsname{7.62}
\expandafter\def\csname result@leave_one_task_out/H0a__drop_libero_goal_6_brms@low952\endcsname{0.09}
\expandafter\def\csname result@leave_one_task_out/H0a__drop_libero_goal_6_brms@high952\endcsname{8.44}
\expandafter\def\csname result@leave_one_task_out/H0a__drop_libero_goal_7_brms@value1\endcsname{4.5}
\expandafter\def\csname result@leave_one_task_out/H0a__drop_libero_goal_7_brms@low1\endcsname{1.1}
\expandafter\def\csname result@leave_one_task_out/H0a__drop_libero_goal_7_brms@high1\endcsname{8.3}
\expandafter\def\csname result@leave_one_task_out/H0a__drop_libero_goal_7_brms@low951\endcsname{0.5}
\expandafter\def\csname result@leave_one_task_out/H0a__drop_libero_goal_7_brms@high951\endcsname{9.0}
\expandafter\def\csname result@leave_one_task_out/H0a__drop_libero_goal_7_brms@value2\endcsname{4.48}
\expandafter\def\csname result@leave_one_task_out/H0a__drop_libero_goal_7_brms@low2\endcsname{1.07}
\expandafter\def\csname result@leave_one_task_out/H0a__drop_libero_goal_7_brms@high2\endcsname{8.29}
\expandafter\def\csname result@leave_one_task_out/H0a__drop_libero_goal_7_brms@low952\endcsname{0.48}
\expandafter\def\csname result@leave_one_task_out/H0a__drop_libero_goal_7_brms@high952\endcsname{8.99}
\expandafter\def\csname result@leave_one_task_out/H0a__drop_libero_spatial_2_brms@value1\endcsname{6.1}
\expandafter\def\csname result@leave_one_task_out/H0a__drop_libero_spatial_2_brms@low1\endcsname{2.6}
\expandafter\def\csname result@leave_one_task_out/H0a__drop_libero_spatial_2_brms@high1\endcsname{10.1}
\expandafter\def\csname result@leave_one_task_out/H0a__drop_libero_spatial_2_brms@low951\endcsname{2.0}
\expandafter\def\csname result@leave_one_task_out/H0a__drop_libero_spatial_2_brms@high951\endcsname{11.0}
\expandafter\def\csname result@leave_one_task_out/H0a__drop_libero_spatial_2_brms@value2\endcsname{6.13}
\expandafter\def\csname result@leave_one_task_out/H0a__drop_libero_spatial_2_brms@low2\endcsname{2.63}
\expandafter\def\csname result@leave_one_task_out/H0a__drop_libero_spatial_2_brms@high2\endcsname{10.10}
\expandafter\def\csname result@leave_one_task_out/H0a__drop_libero_spatial_2_brms@low952\endcsname{2.04}
\expandafter\def\csname result@leave_one_task_out/H0a__drop_libero_spatial_2_brms@high952\endcsname{11.02}
\expandafter\def\csname result@leave_one_task_out/H0a__drop_libero_spatial_8_brms@value1\endcsname{6.0}
\expandafter\def\csname result@leave_one_task_out/H0a__drop_libero_spatial_8_brms@low1\endcsname{2.5}
\expandafter\def\csname result@leave_one_task_out/H0a__drop_libero_spatial_8_brms@high1\endcsname{10.0}
\expandafter\def\csname result@leave_one_task_out/H0a__drop_libero_spatial_8_brms@low951\endcsname{1.8}
\expandafter\def\csname result@leave_one_task_out/H0a__drop_libero_spatial_8_brms@high951\endcsname{10.8}
\expandafter\def\csname result@leave_one_task_out/H0a__drop_libero_spatial_8_brms@value2\endcsname{6.03}
\expandafter\def\csname result@leave_one_task_out/H0a__drop_libero_spatial_8_brms@low2\endcsname{2.45}
\expandafter\def\csname result@leave_one_task_out/H0a__drop_libero_spatial_8_brms@high2\endcsname{10.02}
\expandafter\def\csname result@leave_one_task_out/H0a__drop_libero_spatial_8_brms@low952\endcsname{1.81}
\expandafter\def\csname result@leave_one_task_out/H0a__drop_libero_spatial_8_brms@high952\endcsname{10.76}
\expandafter\def\csname result@leave_one_task_out/H0b__drop_libero_10_5_brms@value1\endcsname{6.8}
\expandafter\def\csname result@leave_one_task_out/H0b__drop_libero_10_5_brms@low1\endcsname{3.2}
\expandafter\def\csname result@leave_one_task_out/H0b__drop_libero_10_5_brms@high1\endcsname{11.0}
\expandafter\def\csname result@leave_one_task_out/H0b__drop_libero_10_5_brms@low951\endcsname{2.6}
\expandafter\def\csname result@leave_one_task_out/H0b__drop_libero_10_5_brms@high951\endcsname{11.9}
\expandafter\def\csname result@leave_one_task_out/H0b__drop_libero_10_5_brms@value2\endcsname{6.79}
\expandafter\def\csname result@leave_one_task_out/H0b__drop_libero_10_5_brms@low2\endcsname{3.21}
\expandafter\def\csname result@leave_one_task_out/H0b__drop_libero_10_5_brms@high2\endcsname{11.02}
\expandafter\def\csname result@leave_one_task_out/H0b__drop_libero_10_5_brms@low952\endcsname{2.63}
\expandafter\def\csname result@leave_one_task_out/H0b__drop_libero_10_5_brms@high952\endcsname{11.95}
\expandafter\def\csname result@leave_one_task_out/H0b__drop_libero_10_9_brms@value1\endcsname{8.8}
\expandafter\def\csname result@leave_one_task_out/H0b__drop_libero_10_9_brms@low1\endcsname{4.7}
\expandafter\def\csname result@leave_one_task_out/H0b__drop_libero_10_9_brms@high1\endcsname{13.3}
\expandafter\def\csname result@leave_one_task_out/H0b__drop_libero_10_9_brms@low951\endcsname{4.2}
\expandafter\def\csname result@leave_one_task_out/H0b__drop_libero_10_9_brms@high951\endcsname{14.2}
\expandafter\def\csname result@leave_one_task_out/H0b__drop_libero_10_9_brms@value2\endcsname{8.76}
\expandafter\def\csname result@leave_one_task_out/H0b__drop_libero_10_9_brms@low2\endcsname{4.74}
\expandafter\def\csname result@leave_one_task_out/H0b__drop_libero_10_9_brms@high2\endcsname{13.27}
\expandafter\def\csname result@leave_one_task_out/H0b__drop_libero_10_9_brms@low952\endcsname{4.17}
\expandafter\def\csname result@leave_one_task_out/H0b__drop_libero_10_9_brms@high952\endcsname{14.22}
\expandafter\def\csname result@leave_one_task_out/H0b__drop_libero_goal_1_brms@value1\endcsname{11.5}
\expandafter\def\csname result@leave_one_task_out/H0b__drop_libero_goal_1_brms@low1\endcsname{7.2}
\expandafter\def\csname result@leave_one_task_out/H0b__drop_libero_goal_1_brms@high1\endcsname{16.6}
\expandafter\def\csname result@leave_one_task_out/H0b__drop_libero_goal_1_brms@low951\endcsname{6.6}
\expandafter\def\csname result@leave_one_task_out/H0b__drop_libero_goal_1_brms@high951\endcsname{17.8}
\expandafter\def\csname result@leave_one_task_out/H0b__drop_libero_goal_1_brms@value2\endcsname{11.48}
\expandafter\def\csname result@leave_one_task_out/H0b__drop_libero_goal_1_brms@low2\endcsname{7.18}
\expandafter\def\csname result@leave_one_task_out/H0b__drop_libero_goal_1_brms@high2\endcsname{16.59}
\expandafter\def\csname result@leave_one_task_out/H0b__drop_libero_goal_1_brms@low952\endcsname{6.56}
\expandafter\def\csname result@leave_one_task_out/H0b__drop_libero_goal_1_brms@high952\endcsname{17.80}
\expandafter\def\csname result@leave_one_task_out/H0b__drop_libero_goal_6_brms@value1\endcsname{8.1}
\expandafter\def\csname result@leave_one_task_out/H0b__drop_libero_goal_6_brms@low1\endcsname{4.3}
\expandafter\def\csname result@leave_one_task_out/H0b__drop_libero_goal_6_brms@high1\endcsname{12.7}
\expandafter\def\csname result@leave_one_task_out/H0b__drop_libero_goal_6_brms@low951\endcsname{3.7}
\expandafter\def\csname result@leave_one_task_out/H0b__drop_libero_goal_6_brms@high951\endcsname{13.6}
\expandafter\def\csname result@leave_one_task_out/H0b__drop_libero_goal_6_brms@value2\endcsname{8.12}
\expandafter\def\csname result@leave_one_task_out/H0b__drop_libero_goal_6_brms@low2\endcsname{4.29}
\expandafter\def\csname result@leave_one_task_out/H0b__drop_libero_goal_6_brms@high2\endcsname{12.66}
\expandafter\def\csname result@leave_one_task_out/H0b__drop_libero_goal_6_brms@low952\endcsname{3.65}
\expandafter\def\csname result@leave_one_task_out/H0b__drop_libero_goal_6_brms@high952\endcsname{13.59}
\expandafter\def\csname result@leave_one_task_out/H0b__drop_libero_goal_7_brms@value1\endcsname{8.8}
\expandafter\def\csname result@leave_one_task_out/H0b__drop_libero_goal_7_brms@low1\endcsname{4.7}
\expandafter\def\csname result@leave_one_task_out/H0b__drop_libero_goal_7_brms@high1\endcsname{13.3}
\expandafter\def\csname result@leave_one_task_out/H0b__drop_libero_goal_7_brms@low951\endcsname{4.2}
\expandafter\def\csname result@leave_one_task_out/H0b__drop_libero_goal_7_brms@high951\endcsname{14.4}
\expandafter\def\csname result@leave_one_task_out/H0b__drop_libero_goal_7_brms@value2\endcsname{8.76}
\expandafter\def\csname result@leave_one_task_out/H0b__drop_libero_goal_7_brms@low2\endcsname{4.74}
\expandafter\def\csname result@leave_one_task_out/H0b__drop_libero_goal_7_brms@high2\endcsname{13.32}
\expandafter\def\csname result@leave_one_task_out/H0b__drop_libero_goal_7_brms@low952\endcsname{4.15}
\expandafter\def\csname result@leave_one_task_out/H0b__drop_libero_goal_7_brms@high952\endcsname{14.39}
\expandafter\def\csname result@leave_one_task_out/H0b__drop_libero_spatial_2_brms@value1\endcsname{8.9}
\expandafter\def\csname result@leave_one_task_out/H0b__drop_libero_spatial_2_brms@low1\endcsname{5.0}
\expandafter\def\csname result@leave_one_task_out/H0b__drop_libero_spatial_2_brms@high1\endcsname{13.4}
\expandafter\def\csname result@leave_one_task_out/H0b__drop_libero_spatial_2_brms@low951\endcsname{4.4}
\expandafter\def\csname result@leave_one_task_out/H0b__drop_libero_spatial_2_brms@high951\endcsname{14.3}
\expandafter\def\csname result@leave_one_task_out/H0b__drop_libero_spatial_2_brms@value2\endcsname{8.95}
\expandafter\def\csname result@leave_one_task_out/H0b__drop_libero_spatial_2_brms@low2\endcsname{4.97}
\expandafter\def\csname result@leave_one_task_out/H0b__drop_libero_spatial_2_brms@high2\endcsname{13.40}
\expandafter\def\csname result@leave_one_task_out/H0b__drop_libero_spatial_2_brms@low952\endcsname{4.39}
\expandafter\def\csname result@leave_one_task_out/H0b__drop_libero_spatial_2_brms@high952\endcsname{14.32}
\expandafter\def\csname result@leave_one_task_out/H0b__drop_libero_spatial_8_brms@value1\endcsname{10.2}
\expandafter\def\csname result@leave_one_task_out/H0b__drop_libero_spatial_8_brms@low1\endcsname{5.9}
\expandafter\def\csname result@leave_one_task_out/H0b__drop_libero_spatial_8_brms@high1\endcsname{15.0}
\expandafter\def\csname result@leave_one_task_out/H0b__drop_libero_spatial_8_brms@low951\endcsname{5.2}
\expandafter\def\csname result@leave_one_task_out/H0b__drop_libero_spatial_8_brms@high951\endcsname{16.0}
\expandafter\def\csname result@leave_one_task_out/H0b__drop_libero_spatial_8_brms@value2\endcsname{10.20}
\expandafter\def\csname result@leave_one_task_out/H0b__drop_libero_spatial_8_brms@low2\endcsname{5.89}
\expandafter\def\csname result@leave_one_task_out/H0b__drop_libero_spatial_8_brms@high2\endcsname{14.96}
\expandafter\def\csname result@leave_one_task_out/H0b__drop_libero_spatial_8_brms@low952\endcsname{5.21}
\expandafter\def\csname result@leave_one_task_out/H0b__drop_libero_spatial_8_brms@high952\endcsname{15.96}
\expandafter\def\csname result@leave_one_task_out/H0c__drop_libero_10_5_brms@value1\endcsname{5.6}
\expandafter\def\csname result@leave_one_task_out/H0c__drop_libero_10_5_brms@low1\endcsname{1.9}
\expandafter\def\csname result@leave_one_task_out/H0c__drop_libero_10_5_brms@high1\endcsname{9.9}
\expandafter\def\csname result@leave_one_task_out/H0c__drop_libero_10_5_brms@low951\endcsname{1.3}
\expandafter\def\csname result@leave_one_task_out/H0c__drop_libero_10_5_brms@high951\endcsname{10.9}
\expandafter\def\csname result@leave_one_task_out/H0c__drop_libero_10_5_brms@value2\endcsname{5.59}
\expandafter\def\csname result@leave_one_task_out/H0c__drop_libero_10_5_brms@low2\endcsname{1.88}
\expandafter\def\csname result@leave_one_task_out/H0c__drop_libero_10_5_brms@high2\endcsname{9.94}
\expandafter\def\csname result@leave_one_task_out/H0c__drop_libero_10_5_brms@low952\endcsname{1.33}
\expandafter\def\csname result@leave_one_task_out/H0c__drop_libero_10_5_brms@high952\endcsname{10.87}
\expandafter\def\csname result@leave_one_task_out/H0c__drop_libero_10_9_brms@value1\endcsname{8.3}
\expandafter\def\csname result@leave_one_task_out/H0c__drop_libero_10_9_brms@low1\endcsname{4.1}
\expandafter\def\csname result@leave_one_task_out/H0c__drop_libero_10_9_brms@high1\endcsname{13.2}
\expandafter\def\csname result@leave_one_task_out/H0c__drop_libero_10_9_brms@low951\endcsname{3.4}
\expandafter\def\csname result@leave_one_task_out/H0c__drop_libero_10_9_brms@high951\endcsname{14.2}
\expandafter\def\csname result@leave_one_task_out/H0c__drop_libero_10_9_brms@value2\endcsname{8.27}
\expandafter\def\csname result@leave_one_task_out/H0c__drop_libero_10_9_brms@low2\endcsname{4.06}
\expandafter\def\csname result@leave_one_task_out/H0c__drop_libero_10_9_brms@high2\endcsname{13.15}
\expandafter\def\csname result@leave_one_task_out/H0c__drop_libero_10_9_brms@low952\endcsname{3.44}
\expandafter\def\csname result@leave_one_task_out/H0c__drop_libero_10_9_brms@high952\endcsname{14.17}
\expandafter\def\csname result@leave_one_task_out/H0c__drop_libero_goal_1_brms@value1\endcsname{10.3}
\expandafter\def\csname result@leave_one_task_out/H0c__drop_libero_goal_1_brms@low1\endcsname{5.7}
\expandafter\def\csname result@leave_one_task_out/H0c__drop_libero_goal_1_brms@high1\endcsname{15.8}
\expandafter\def\csname result@leave_one_task_out/H0c__drop_libero_goal_1_brms@low951\endcsname{5.0}
\expandafter\def\csname result@leave_one_task_out/H0c__drop_libero_goal_1_brms@high951\endcsname{16.9}
\expandafter\def\csname result@leave_one_task_out/H0c__drop_libero_goal_1_brms@value2\endcsname{10.29}
\expandafter\def\csname result@leave_one_task_out/H0c__drop_libero_goal_1_brms@low2\endcsname{5.69}
\expandafter\def\csname result@leave_one_task_out/H0c__drop_libero_goal_1_brms@high2\endcsname{15.78}
\expandafter\def\csname result@leave_one_task_out/H0c__drop_libero_goal_1_brms@low952\endcsname{4.97}
\expandafter\def\csname result@leave_one_task_out/H0c__drop_libero_goal_1_brms@high952\endcsname{16.86}
\expandafter\def\csname result@leave_one_task_out/H0c__drop_libero_goal_6_brms@value1\endcsname{9.1}
\expandafter\def\csname result@leave_one_task_out/H0c__drop_libero_goal_6_brms@low1\endcsname{4.7}
\expandafter\def\csname result@leave_one_task_out/H0c__drop_libero_goal_6_brms@high1\endcsname{13.9}
\expandafter\def\csname result@leave_one_task_out/H0c__drop_libero_goal_6_brms@low951\endcsname{4.1}
\expandafter\def\csname result@leave_one_task_out/H0c__drop_libero_goal_6_brms@high951\endcsname{14.8}
\expandafter\def\csname result@leave_one_task_out/H0c__drop_libero_goal_6_brms@value2\endcsname{9.08}
\expandafter\def\csname result@leave_one_task_out/H0c__drop_libero_goal_6_brms@low2\endcsname{4.73}
\expandafter\def\csname result@leave_one_task_out/H0c__drop_libero_goal_6_brms@high2\endcsname{13.90}
\expandafter\def\csname result@leave_one_task_out/H0c__drop_libero_goal_6_brms@low952\endcsname{4.07}
\expandafter\def\csname result@leave_one_task_out/H0c__drop_libero_goal_6_brms@high952\endcsname{14.83}
\expandafter\def\csname result@leave_one_task_out/H0c__drop_libero_goal_7_brms@value1\endcsname{8.2}
\expandafter\def\csname result@leave_one_task_out/H0c__drop_libero_goal_7_brms@low1\endcsname{4.1}
\expandafter\def\csname result@leave_one_task_out/H0c__drop_libero_goal_7_brms@high1\endcsname{13.0}
\expandafter\def\csname result@leave_one_task_out/H0c__drop_libero_goal_7_brms@low951\endcsname{3.5}
\expandafter\def\csname result@leave_one_task_out/H0c__drop_libero_goal_7_brms@high951\endcsname{14.0}
\expandafter\def\csname result@leave_one_task_out/H0c__drop_libero_goal_7_brms@value2\endcsname{8.24}
\expandafter\def\csname result@leave_one_task_out/H0c__drop_libero_goal_7_brms@low2\endcsname{4.10}
\expandafter\def\csname result@leave_one_task_out/H0c__drop_libero_goal_7_brms@high2\endcsname{13.03}
\expandafter\def\csname result@leave_one_task_out/H0c__drop_libero_goal_7_brms@low952\endcsname{3.54}
\expandafter\def\csname result@leave_one_task_out/H0c__drop_libero_goal_7_brms@high952\endcsname{14.04}
\expandafter\def\csname result@leave_one_task_out/H0c__drop_libero_spatial_2_brms@value1\endcsname{8.8}
\expandafter\def\csname result@leave_one_task_out/H0c__drop_libero_spatial_2_brms@low1\endcsname{4.6}
\expandafter\def\csname result@leave_one_task_out/H0c__drop_libero_spatial_2_brms@high1\endcsname{13.6}
\expandafter\def\csname result@leave_one_task_out/H0c__drop_libero_spatial_2_brms@low951\endcsname{3.9}
\expandafter\def\csname result@leave_one_task_out/H0c__drop_libero_spatial_2_brms@high951\endcsname{14.6}
\expandafter\def\csname result@leave_one_task_out/H0c__drop_libero_spatial_2_brms@value2\endcsname{8.79}
\expandafter\def\csname result@leave_one_task_out/H0c__drop_libero_spatial_2_brms@low2\endcsname{4.58}
\expandafter\def\csname result@leave_one_task_out/H0c__drop_libero_spatial_2_brms@high2\endcsname{13.62}
\expandafter\def\csname result@leave_one_task_out/H0c__drop_libero_spatial_2_brms@low952\endcsname{3.89}
\expandafter\def\csname result@leave_one_task_out/H0c__drop_libero_spatial_2_brms@high952\endcsname{14.59}
\expandafter\def\csname result@leave_one_task_out/H0c__drop_libero_spatial_8_brms@value1\endcsname{9.1}
\expandafter\def\csname result@leave_one_task_out/H0c__drop_libero_spatial_8_brms@low1\endcsname{4.7}
\expandafter\def\csname result@leave_one_task_out/H0c__drop_libero_spatial_8_brms@high1\endcsname{14.0}
\expandafter\def\csname result@leave_one_task_out/H0c__drop_libero_spatial_8_brms@low951\endcsname{4.0}
\expandafter\def\csname result@leave_one_task_out/H0c__drop_libero_spatial_8_brms@high951\endcsname{14.9}
\expandafter\def\csname result@leave_one_task_out/H0c__drop_libero_spatial_8_brms@value2\endcsname{9.06}
\expandafter\def\csname result@leave_one_task_out/H0c__drop_libero_spatial_8_brms@low2\endcsname{4.71}
\expandafter\def\csname result@leave_one_task_out/H0c__drop_libero_spatial_8_brms@high2\endcsname{13.95}
\expandafter\def\csname result@leave_one_task_out/H0c__drop_libero_spatial_8_brms@low952\endcsname{4.04}
\expandafter\def\csname result@leave_one_task_out/H0c__drop_libero_spatial_8_brms@high952\endcsname{14.86}
\expandafter\def\csname result@leave_one_task_out/H1__drop_libero_10_5_brms@value1\endcsname{-0.9}
\expandafter\def\csname result@leave_one_task_out/H1__drop_libero_10_5_brms@low1\endcsname{-3.9}
\expandafter\def\csname result@leave_one_task_out/H1__drop_libero_10_5_brms@high1\endcsname{2.0}
\expandafter\def\csname result@leave_one_task_out/H1__drop_libero_10_5_brms@low951\endcsname{-4.6}
\expandafter\def\csname result@leave_one_task_out/H1__drop_libero_10_5_brms@high951\endcsname{2.6}
\expandafter\def\csname result@leave_one_task_out/H1__drop_libero_10_5_brms@value2\endcsname{-0.89}
\expandafter\def\csname result@leave_one_task_out/H1__drop_libero_10_5_brms@low2\endcsname{-3.92}
\expandafter\def\csname result@leave_one_task_out/H1__drop_libero_10_5_brms@high2\endcsname{1.98}
\expandafter\def\csname result@leave_one_task_out/H1__drop_libero_10_5_brms@low952\endcsname{-4.64}
\expandafter\def\csname result@leave_one_task_out/H1__drop_libero_10_5_brms@high952\endcsname{2.57}
\expandafter\def\csname result@leave_one_task_out/H1__drop_libero_10_9_brms@value1\endcsname{0.4}
\expandafter\def\csname result@leave_one_task_out/H1__drop_libero_10_9_brms@low1\endcsname{-2.5}
\expandafter\def\csname result@leave_one_task_out/H1__drop_libero_10_9_brms@high1\endcsname{3.3}
\expandafter\def\csname result@leave_one_task_out/H1__drop_libero_10_9_brms@low951\endcsname{-3.1}
\expandafter\def\csname result@leave_one_task_out/H1__drop_libero_10_9_brms@high951\endcsname{4.0}
\expandafter\def\csname result@leave_one_task_out/H1__drop_libero_10_9_brms@value2\endcsname{0.37}
\expandafter\def\csname result@leave_one_task_out/H1__drop_libero_10_9_brms@low2\endcsname{-2.51}
\expandafter\def\csname result@leave_one_task_out/H1__drop_libero_10_9_brms@high2\endcsname{3.30}
\expandafter\def\csname result@leave_one_task_out/H1__drop_libero_10_9_brms@low952\endcsname{-3.06}
\expandafter\def\csname result@leave_one_task_out/H1__drop_libero_10_9_brms@high952\endcsname{3.98}
\expandafter\def\csname result@leave_one_task_out/H1__drop_libero_goal_1_brms@value1\endcsname{-1.5}
\expandafter\def\csname result@leave_one_task_out/H1__drop_libero_goal_1_brms@low1\endcsname{-4.2}
\expandafter\def\csname result@leave_one_task_out/H1__drop_libero_goal_1_brms@high1\endcsname{1.1}
\expandafter\def\csname result@leave_one_task_out/H1__drop_libero_goal_1_brms@low951\endcsname{-4.8}
\expandafter\def\csname result@leave_one_task_out/H1__drop_libero_goal_1_brms@high951\endcsname{1.5}
\expandafter\def\csname result@leave_one_task_out/H1__drop_libero_goal_1_brms@value2\endcsname{-1.48}
\expandafter\def\csname result@leave_one_task_out/H1__drop_libero_goal_1_brms@low2\endcsname{-4.18}
\expandafter\def\csname result@leave_one_task_out/H1__drop_libero_goal_1_brms@high2\endcsname{1.08}
\expandafter\def\csname result@leave_one_task_out/H1__drop_libero_goal_1_brms@low952\endcsname{-4.82}
\expandafter\def\csname result@leave_one_task_out/H1__drop_libero_goal_1_brms@high952\endcsname{1.54}
\expandafter\def\csname result@leave_one_task_out/H1__drop_libero_goal_6_brms@value1\endcsname{-0.4}
\expandafter\def\csname result@leave_one_task_out/H1__drop_libero_goal_6_brms@low1\endcsname{-3.4}
\expandafter\def\csname result@leave_one_task_out/H1__drop_libero_goal_6_brms@high1\endcsname{2.5}
\expandafter\def\csname result@leave_one_task_out/H1__drop_libero_goal_6_brms@low951\endcsname{-4.0}
\expandafter\def\csname result@leave_one_task_out/H1__drop_libero_goal_6_brms@high951\endcsname{3.2}
\expandafter\def\csname result@leave_one_task_out/H1__drop_libero_goal_6_brms@value2\endcsname{-0.38}
\expandafter\def\csname result@leave_one_task_out/H1__drop_libero_goal_6_brms@low2\endcsname{-3.38}
\expandafter\def\csname result@leave_one_task_out/H1__drop_libero_goal_6_brms@high2\endcsname{2.54}
\expandafter\def\csname result@leave_one_task_out/H1__drop_libero_goal_6_brms@low952\endcsname{-3.99}
\expandafter\def\csname result@leave_one_task_out/H1__drop_libero_goal_6_brms@high952\endcsname{3.17}
\expandafter\def\csname result@leave_one_task_out/H1__drop_libero_goal_7_brms@value1\endcsname{-0.7}
\expandafter\def\csname result@leave_one_task_out/H1__drop_libero_goal_7_brms@low1\endcsname{-3.7}
\expandafter\def\csname result@leave_one_task_out/H1__drop_libero_goal_7_brms@high1\endcsname{2.2}
\expandafter\def\csname result@leave_one_task_out/H1__drop_libero_goal_7_brms@low951\endcsname{-4.4}
\expandafter\def\csname result@leave_one_task_out/H1__drop_libero_goal_7_brms@high951\endcsname{2.8}
\expandafter\def\csname result@leave_one_task_out/H1__drop_libero_goal_7_brms@value2\endcsname{-0.71}
\expandafter\def\csname result@leave_one_task_out/H1__drop_libero_goal_7_brms@low2\endcsname{-3.74}
\expandafter\def\csname result@leave_one_task_out/H1__drop_libero_goal_7_brms@high2\endcsname{2.21}
\expandafter\def\csname result@leave_one_task_out/H1__drop_libero_goal_7_brms@low952\endcsname{-4.36}
\expandafter\def\csname result@leave_one_task_out/H1__drop_libero_goal_7_brms@high952\endcsname{2.84}
\expandafter\def\csname result@leave_one_task_out/H1__drop_libero_spatial_2_brms@value1\endcsname{-0.0}
\expandafter\def\csname result@leave_one_task_out/H1__drop_libero_spatial_2_brms@low1\endcsname{-3.0}
\expandafter\def\csname result@leave_one_task_out/H1__drop_libero_spatial_2_brms@high1\endcsname{3.0}
\expandafter\def\csname result@leave_one_task_out/H1__drop_libero_spatial_2_brms@low951\endcsname{-3.7}
\expandafter\def\csname result@leave_one_task_out/H1__drop_libero_spatial_2_brms@high951\endcsname{3.6}
\expandafter\def\csname result@leave_one_task_out/H1__drop_libero_spatial_2_brms@value2\endcsname{-0.00}
\expandafter\def\csname result@leave_one_task_out/H1__drop_libero_spatial_2_brms@low2\endcsname{-3.01}
\expandafter\def\csname result@leave_one_task_out/H1__drop_libero_spatial_2_brms@high2\endcsname{2.99}
\expandafter\def\csname result@leave_one_task_out/H1__drop_libero_spatial_2_brms@low952\endcsname{-3.69}
\expandafter\def\csname result@leave_one_task_out/H1__drop_libero_spatial_2_brms@high952\endcsname{3.65}
\expandafter\def\csname result@leave_one_task_out/H1__drop_libero_spatial_8_brms@value1\endcsname{-1.3}
\expandafter\def\csname result@leave_one_task_out/H1__drop_libero_spatial_8_brms@low1\endcsname{-4.4}
\expandafter\def\csname result@leave_one_task_out/H1__drop_libero_spatial_8_brms@high1\endcsname{1.5}
\expandafter\def\csname result@leave_one_task_out/H1__drop_libero_spatial_8_brms@low951\endcsname{-5.1}
\expandafter\def\csname result@leave_one_task_out/H1__drop_libero_spatial_8_brms@high951\endcsname{2.1}
\expandafter\def\csname result@leave_one_task_out/H1__drop_libero_spatial_8_brms@value2\endcsname{-1.34}
\expandafter\def\csname result@leave_one_task_out/H1__drop_libero_spatial_8_brms@low2\endcsname{-4.42}
\expandafter\def\csname result@leave_one_task_out/H1__drop_libero_spatial_8_brms@high2\endcsname{1.50}
\expandafter\def\csname result@leave_one_task_out/H1__drop_libero_spatial_8_brms@low952\endcsname{-5.12}
\expandafter\def\csname result@leave_one_task_out/H1__drop_libero_spatial_8_brms@high952\endcsname{2.14}
\expandafter\def\csname result@leave_one_task_out/H2__drop_libero_10_5_brms@value1\endcsname{-0.0}
\expandafter\def\csname result@leave_one_task_out/H2__drop_libero_10_5_brms@low1\endcsname{-2.7}
\expandafter\def\csname result@leave_one_task_out/H2__drop_libero_10_5_brms@high1\endcsname{2.6}
\expandafter\def\csname result@leave_one_task_out/H2__drop_libero_10_5_brms@low951\endcsname{-3.3}
\expandafter\def\csname result@leave_one_task_out/H2__drop_libero_10_5_brms@high951\endcsname{3.2}
\expandafter\def\csname result@leave_one_task_out/H2__drop_libero_10_5_brms@value2\endcsname{-0.01}
\expandafter\def\csname result@leave_one_task_out/H2__drop_libero_10_5_brms@low2\endcsname{-2.67}
\expandafter\def\csname result@leave_one_task_out/H2__drop_libero_10_5_brms@high2\endcsname{2.59}
\expandafter\def\csname result@leave_one_task_out/H2__drop_libero_10_5_brms@low952\endcsname{-3.26}
\expandafter\def\csname result@leave_one_task_out/H2__drop_libero_10_5_brms@high952\endcsname{3.17}
\expandafter\def\csname result@leave_one_task_out/H2__drop_libero_10_9_brms@value1\endcsname{0.3}
\expandafter\def\csname result@leave_one_task_out/H2__drop_libero_10_9_brms@low1\endcsname{-2.5}
\expandafter\def\csname result@leave_one_task_out/H2__drop_libero_10_9_brms@high1\endcsname{3.2}
\expandafter\def\csname result@leave_one_task_out/H2__drop_libero_10_9_brms@low951\endcsname{-3.1}
\expandafter\def\csname result@leave_one_task_out/H2__drop_libero_10_9_brms@high951\endcsname{3.8}
\expandafter\def\csname result@leave_one_task_out/H2__drop_libero_10_9_brms@value2\endcsname{0.35}
\expandafter\def\csname result@leave_one_task_out/H2__drop_libero_10_9_brms@low2\endcsname{-2.48}
\expandafter\def\csname result@leave_one_task_out/H2__drop_libero_10_9_brms@high2\endcsname{3.25}
\expandafter\def\csname result@leave_one_task_out/H2__drop_libero_10_9_brms@low952\endcsname{-3.12}
\expandafter\def\csname result@leave_one_task_out/H2__drop_libero_10_9_brms@high952\endcsname{3.85}
\expandafter\def\csname result@leave_one_task_out/H2__drop_libero_goal_1_brms@value1\endcsname{-0.2}
\expandafter\def\csname result@leave_one_task_out/H2__drop_libero_goal_1_brms@low1\endcsname{-2.6}
\expandafter\def\csname result@leave_one_task_out/H2__drop_libero_goal_1_brms@high1\endcsname{2.1}
\expandafter\def\csname result@leave_one_task_out/H2__drop_libero_goal_1_brms@low951\endcsname{-3.2}
\expandafter\def\csname result@leave_one_task_out/H2__drop_libero_goal_1_brms@high951\endcsname{2.7}
\expandafter\def\csname result@leave_one_task_out/H2__drop_libero_goal_1_brms@value2\endcsname{-0.24}
\expandafter\def\csname result@leave_one_task_out/H2__drop_libero_goal_1_brms@low2\endcsname{-2.65}
\expandafter\def\csname result@leave_one_task_out/H2__drop_libero_goal_1_brms@high2\endcsname{2.13}
\expandafter\def\csname result@leave_one_task_out/H2__drop_libero_goal_1_brms@low952\endcsname{-3.17}
\expandafter\def\csname result@leave_one_task_out/H2__drop_libero_goal_1_brms@high952\endcsname{2.67}
\expandafter\def\csname result@leave_one_task_out/H2__drop_libero_goal_6_brms@value1\endcsname{0.4}
\expandafter\def\csname result@leave_one_task_out/H2__drop_libero_goal_6_brms@low1\endcsname{-2.4}
\expandafter\def\csname result@leave_one_task_out/H2__drop_libero_goal_6_brms@high1\endcsname{3.2}
\expandafter\def\csname result@leave_one_task_out/H2__drop_libero_goal_6_brms@low951\endcsname{-3.0}
\expandafter\def\csname result@leave_one_task_out/H2__drop_libero_goal_6_brms@high951\endcsname{3.8}
\expandafter\def\csname result@leave_one_task_out/H2__drop_libero_goal_6_brms@value2\endcsname{0.37}
\expandafter\def\csname result@leave_one_task_out/H2__drop_libero_goal_6_brms@low2\endcsname{-2.36}
\expandafter\def\csname result@leave_one_task_out/H2__drop_libero_goal_6_brms@high2\endcsname{3.21}
\expandafter\def\csname result@leave_one_task_out/H2__drop_libero_goal_6_brms@low952\endcsname{-2.95}
\expandafter\def\csname result@leave_one_task_out/H2__drop_libero_goal_6_brms@high952\endcsname{3.83}
\expandafter\def\csname result@leave_one_task_out/H2__drop_libero_goal_7_brms@value1\endcsname{-0.2}
\expandafter\def\csname result@leave_one_task_out/H2__drop_libero_goal_7_brms@low1\endcsname{-3.1}
\expandafter\def\csname result@leave_one_task_out/H2__drop_libero_goal_7_brms@high1\endcsname{2.5}
\expandafter\def\csname result@leave_one_task_out/H2__drop_libero_goal_7_brms@low951\endcsname{-3.7}
\expandafter\def\csname result@leave_one_task_out/H2__drop_libero_goal_7_brms@high951\endcsname{3.1}
\expandafter\def\csname result@leave_one_task_out/H2__drop_libero_goal_7_brms@value2\endcsname{-0.24}
\expandafter\def\csname result@leave_one_task_out/H2__drop_libero_goal_7_brms@low2\endcsname{-3.07}
\expandafter\def\csname result@leave_one_task_out/H2__drop_libero_goal_7_brms@high2\endcsname{2.51}
\expandafter\def\csname result@leave_one_task_out/H2__drop_libero_goal_7_brms@low952\endcsname{-3.67}
\expandafter\def\csname result@leave_one_task_out/H2__drop_libero_goal_7_brms@high952\endcsname{3.12}
\expandafter\def\csname result@leave_one_task_out/H2__drop_libero_spatial_2_brms@value1\endcsname{0.1}
\expandafter\def\csname result@leave_one_task_out/H2__drop_libero_spatial_2_brms@low1\endcsname{-2.7}
\expandafter\def\csname result@leave_one_task_out/H2__drop_libero_spatial_2_brms@high1\endcsname{2.9}
\expandafter\def\csname result@leave_one_task_out/H2__drop_libero_spatial_2_brms@low951\endcsname{-3.3}
\expandafter\def\csname result@leave_one_task_out/H2__drop_libero_spatial_2_brms@high951\endcsname{3.5}
\expandafter\def\csname result@leave_one_task_out/H2__drop_libero_spatial_2_brms@value2\endcsname{0.12}
\expandafter\def\csname result@leave_one_task_out/H2__drop_libero_spatial_2_brms@low2\endcsname{-2.73}
\expandafter\def\csname result@leave_one_task_out/H2__drop_libero_spatial_2_brms@high2\endcsname{2.94}
\expandafter\def\csname result@leave_one_task_out/H2__drop_libero_spatial_2_brms@low952\endcsname{-3.33}
\expandafter\def\csname result@leave_one_task_out/H2__drop_libero_spatial_2_brms@high952\endcsname{3.52}
\expandafter\def\csname result@leave_one_task_out/H2__drop_libero_spatial_8_brms@value1\endcsname{-0.4}
\expandafter\def\csname result@leave_one_task_out/H2__drop_libero_spatial_8_brms@low1\endcsname{-3.2}
\expandafter\def\csname result@leave_one_task_out/H2__drop_libero_spatial_8_brms@high1\endcsname{2.4}
\expandafter\def\csname result@leave_one_task_out/H2__drop_libero_spatial_8_brms@low951\endcsname{-3.8}
\expandafter\def\csname result@leave_one_task_out/H2__drop_libero_spatial_8_brms@high951\endcsname{3.0}
\expandafter\def\csname result@leave_one_task_out/H2__drop_libero_spatial_8_brms@value2\endcsname{-0.36}
\expandafter\def\csname result@leave_one_task_out/H2__drop_libero_spatial_8_brms@low2\endcsname{-3.19}
\expandafter\def\csname result@leave_one_task_out/H2__drop_libero_spatial_8_brms@high2\endcsname{2.41}
\expandafter\def\csname result@leave_one_task_out/H2__drop_libero_spatial_8_brms@low952\endcsname{-3.79}
\expandafter\def\csname result@leave_one_task_out/H2__drop_libero_spatial_8_brms@high952\endcsname{2.97}
\expandafter\def\csname result@leave_one_task_out/H3__drop_libero_10_5_brms@value1\endcsname{-3.3}
\expandafter\def\csname result@leave_one_task_out/H3__drop_libero_10_5_brms@low1\endcsname{-6.8}
\expandafter\def\csname result@leave_one_task_out/H3__drop_libero_10_5_brms@high1\endcsname{-0.1}
\expandafter\def\csname result@leave_one_task_out/H3__drop_libero_10_5_brms@low951\endcsname{-7.6}
\expandafter\def\csname result@leave_one_task_out/H3__drop_libero_10_5_brms@high951\endcsname{0.5}
\expandafter\def\csname result@leave_one_task_out/H3__drop_libero_10_5_brms@value2\endcsname{-3.26}
\expandafter\def\csname result@leave_one_task_out/H3__drop_libero_10_5_brms@low2\endcsname{-6.79}
\expandafter\def\csname result@leave_one_task_out/H3__drop_libero_10_5_brms@high2\endcsname{-0.10}
\expandafter\def\csname result@leave_one_task_out/H3__drop_libero_10_5_brms@low952\endcsname{-7.58}
\expandafter\def\csname result@leave_one_task_out/H3__drop_libero_10_5_brms@high952\endcsname{0.49}
\expandafter\def\csname result@leave_one_task_out/H3__drop_libero_10_9_brms@value1\endcsname{0.3}
\expandafter\def\csname result@leave_one_task_out/H3__drop_libero_10_9_brms@low1\endcsname{-2.7}
\expandafter\def\csname result@leave_one_task_out/H3__drop_libero_10_9_brms@high1\endcsname{3.4}
\expandafter\def\csname result@leave_one_task_out/H3__drop_libero_10_9_brms@low951\endcsname{-3.4}
\expandafter\def\csname result@leave_one_task_out/H3__drop_libero_10_9_brms@high951\endcsname{4.1}
\expandafter\def\csname result@leave_one_task_out/H3__drop_libero_10_9_brms@value2\endcsname{0.32}
\expandafter\def\csname result@leave_one_task_out/H3__drop_libero_10_9_brms@low2\endcsname{-2.71}
\expandafter\def\csname result@leave_one_task_out/H3__drop_libero_10_9_brms@high2\endcsname{3.42}
\expandafter\def\csname result@leave_one_task_out/H3__drop_libero_10_9_brms@low952\endcsname{-3.40}
\expandafter\def\csname result@leave_one_task_out/H3__drop_libero_10_9_brms@high952\endcsname{4.13}
\expandafter\def\csname result@leave_one_task_out/H3__drop_libero_goal_1_brms@value1\endcsname{-1.5}
\expandafter\def\csname result@leave_one_task_out/H3__drop_libero_goal_1_brms@low1\endcsname{-4.4}
\expandafter\def\csname result@leave_one_task_out/H3__drop_libero_goal_1_brms@high1\endcsname{1.3}
\expandafter\def\csname result@leave_one_task_out/H3__drop_libero_goal_1_brms@low951\endcsname{-5.1}
\expandafter\def\csname result@leave_one_task_out/H3__drop_libero_goal_1_brms@high951\endcsname{1.9}
\expandafter\def\csname result@leave_one_task_out/H3__drop_libero_goal_1_brms@value2\endcsname{-1.52}
\expandafter\def\csname result@leave_one_task_out/H3__drop_libero_goal_1_brms@low2\endcsname{-4.43}
\expandafter\def\csname result@leave_one_task_out/H3__drop_libero_goal_1_brms@high2\endcsname{1.28}
\expandafter\def\csname result@leave_one_task_out/H3__drop_libero_goal_1_brms@low952\endcsname{-5.07}
\expandafter\def\csname result@leave_one_task_out/H3__drop_libero_goal_1_brms@high952\endcsname{1.87}
\expandafter\def\csname result@leave_one_task_out/H3__drop_libero_goal_6_brms@value1\endcsname{-2.8}
\expandafter\def\csname result@leave_one_task_out/H3__drop_libero_goal_6_brms@low1\endcsname{-6.2}
\expandafter\def\csname result@leave_one_task_out/H3__drop_libero_goal_6_brms@high1\endcsname{0.4}
\expandafter\def\csname result@leave_one_task_out/H3__drop_libero_goal_6_brms@low951\endcsname{-7.0}
\expandafter\def\csname result@leave_one_task_out/H3__drop_libero_goal_6_brms@high951\endcsname{1.0}
\expandafter\def\csname result@leave_one_task_out/H3__drop_libero_goal_6_brms@value2\endcsname{-2.78}
\expandafter\def\csname result@leave_one_task_out/H3__drop_libero_goal_6_brms@low2\endcsname{-6.23}
\expandafter\def\csname result@leave_one_task_out/H3__drop_libero_goal_6_brms@high2\endcsname{0.35}
\expandafter\def\csname result@leave_one_task_out/H3__drop_libero_goal_6_brms@low952\endcsname{-6.96}
\expandafter\def\csname result@leave_one_task_out/H3__drop_libero_goal_6_brms@high952\endcsname{0.95}
\expandafter\def\csname result@leave_one_task_out/H3__drop_libero_goal_7_brms@value1\endcsname{-1.5}
\expandafter\def\csname result@leave_one_task_out/H3__drop_libero_goal_7_brms@low1\endcsname{-4.9}
\expandafter\def\csname result@leave_one_task_out/H3__drop_libero_goal_7_brms@high1\endcsname{1.7}
\expandafter\def\csname result@leave_one_task_out/H3__drop_libero_goal_7_brms@low951\endcsname{-5.6}
\expandafter\def\csname result@leave_one_task_out/H3__drop_libero_goal_7_brms@high951\endcsname{2.3}
\expandafter\def\csname result@leave_one_task_out/H3__drop_libero_goal_7_brms@value2\endcsname{-1.53}
\expandafter\def\csname result@leave_one_task_out/H3__drop_libero_goal_7_brms@low2\endcsname{-4.85}
\expandafter\def\csname result@leave_one_task_out/H3__drop_libero_goal_7_brms@high2\endcsname{1.66}
\expandafter\def\csname result@leave_one_task_out/H3__drop_libero_goal_7_brms@low952\endcsname{-5.59}
\expandafter\def\csname result@leave_one_task_out/H3__drop_libero_goal_7_brms@high952\endcsname{2.34}
\expandafter\def\csname result@leave_one_task_out/H3__drop_libero_spatial_2_brms@value1\endcsname{-1.8}
\expandafter\def\csname result@leave_one_task_out/H3__drop_libero_spatial_2_brms@low1\endcsname{-5.1}
\expandafter\def\csname result@leave_one_task_out/H3__drop_libero_spatial_2_brms@high1\endcsname{1.4}
\expandafter\def\csname result@leave_one_task_out/H3__drop_libero_spatial_2_brms@low951\endcsname{-5.9}
\expandafter\def\csname result@leave_one_task_out/H3__drop_libero_spatial_2_brms@high951\endcsname{2.0}
\expandafter\def\csname result@leave_one_task_out/H3__drop_libero_spatial_2_brms@value2\endcsname{-1.80}
\expandafter\def\csname result@leave_one_task_out/H3__drop_libero_spatial_2_brms@low2\endcsname{-5.11}
\expandafter\def\csname result@leave_one_task_out/H3__drop_libero_spatial_2_brms@high2\endcsname{1.36}
\expandafter\def\csname result@leave_one_task_out/H3__drop_libero_spatial_2_brms@low952\endcsname{-5.86}
\expandafter\def\csname result@leave_one_task_out/H3__drop_libero_spatial_2_brms@high952\endcsname{2.00}
\expandafter\def\csname result@leave_one_task_out/H3__drop_libero_spatial_8_brms@value1\endcsname{-2.7}
\expandafter\def\csname result@leave_one_task_out/H3__drop_libero_spatial_8_brms@low1\endcsname{-6.2}
\expandafter\def\csname result@leave_one_task_out/H3__drop_libero_spatial_8_brms@high1\endcsname{0.5}
\expandafter\def\csname result@leave_one_task_out/H3__drop_libero_spatial_8_brms@low951\endcsname{-7.0}
\expandafter\def\csname result@leave_one_task_out/H3__drop_libero_spatial_8_brms@high951\endcsname{1.1}
\expandafter\def\csname result@leave_one_task_out/H3__drop_libero_spatial_8_brms@value2\endcsname{-2.69}
\expandafter\def\csname result@leave_one_task_out/H3__drop_libero_spatial_8_brms@low2\endcsname{-6.19}
\expandafter\def\csname result@leave_one_task_out/H3__drop_libero_spatial_8_brms@high2\endcsname{0.52}
\expandafter\def\csname result@leave_one_task_out/H3__drop_libero_spatial_8_brms@low952\endcsname{-7.02}
\expandafter\def\csname result@leave_one_task_out/H3__drop_libero_spatial_8_brms@high952\endcsname{1.14}
\expandafter\def\csname result@leave_one_task_out/H4__drop_libero_10_5_brms@value1\endcsname{-1.4}
\expandafter\def\csname result@leave_one_task_out/H4__drop_libero_10_5_brms@low1\endcsname{-5.3}
\expandafter\def\csname result@leave_one_task_out/H4__drop_libero_10_5_brms@high1\endcsname{2.6}
\expandafter\def\csname result@leave_one_task_out/H4__drop_libero_10_5_brms@low951\endcsname{-6.0}
\expandafter\def\csname result@leave_one_task_out/H4__drop_libero_10_5_brms@high951\endcsname{3.4}
\expandafter\def\csname result@leave_one_task_out/H4__drop_libero_10_5_brms@value2\endcsname{-1.35}
\expandafter\def\csname result@leave_one_task_out/H4__drop_libero_10_5_brms@low2\endcsname{-5.27}
\expandafter\def\csname result@leave_one_task_out/H4__drop_libero_10_5_brms@high2\endcsname{2.64}
\expandafter\def\csname result@leave_one_task_out/H4__drop_libero_10_5_brms@low952\endcsname{-6.03}
\expandafter\def\csname result@leave_one_task_out/H4__drop_libero_10_5_brms@high952\endcsname{3.38}
\expandafter\def\csname result@leave_one_task_out/H4__drop_libero_10_9_brms@value1\endcsname{0.8}
\expandafter\def\csname result@leave_one_task_out/H4__drop_libero_10_9_brms@low1\endcsname{-2.9}
\expandafter\def\csname result@leave_one_task_out/H4__drop_libero_10_9_brms@high1\endcsname{4.8}
\expandafter\def\csname result@leave_one_task_out/H4__drop_libero_10_9_brms@low951\endcsname{-3.7}
\expandafter\def\csname result@leave_one_task_out/H4__drop_libero_10_9_brms@high951\endcsname{5.5}
\expandafter\def\csname result@leave_one_task_out/H4__drop_libero_10_9_brms@value2\endcsname{0.84}
\expandafter\def\csname result@leave_one_task_out/H4__drop_libero_10_9_brms@low2\endcsname{-2.94}
\expandafter\def\csname result@leave_one_task_out/H4__drop_libero_10_9_brms@high2\endcsname{4.76}
\expandafter\def\csname result@leave_one_task_out/H4__drop_libero_10_9_brms@low952\endcsname{-3.69}
\expandafter\def\csname result@leave_one_task_out/H4__drop_libero_10_9_brms@high952\endcsname{5.46}
\expandafter\def\csname result@leave_one_task_out/H4__drop_libero_goal_1_brms@value1\endcsname{-1.0}
\expandafter\def\csname result@leave_one_task_out/H4__drop_libero_goal_1_brms@low1\endcsname{-4.9}
\expandafter\def\csname result@leave_one_task_out/H4__drop_libero_goal_1_brms@high1\endcsname{3.0}
\expandafter\def\csname result@leave_one_task_out/H4__drop_libero_goal_1_brms@low951\endcsname{-5.7}
\expandafter\def\csname result@leave_one_task_out/H4__drop_libero_goal_1_brms@high951\endcsname{3.8}
\expandafter\def\csname result@leave_one_task_out/H4__drop_libero_goal_1_brms@value2\endcsname{-0.99}
\expandafter\def\csname result@leave_one_task_out/H4__drop_libero_goal_1_brms@low2\endcsname{-4.95}
\expandafter\def\csname result@leave_one_task_out/H4__drop_libero_goal_1_brms@high2\endcsname{3.00}
\expandafter\def\csname result@leave_one_task_out/H4__drop_libero_goal_1_brms@low952\endcsname{-5.71}
\expandafter\def\csname result@leave_one_task_out/H4__drop_libero_goal_1_brms@high952\endcsname{3.78}
\expandafter\def\csname result@leave_one_task_out/H4__drop_libero_goal_6_brms@value1\endcsname{-0.1}
\expandafter\def\csname result@leave_one_task_out/H4__drop_libero_goal_6_brms@low1\endcsname{-3.9}
\expandafter\def\csname result@leave_one_task_out/H4__drop_libero_goal_6_brms@high1\endcsname{3.6}
\expandafter\def\csname result@leave_one_task_out/H4__drop_libero_goal_6_brms@low951\endcsname{-4.7}
\expandafter\def\csname result@leave_one_task_out/H4__drop_libero_goal_6_brms@high951\endcsname{4.3}
\expandafter\def\csname result@leave_one_task_out/H4__drop_libero_goal_6_brms@value2\endcsname{-0.14}
\expandafter\def\csname result@leave_one_task_out/H4__drop_libero_goal_6_brms@low2\endcsname{-3.94}
\expandafter\def\csname result@leave_one_task_out/H4__drop_libero_goal_6_brms@high2\endcsname{3.55}
\expandafter\def\csname result@leave_one_task_out/H4__drop_libero_goal_6_brms@low952\endcsname{-4.66}
\expandafter\def\csname result@leave_one_task_out/H4__drop_libero_goal_6_brms@high952\endcsname{4.32}
\expandafter\def\csname result@leave_one_task_out/H4__drop_libero_goal_7_brms@value1\endcsname{-0.5}
\expandafter\def\csname result@leave_one_task_out/H4__drop_libero_goal_7_brms@low1\endcsname{-4.4}
\expandafter\def\csname result@leave_one_task_out/H4__drop_libero_goal_7_brms@high1\endcsname{3.5}
\expandafter\def\csname result@leave_one_task_out/H4__drop_libero_goal_7_brms@low951\endcsname{-5.2}
\expandafter\def\csname result@leave_one_task_out/H4__drop_libero_goal_7_brms@high951\endcsname{4.2}
\expandafter\def\csname result@leave_one_task_out/H4__drop_libero_goal_7_brms@value2\endcsname{-0.47}
\expandafter\def\csname result@leave_one_task_out/H4__drop_libero_goal_7_brms@low2\endcsname{-4.45}
\expandafter\def\csname result@leave_one_task_out/H4__drop_libero_goal_7_brms@high2\endcsname{3.51}
\expandafter\def\csname result@leave_one_task_out/H4__drop_libero_goal_7_brms@low952\endcsname{-5.21}
\expandafter\def\csname result@leave_one_task_out/H4__drop_libero_goal_7_brms@high952\endcsname{4.23}
\expandafter\def\csname result@leave_one_task_out/H4__drop_libero_spatial_2_brms@value1\endcsname{-3.1}
\expandafter\def\csname result@leave_one_task_out/H4__drop_libero_spatial_2_brms@low1\endcsname{-7.2}
\expandafter\def\csname result@leave_one_task_out/H4__drop_libero_spatial_2_brms@high1\endcsname{0.9}
\expandafter\def\csname result@leave_one_task_out/H4__drop_libero_spatial_2_brms@low951\endcsname{-8.0}
\expandafter\def\csname result@leave_one_task_out/H4__drop_libero_spatial_2_brms@high951\endcsname{1.6}
\expandafter\def\csname result@leave_one_task_out/H4__drop_libero_spatial_2_brms@value2\endcsname{-3.09}
\expandafter\def\csname result@leave_one_task_out/H4__drop_libero_spatial_2_brms@low2\endcsname{-7.16}
\expandafter\def\csname result@leave_one_task_out/H4__drop_libero_spatial_2_brms@high2\endcsname{0.88}
\expandafter\def\csname result@leave_one_task_out/H4__drop_libero_spatial_2_brms@low952\endcsname{-8.00}
\expandafter\def\csname result@leave_one_task_out/H4__drop_libero_spatial_2_brms@high952\endcsname{1.63}
\expandafter\def\csname result@leave_one_task_out/H4__drop_libero_spatial_8_brms@value1\endcsname{-1.7}
\expandafter\def\csname result@leave_one_task_out/H4__drop_libero_spatial_8_brms@low1\endcsname{-5.4}
\expandafter\def\csname result@leave_one_task_out/H4__drop_libero_spatial_8_brms@high1\endcsname{1.9}
\expandafter\def\csname result@leave_one_task_out/H4__drop_libero_spatial_8_brms@low951\endcsname{-6.2}
\expandafter\def\csname result@leave_one_task_out/H4__drop_libero_spatial_8_brms@high951\endcsname{2.7}
\expandafter\def\csname result@leave_one_task_out/H4__drop_libero_spatial_8_brms@value2\endcsname{-1.69}
\expandafter\def\csname result@leave_one_task_out/H4__drop_libero_spatial_8_brms@low2\endcsname{-5.40}
\expandafter\def\csname result@leave_one_task_out/H4__drop_libero_spatial_8_brms@high2\endcsname{1.93}
\expandafter\def\csname result@leave_one_task_out/H4__drop_libero_spatial_8_brms@low952\endcsname{-6.15}
\expandafter\def\csname result@leave_one_task_out/H4__drop_libero_spatial_8_brms@high952\endcsname{2.67}
\expandafter\def\csname result@leave_one_task_out/H5__drop_libero_10_5_brms@value1\endcsname{-3.7}
\expandafter\def\csname result@leave_one_task_out/H5__drop_libero_10_5_brms@low1\endcsname{-7.0}
\expandafter\def\csname result@leave_one_task_out/H5__drop_libero_10_5_brms@high1\endcsname{-0.9}
\expandafter\def\csname result@leave_one_task_out/H5__drop_libero_10_5_brms@low951\endcsname{-7.8}
\expandafter\def\csname result@leave_one_task_out/H5__drop_libero_10_5_brms@high951\endcsname{-0.5}
\expandafter\def\csname result@leave_one_task_out/H5__drop_libero_10_5_brms@value2\endcsname{-3.69}
\expandafter\def\csname result@leave_one_task_out/H5__drop_libero_10_5_brms@low2\endcsname{-6.99}
\expandafter\def\csname result@leave_one_task_out/H5__drop_libero_10_5_brms@high2\endcsname{-0.89}
\expandafter\def\csname result@leave_one_task_out/H5__drop_libero_10_5_brms@low952\endcsname{-7.80}
\expandafter\def\csname result@leave_one_task_out/H5__drop_libero_10_5_brms@high952\endcsname{-0.48}
\expandafter\def\csname result@leave_one_task_out/H5__drop_libero_10_9_brms@value1\endcsname{-0.8}
\expandafter\def\csname result@leave_one_task_out/H5__drop_libero_10_9_brms@low1\endcsname{-3.7}
\expandafter\def\csname result@leave_one_task_out/H5__drop_libero_10_9_brms@high1\endcsname{1.9}
\expandafter\def\csname result@leave_one_task_out/H5__drop_libero_10_9_brms@low951\endcsname{-4.4}
\expandafter\def\csname result@leave_one_task_out/H5__drop_libero_10_9_brms@high951\endcsname{2.4}
\expandafter\def\csname result@leave_one_task_out/H5__drop_libero_10_9_brms@value2\endcsname{-0.85}
\expandafter\def\csname result@leave_one_task_out/H5__drop_libero_10_9_brms@low2\endcsname{-3.72}
\expandafter\def\csname result@leave_one_task_out/H5__drop_libero_10_9_brms@high2\endcsname{1.89}
\expandafter\def\csname result@leave_one_task_out/H5__drop_libero_10_9_brms@low952\endcsname{-4.39}
\expandafter\def\csname result@leave_one_task_out/H5__drop_libero_10_9_brms@high952\endcsname{2.45}
\expandafter\def\csname result@leave_one_task_out/H5__drop_libero_goal_1_brms@value1\endcsname{-0.7}
\expandafter\def\csname result@leave_one_task_out/H5__drop_libero_goal_1_brms@low1\endcsname{-3.4}
\expandafter\def\csname result@leave_one_task_out/H5__drop_libero_goal_1_brms@high1\endcsname{1.8}
\expandafter\def\csname result@leave_one_task_out/H5__drop_libero_goal_1_brms@low951\endcsname{-4.1}
\expandafter\def\csname result@leave_one_task_out/H5__drop_libero_goal_1_brms@high951\endcsname{2.3}
\expandafter\def\csname result@leave_one_task_out/H5__drop_libero_goal_1_brms@value2\endcsname{-0.73}
\expandafter\def\csname result@leave_one_task_out/H5__drop_libero_goal_1_brms@low2\endcsname{-3.45}
\expandafter\def\csname result@leave_one_task_out/H5__drop_libero_goal_1_brms@high2\endcsname{1.80}
\expandafter\def\csname result@leave_one_task_out/H5__drop_libero_goal_1_brms@low952\endcsname{-4.06}
\expandafter\def\csname result@leave_one_task_out/H5__drop_libero_goal_1_brms@high952\endcsname{2.34}
\expandafter\def\csname result@leave_one_task_out/H5__drop_libero_goal_6_brms@value1\endcsname{-2.3}
\expandafter\def\csname result@leave_one_task_out/H5__drop_libero_goal_6_brms@low1\endcsname{-5.3}
\expandafter\def\csname result@leave_one_task_out/H5__drop_libero_goal_6_brms@high1\endcsname{0.4}
\expandafter\def\csname result@leave_one_task_out/H5__drop_libero_goal_6_brms@low951\endcsname{-6.0}
\expandafter\def\csname result@leave_one_task_out/H5__drop_libero_goal_6_brms@high951\endcsname{0.9}
\expandafter\def\csname result@leave_one_task_out/H5__drop_libero_goal_6_brms@value2\endcsname{-2.33}
\expandafter\def\csname result@leave_one_task_out/H5__drop_libero_goal_6_brms@low2\endcsname{-5.31}
\expandafter\def\csname result@leave_one_task_out/H5__drop_libero_goal_6_brms@high2\endcsname{0.39}
\expandafter\def\csname result@leave_one_task_out/H5__drop_libero_goal_6_brms@low952\endcsname{-6.03}
\expandafter\def\csname result@leave_one_task_out/H5__drop_libero_goal_6_brms@high952\endcsname{0.88}
\expandafter\def\csname result@leave_one_task_out/H5__drop_libero_goal_7_brms@value1\endcsname{-1.3}
\expandafter\def\csname result@leave_one_task_out/H5__drop_libero_goal_7_brms@low1\endcsname{-4.2}
\expandafter\def\csname result@leave_one_task_out/H5__drop_libero_goal_7_brms@high1\endcsname{1.4}
\expandafter\def\csname result@leave_one_task_out/H5__drop_libero_goal_7_brms@low951\endcsname{-5.0}
\expandafter\def\csname result@leave_one_task_out/H5__drop_libero_goal_7_brms@high951\endcsname{1.9}
\expandafter\def\csname result@leave_one_task_out/H5__drop_libero_goal_7_brms@value2\endcsname{-1.30}
\expandafter\def\csname result@leave_one_task_out/H5__drop_libero_goal_7_brms@low2\endcsname{-4.24}
\expandafter\def\csname result@leave_one_task_out/H5__drop_libero_goal_7_brms@high2\endcsname{1.44}
\expandafter\def\csname result@leave_one_task_out/H5__drop_libero_goal_7_brms@low952\endcsname{-4.96}
\expandafter\def\csname result@leave_one_task_out/H5__drop_libero_goal_7_brms@high952\endcsname{1.94}
\expandafter\def\csname result@leave_one_task_out/H5__drop_libero_spatial_2_brms@value1\endcsname{-1.9}
\expandafter\def\csname result@leave_one_task_out/H5__drop_libero_spatial_2_brms@low1\endcsname{-5.0}
\expandafter\def\csname result@leave_one_task_out/H5__drop_libero_spatial_2_brms@high1\endcsname{1.1}
\expandafter\def\csname result@leave_one_task_out/H5__drop_libero_spatial_2_brms@low951\endcsname{-5.7}
\expandafter\def\csname result@leave_one_task_out/H5__drop_libero_spatial_2_brms@high951\endcsname{1.7}
\expandafter\def\csname result@leave_one_task_out/H5__drop_libero_spatial_2_brms@value2\endcsname{-1.88}
\expandafter\def\csname result@leave_one_task_out/H5__drop_libero_spatial_2_brms@low2\endcsname{-5.03}
\expandafter\def\csname result@leave_one_task_out/H5__drop_libero_spatial_2_brms@high2\endcsname{1.05}
\expandafter\def\csname result@leave_one_task_out/H5__drop_libero_spatial_2_brms@low952\endcsname{-5.70}
\expandafter\def\csname result@leave_one_task_out/H5__drop_libero_spatial_2_brms@high952\endcsname{1.66}
\expandafter\def\csname result@leave_one_task_out/H5__drop_libero_spatial_8_brms@value1\endcsname{-2.1}
\expandafter\def\csname result@leave_one_task_out/H5__drop_libero_spatial_8_brms@low1\endcsname{-5.3}
\expandafter\def\csname result@leave_one_task_out/H5__drop_libero_spatial_8_brms@high1\endcsname{0.8}
\expandafter\def\csname result@leave_one_task_out/H5__drop_libero_spatial_8_brms@low951\endcsname{-6.0}
\expandafter\def\csname result@leave_one_task_out/H5__drop_libero_spatial_8_brms@high951\endcsname{1.3}
\expandafter\def\csname result@leave_one_task_out/H5__drop_libero_spatial_8_brms@value2\endcsname{-2.14}
\expandafter\def\csname result@leave_one_task_out/H5__drop_libero_spatial_8_brms@low2\endcsname{-5.28}
\expandafter\def\csname result@leave_one_task_out/H5__drop_libero_spatial_8_brms@high2\endcsname{0.77}
\expandafter\def\csname result@leave_one_task_out/H5__drop_libero_spatial_8_brms@low952\endcsname{-5.98}
\expandafter\def\csname result@leave_one_task_out/H5__drop_libero_spatial_8_brms@high952\endcsname{1.31}
\expandafter\def\csname result@leave_one_task_out/H6b__drop_libero_10_5_brms@value1\endcsname{1.7}
\expandafter\def\csname result@leave_one_task_out/H6b__drop_libero_10_5_brms@low1\endcsname{-1.1}
\expandafter\def\csname result@leave_one_task_out/H6b__drop_libero_10_5_brms@high1\endcsname{4.9}
\expandafter\def\csname result@leave_one_task_out/H6b__drop_libero_10_5_brms@low951\endcsname{-1.7}
\expandafter\def\csname result@leave_one_task_out/H6b__drop_libero_10_5_brms@high951\endcsname{5.7}
\expandafter\def\csname result@leave_one_task_out/H6b__drop_libero_10_5_brms@value2\endcsname{1.74}
\expandafter\def\csname result@leave_one_task_out/H6b__drop_libero_10_5_brms@low2\endcsname{-1.12}
\expandafter\def\csname result@leave_one_task_out/H6b__drop_libero_10_5_brms@high2\endcsname{4.93}
\expandafter\def\csname result@leave_one_task_out/H6b__drop_libero_10_5_brms@low952\endcsname{-1.72}
\expandafter\def\csname result@leave_one_task_out/H6b__drop_libero_10_5_brms@high952\endcsname{5.67}
\expandafter\def\csname result@leave_one_task_out/H6b__drop_libero_10_9_brms@value1\endcsname{2.7}
\expandafter\def\csname result@leave_one_task_out/H6b__drop_libero_10_9_brms@low1\endcsname{-0.3}
\expandafter\def\csname result@leave_one_task_out/H6b__drop_libero_10_9_brms@high1\endcsname{6.1}
\expandafter\def\csname result@leave_one_task_out/H6b__drop_libero_10_9_brms@low951\endcsname{-0.8}
\expandafter\def\csname result@leave_one_task_out/H6b__drop_libero_10_9_brms@high951\endcsname{6.9}
\expandafter\def\csname result@leave_one_task_out/H6b__drop_libero_10_9_brms@value2\endcsname{2.65}
\expandafter\def\csname result@leave_one_task_out/H6b__drop_libero_10_9_brms@low2\endcsname{-0.26}
\expandafter\def\csname result@leave_one_task_out/H6b__drop_libero_10_9_brms@high2\endcsname{6.14}
\expandafter\def\csname result@leave_one_task_out/H6b__drop_libero_10_9_brms@low952\endcsname{-0.81}
\expandafter\def\csname result@leave_one_task_out/H6b__drop_libero_10_9_brms@high952\endcsname{6.90}
\expandafter\def\csname result@leave_one_task_out/H6b__drop_libero_goal_1_brms@value1\endcsname{1.7}
\expandafter\def\csname result@leave_one_task_out/H6b__drop_libero_goal_1_brms@low1\endcsname{-1.1}
\expandafter\def\csname result@leave_one_task_out/H6b__drop_libero_goal_1_brms@high1\endcsname{5.0}
\expandafter\def\csname result@leave_one_task_out/H6b__drop_libero_goal_1_brms@low951\endcsname{-1.7}
\expandafter\def\csname result@leave_one_task_out/H6b__drop_libero_goal_1_brms@high951\endcsname{5.9}
\expandafter\def\csname result@leave_one_task_out/H6b__drop_libero_goal_1_brms@value2\endcsname{1.74}
\expandafter\def\csname result@leave_one_task_out/H6b__drop_libero_goal_1_brms@low2\endcsname{-1.12}
\expandafter\def\csname result@leave_one_task_out/H6b__drop_libero_goal_1_brms@high2\endcsname{5.05}
\expandafter\def\csname result@leave_one_task_out/H6b__drop_libero_goal_1_brms@low952\endcsname{-1.68}
\expandafter\def\csname result@leave_one_task_out/H6b__drop_libero_goal_1_brms@high952\endcsname{5.85}
\expandafter\def\csname result@leave_one_task_out/H6b__drop_libero_goal_6_brms@value1\endcsname{2.7}
\expandafter\def\csname result@leave_one_task_out/H6b__drop_libero_goal_6_brms@low1\endcsname{-0.1}
\expandafter\def\csname result@leave_one_task_out/H6b__drop_libero_goal_6_brms@high1\endcsname{6.0}
\expandafter\def\csname result@leave_one_task_out/H6b__drop_libero_goal_6_brms@low951\endcsname{-0.7}
\expandafter\def\csname result@leave_one_task_out/H6b__drop_libero_goal_6_brms@high951\endcsname{6.8}
\expandafter\def\csname result@leave_one_task_out/H6b__drop_libero_goal_6_brms@value2\endcsname{2.75}
\expandafter\def\csname result@leave_one_task_out/H6b__drop_libero_goal_6_brms@low2\endcsname{-0.13}
\expandafter\def\csname result@leave_one_task_out/H6b__drop_libero_goal_6_brms@high2\endcsname{6.05}
\expandafter\def\csname result@leave_one_task_out/H6b__drop_libero_goal_6_brms@low952\endcsname{-0.66}
\expandafter\def\csname result@leave_one_task_out/H6b__drop_libero_goal_6_brms@high952\endcsname{6.78}
\expandafter\def\csname result@leave_one_task_out/H6b__drop_libero_goal_7_brms@value1\endcsname{1.5}
\expandafter\def\csname result@leave_one_task_out/H6b__drop_libero_goal_7_brms@low1\endcsname{-1.5}
\expandafter\def\csname result@leave_one_task_out/H6b__drop_libero_goal_7_brms@high1\endcsname{4.8}
\expandafter\def\csname result@leave_one_task_out/H6b__drop_libero_goal_7_brms@low951\endcsname{-2.1}
\expandafter\def\csname result@leave_one_task_out/H6b__drop_libero_goal_7_brms@high951\endcsname{5.5}
\expandafter\def\csname result@leave_one_task_out/H6b__drop_libero_goal_7_brms@value2\endcsname{1.51}
\expandafter\def\csname result@leave_one_task_out/H6b__drop_libero_goal_7_brms@low2\endcsname{-1.49}
\expandafter\def\csname result@leave_one_task_out/H6b__drop_libero_goal_7_brms@high2\endcsname{4.78}
\expandafter\def\csname result@leave_one_task_out/H6b__drop_libero_goal_7_brms@low952\endcsname{-2.09}
\expandafter\def\csname result@leave_one_task_out/H6b__drop_libero_goal_7_brms@high952\endcsname{5.54}
\expandafter\def\csname result@leave_one_task_out/H6b__drop_libero_spatial_2_brms@value1\endcsname{2.3}
\expandafter\def\csname result@leave_one_task_out/H6b__drop_libero_spatial_2_brms@low1\endcsname{-0.7}
\expandafter\def\csname result@leave_one_task_out/H6b__drop_libero_spatial_2_brms@high1\endcsname{5.7}
\expandafter\def\csname result@leave_one_task_out/H6b__drop_libero_spatial_2_brms@low951\endcsname{-1.3}
\expandafter\def\csname result@leave_one_task_out/H6b__drop_libero_spatial_2_brms@high951\endcsname{6.5}
\expandafter\def\csname result@leave_one_task_out/H6b__drop_libero_spatial_2_brms@value2\endcsname{2.34}
\expandafter\def\csname result@leave_one_task_out/H6b__drop_libero_spatial_2_brms@low2\endcsname{-0.75}
\expandafter\def\csname result@leave_one_task_out/H6b__drop_libero_spatial_2_brms@high2\endcsname{5.74}
\expandafter\def\csname result@leave_one_task_out/H6b__drop_libero_spatial_2_brms@low952\endcsname{-1.30}
\expandafter\def\csname result@leave_one_task_out/H6b__drop_libero_spatial_2_brms@high952\endcsname{6.47}
\expandafter\def\csname result@leave_one_task_out/H6b__drop_libero_spatial_8_brms@value1\endcsname{2.7}
\expandafter\def\csname result@leave_one_task_out/H6b__drop_libero_spatial_8_brms@low1\endcsname{-0.6}
\expandafter\def\csname result@leave_one_task_out/H6b__drop_libero_spatial_8_brms@high1\endcsname{6.3}
\expandafter\def\csname result@leave_one_task_out/H6b__drop_libero_spatial_8_brms@low951\endcsname{-1.3}
\expandafter\def\csname result@leave_one_task_out/H6b__drop_libero_spatial_8_brms@high951\endcsname{7.0}
\expandafter\def\csname result@leave_one_task_out/H6b__drop_libero_spatial_8_brms@value2\endcsname{2.68}
\expandafter\def\csname result@leave_one_task_out/H6b__drop_libero_spatial_8_brms@low2\endcsname{-0.64}
\expandafter\def\csname result@leave_one_task_out/H6b__drop_libero_spatial_8_brms@high2\endcsname{6.28}
\expandafter\def\csname result@leave_one_task_out/H6b__drop_libero_spatial_8_brms@low952\endcsname{-1.27}
\expandafter\def\csname result@leave_one_task_out/H6b__drop_libero_spatial_8_brms@high952\endcsname{7.04}
\expandafter\def\csname result@leave_one_task_out/H6c__drop_libero_10_5_brms@value1\endcsname{5.4}
\expandafter\def\csname result@leave_one_task_out/H6c__drop_libero_10_5_brms@low1\endcsname{1.1}
\expandafter\def\csname result@leave_one_task_out/H6c__drop_libero_10_5_brms@high1\endcsname{10.5}
\expandafter\def\csname result@leave_one_task_out/H6c__drop_libero_10_5_brms@low951\endcsname{0.4}
\expandafter\def\csname result@leave_one_task_out/H6c__drop_libero_10_5_brms@high951\endcsname{11.7}
\expandafter\def\csname result@leave_one_task_out/H6c__drop_libero_10_5_brms@value2\endcsname{5.40}
\expandafter\def\csname result@leave_one_task_out/H6c__drop_libero_10_5_brms@low2\endcsname{1.12}
\expandafter\def\csname result@leave_one_task_out/H6c__drop_libero_10_5_brms@high2\endcsname{10.55}
\expandafter\def\csname result@leave_one_task_out/H6c__drop_libero_10_5_brms@low952\endcsname{0.43}
\expandafter\def\csname result@leave_one_task_out/H6c__drop_libero_10_5_brms@high952\endcsname{11.67}
\expandafter\def\csname result@leave_one_task_out/H6c__drop_libero_10_9_brms@value1\endcsname{3.6}
\expandafter\def\csname result@leave_one_task_out/H6c__drop_libero_10_9_brms@low1\endcsname{-0.5}
\expandafter\def\csname result@leave_one_task_out/H6c__drop_libero_10_9_brms@high1\endcsname{8.2}
\expandafter\def\csname result@leave_one_task_out/H6c__drop_libero_10_9_brms@low951\endcsname{-1.3}
\expandafter\def\csname result@leave_one_task_out/H6c__drop_libero_10_9_brms@high951\endcsname{9.3}
\expandafter\def\csname result@leave_one_task_out/H6c__drop_libero_10_9_brms@value2\endcsname{3.57}
\expandafter\def\csname result@leave_one_task_out/H6c__drop_libero_10_9_brms@low2\endcsname{-0.46}
\expandafter\def\csname result@leave_one_task_out/H6c__drop_libero_10_9_brms@high2\endcsname{8.18}
\expandafter\def\csname result@leave_one_task_out/H6c__drop_libero_10_9_brms@low952\endcsname{-1.27}
\expandafter\def\csname result@leave_one_task_out/H6c__drop_libero_10_9_brms@high952\endcsname{9.28}
\expandafter\def\csname result@leave_one_task_out/H6c__drop_libero_goal_1_brms@value1\endcsname{2.5}
\expandafter\def\csname result@leave_one_task_out/H6c__drop_libero_goal_1_brms@low1\endcsname{-1.3}
\expandafter\def\csname result@leave_one_task_out/H6c__drop_libero_goal_1_brms@high1\endcsname{6.6}
\expandafter\def\csname result@leave_one_task_out/H6c__drop_libero_goal_1_brms@low951\endcsname{-2.1}
\expandafter\def\csname result@leave_one_task_out/H6c__drop_libero_goal_1_brms@high951\endcsname{7.6}
\expandafter\def\csname result@leave_one_task_out/H6c__drop_libero_goal_1_brms@value2\endcsname{2.46}
\expandafter\def\csname result@leave_one_task_out/H6c__drop_libero_goal_1_brms@low2\endcsname{-1.33}
\expandafter\def\csname result@leave_one_task_out/H6c__drop_libero_goal_1_brms@high2\endcsname{6.61}
\expandafter\def\csname result@leave_one_task_out/H6c__drop_libero_goal_1_brms@low952\endcsname{-2.09}
\expandafter\def\csname result@leave_one_task_out/H6c__drop_libero_goal_1_brms@high952\endcsname{7.63}
\expandafter\def\csname result@leave_one_task_out/H6c__drop_libero_goal_6_brms@value1\endcsname{5.2}
\expandafter\def\csname result@leave_one_task_out/H6c__drop_libero_goal_6_brms@low1\endcsname{1.1}
\expandafter\def\csname result@leave_one_task_out/H6c__drop_libero_goal_6_brms@high1\endcsname{10.1}
\expandafter\def\csname result@leave_one_task_out/H6c__drop_libero_goal_6_brms@low951\endcsname{0.4}
\expandafter\def\csname result@leave_one_task_out/H6c__drop_libero_goal_6_brms@high951\endcsname{11.1}
\expandafter\def\csname result@leave_one_task_out/H6c__drop_libero_goal_6_brms@value2\endcsname{5.16}
\expandafter\def\csname result@leave_one_task_out/H6c__drop_libero_goal_6_brms@low2\endcsname{1.06}
\expandafter\def\csname result@leave_one_task_out/H6c__drop_libero_goal_6_brms@high2\endcsname{10.06}
\expandafter\def\csname result@leave_one_task_out/H6c__drop_libero_goal_6_brms@low952\endcsname{0.39}
\expandafter\def\csname result@leave_one_task_out/H6c__drop_libero_goal_6_brms@high952\endcsname{11.13}
\expandafter\def\csname result@leave_one_task_out/H6c__drop_libero_goal_7_brms@value1\endcsname{2.8}
\expandafter\def\csname result@leave_one_task_out/H6c__drop_libero_goal_7_brms@low1\endcsname{-1.2}
\expandafter\def\csname result@leave_one_task_out/H6c__drop_libero_goal_7_brms@high1\endcsname{7.3}
\expandafter\def\csname result@leave_one_task_out/H6c__drop_libero_goal_7_brms@low951\endcsname{-2.0}
\expandafter\def\csname result@leave_one_task_out/H6c__drop_libero_goal_7_brms@high951\endcsname{8.3}
\expandafter\def\csname result@leave_one_task_out/H6c__drop_libero_goal_7_brms@value2\endcsname{2.76}
\expandafter\def\csname result@leave_one_task_out/H6c__drop_libero_goal_7_brms@low2\endcsname{-1.24}
\expandafter\def\csname result@leave_one_task_out/H6c__drop_libero_goal_7_brms@high2\endcsname{7.30}
\expandafter\def\csname result@leave_one_task_out/H6c__drop_libero_goal_7_brms@low952\endcsname{-2.01}
\expandafter\def\csname result@leave_one_task_out/H6c__drop_libero_goal_7_brms@high952\endcsname{8.29}
\expandafter\def\csname result@leave_one_task_out/H6c__drop_libero_spatial_2_brms@value1\endcsname{4.3}
\expandafter\def\csname result@leave_one_task_out/H6c__drop_libero_spatial_2_brms@low1\endcsname{-0.0}
\expandafter\def\csname result@leave_one_task_out/H6c__drop_libero_spatial_2_brms@high1\endcsname{9.2}
\expandafter\def\csname result@leave_one_task_out/H6c__drop_libero_spatial_2_brms@low951\endcsname{-0.9}
\expandafter\def\csname result@leave_one_task_out/H6c__drop_libero_spatial_2_brms@high951\endcsname{10.3}
\expandafter\def\csname result@leave_one_task_out/H6c__drop_libero_spatial_2_brms@value2\endcsname{4.31}
\expandafter\def\csname result@leave_one_task_out/H6c__drop_libero_spatial_2_brms@low2\endcsname{-0.04}
\expandafter\def\csname result@leave_one_task_out/H6c__drop_libero_spatial_2_brms@high2\endcsname{9.20}
\expandafter\def\csname result@leave_one_task_out/H6c__drop_libero_spatial_2_brms@low952\endcsname{-0.87}
\expandafter\def\csname result@leave_one_task_out/H6c__drop_libero_spatial_2_brms@high952\endcsname{10.31}
\expandafter\def\csname result@leave_one_task_out/H6c__drop_libero_spatial_8_brms@value1\endcsname{4.8}
\expandafter\def\csname result@leave_one_task_out/H6c__drop_libero_spatial_8_brms@low1\endcsname{0.4}
\expandafter\def\csname result@leave_one_task_out/H6c__drop_libero_spatial_8_brms@high1\endcsname{9.7}
\expandafter\def\csname result@leave_one_task_out/H6c__drop_libero_spatial_8_brms@low951\endcsname{-0.4}
\expandafter\def\csname result@leave_one_task_out/H6c__drop_libero_spatial_8_brms@high951\endcsname{10.8}
\expandafter\def\csname result@leave_one_task_out/H6c__drop_libero_spatial_8_brms@value2\endcsname{4.75}
\expandafter\def\csname result@leave_one_task_out/H6c__drop_libero_spatial_8_brms@low2\endcsname{0.39}
\expandafter\def\csname result@leave_one_task_out/H6c__drop_libero_spatial_8_brms@high2\endcsname{9.72}
\expandafter\def\csname result@leave_one_task_out/H6c__drop_libero_spatial_8_brms@low952\endcsname{-0.37}
\expandafter\def\csname result@leave_one_task_out/H6c__drop_libero_spatial_8_brms@high952\endcsname{10.79}
\expandafter\def\csname result@main/H0a_brms@value1\endcsname{5.5}
\expandafter\def\csname result@main/H0a_brms@low1\endcsname{2.3}
\expandafter\def\csname result@main/H0a_brms@high1\endcsname{9.0}
\expandafter\def\csname result@main/H0a_brms@low951\endcsname{1.8}
\expandafter\def\csname result@main/H0a_brms@high951\endcsname{9.8}
\expandafter\def\csname result@main/H0a_brms@value2\endcsname{5.50}
\expandafter\def\csname result@main/H0a_brms@low2\endcsname{2.28}
\expandafter\def\csname result@main/H0a_brms@high2\endcsname{9.03}
\expandafter\def\csname result@main/H0a_brms@low952\endcsname{1.75}
\expandafter\def\csname result@main/H0a_brms@high952\endcsname{9.84}
\expandafter\def\csname result@main/H0a_glmer@value1\endcsname{5.6}
\expandafter\def\csname result@main/H0a_glmer@low1\endcsname{2.3}
\expandafter\def\csname result@main/H0a_glmer@high1\endcsname{9.3}
\expandafter\def\csname result@main/H0a_glmer@low951\endcsname{1.7}
\expandafter\def\csname result@main/H0a_glmer@high951\endcsname{10.1}
\expandafter\def\csname result@main/H0a_glmer@value2\endcsname{5.64}
\expandafter\def\csname result@main/H0a_glmer@low2\endcsname{2.27}
\expandafter\def\csname result@main/H0a_glmer@high2\endcsname{9.25}
\expandafter\def\csname result@main/H0a_glmer@low952\endcsname{1.72}
\expandafter\def\csname result@main/H0a_glmer@high952\endcsname{10.07}
\expandafter\def\csname result@main/H0b_brms@value1\endcsname{9.1}
\expandafter\def\csname result@main/H0b_brms@low1\endcsname{5.3}
\expandafter\def\csname result@main/H0b_brms@high1\endcsname{13.4}
\expandafter\def\csname result@main/H0b_brms@low951\endcsname{4.8}
\expandafter\def\csname result@main/H0b_brms@high951\endcsname{14.3}
\expandafter\def\csname result@main/H0b_brms@value2\endcsname{9.07}
\expandafter\def\csname result@main/H0b_brms@low2\endcsname{5.29}
\expandafter\def\csname result@main/H0b_brms@high2\endcsname{13.36}
\expandafter\def\csname result@main/H0b_brms@low952\endcsname{4.76}
\expandafter\def\csname result@main/H0b_brms@high952\endcsname{14.28}
\expandafter\def\csname result@main/H0b_glmer@value1\endcsname{9.2}
\expandafter\def\csname result@main/H0b_glmer@low1\endcsname{5.2}
\expandafter\def\csname result@main/H0b_glmer@high1\endcsname{13.2}
\expandafter\def\csname result@main/H0b_glmer@low951\endcsname{4.5}
\expandafter\def\csname result@main/H0b_glmer@high951\endcsname{14.2}
\expandafter\def\csname result@main/H0b_glmer@value2\endcsname{9.19}
\expandafter\def\csname result@main/H0b_glmer@low2\endcsname{5.17}
\expandafter\def\csname result@main/H0b_glmer@high2\endcsname{13.23}
\expandafter\def\csname result@main/H0b_glmer@low952\endcsname{4.48}
\expandafter\def\csname result@main/H0b_glmer@high952\endcsname{14.17}
\expandafter\def\csname result@main/H0c_brms@value1\endcsname{8.5}
\expandafter\def\csname result@main/H0c_brms@low1\endcsname{4.6}
\expandafter\def\csname result@main/H0c_brms@high1\endcsname{12.9}
\expandafter\def\csname result@main/H0c_brms@low951\endcsname{4.0}
\expandafter\def\csname result@main/H0c_brms@high951\endcsname{13.9}
\expandafter\def\csname result@main/H0c_brms@value2\endcsname{8.50}
\expandafter\def\csname result@main/H0c_brms@low2\endcsname{4.60}
\expandafter\def\csname result@main/H0c_brms@high2\endcsname{12.93}
\expandafter\def\csname result@main/H0c_brms@low952\endcsname{4.00}
\expandafter\def\csname result@main/H0c_brms@high952\endcsname{13.91}
\expandafter\def\csname result@main/H0c_glmer@value1\endcsname{8.9}
\expandafter\def\csname result@main/H0c_glmer@low1\endcsname{4.4}
\expandafter\def\csname result@main/H0c_glmer@high1\endcsname{13.2}
\expandafter\def\csname result@main/H0c_glmer@low951\endcsname{3.9}
\expandafter\def\csname result@main/H0c_glmer@high951\endcsname{13.9}
\expandafter\def\csname result@main/H0c_glmer@value2\endcsname{8.88}
\expandafter\def\csname result@main/H0c_glmer@low2\endcsname{4.42}
\expandafter\def\csname result@main/H0c_glmer@high2\endcsname{13.16}
\expandafter\def\csname result@main/H0c_glmer@low952\endcsname{3.93}
\expandafter\def\csname result@main/H0c_glmer@high952\endcsname{13.87}
\expandafter\def\csname result@main/H1_brms@value1\endcsname{-0.6}
\expandafter\def\csname result@main/H1_brms@low1\endcsname{-3.4}
\expandafter\def\csname result@main/H1_brms@high1\endcsname{2.1}
\expandafter\def\csname result@main/H1_brms@low951\endcsname{-4.0}
\expandafter\def\csname result@main/H1_brms@high951\endcsname{2.7}
\expandafter\def\csname result@main/H1_brms@value2\endcsname{-0.64}
\expandafter\def\csname result@main/H1_brms@low2\endcsname{-3.43}
\expandafter\def\csname result@main/H1_brms@high2\endcsname{2.09}
\expandafter\def\csname result@main/H1_brms@low952\endcsname{-4.00}
\expandafter\def\csname result@main/H1_brms@high952\endcsname{2.66}
\expandafter\def\csname result@main/H1_glmer@value1\endcsname{-0.6}
\expandafter\def\csname result@main/H1_glmer@low1\endcsname{-3.2}
\expandafter\def\csname result@main/H1_glmer@high1\endcsname{2.0}
\expandafter\def\csname result@main/H1_glmer@low951\endcsname{-3.7}
\expandafter\def\csname result@main/H1_glmer@high951\endcsname{2.5}
\expandafter\def\csname result@main/H1_glmer@value2\endcsname{-0.64}
\expandafter\def\csname result@main/H1_glmer@low2\endcsname{-3.17}
\expandafter\def\csname result@main/H1_glmer@high2\endcsname{1.98}
\expandafter\def\csname result@main/H1_glmer@low952\endcsname{-3.71}
\expandafter\def\csname result@main/H1_glmer@high952\endcsname{2.47}
\expandafter\def\csname result@main/H2_brms@value1\endcsname{0.0}
\expandafter\def\csname result@main/H2_brms@low1\endcsname{-2.5}
\expandafter\def\csname result@main/H2_brms@high1\endcsname{2.5}
\expandafter\def\csname result@main/H2_brms@low951\endcsname{-3.1}
\expandafter\def\csname result@main/H2_brms@high951\endcsname{3.0}
\expandafter\def\csname result@main/H2_brms@value2\endcsname{0.00}
\expandafter\def\csname result@main/H2_brms@low2\endcsname{-2.53}
\expandafter\def\csname result@main/H2_brms@high2\endcsname{2.55}
\expandafter\def\csname result@main/H2_brms@low952\endcsname{-3.10}
\expandafter\def\csname result@main/H2_brms@high952\endcsname{3.04}
\expandafter\def\csname result@main/H2_glmer@value1\endcsname{-0.0}
\expandafter\def\csname result@main/H2_glmer@low1\endcsname{-2.5}
\expandafter\def\csname result@main/H2_glmer@high1\endcsname{2.3}
\expandafter\def\csname result@main/H2_glmer@low951\endcsname{-2.9}
\expandafter\def\csname result@main/H2_glmer@high951\endcsname{3.0}
\expandafter\def\csname result@main/H2_glmer@value2\endcsname{-0.00}
\expandafter\def\csname result@main/H2_glmer@low2\endcsname{-2.46}
\expandafter\def\csname result@main/H2_glmer@high2\endcsname{2.35}
\expandafter\def\csname result@main/H2_glmer@low952\endcsname{-2.85}
\expandafter\def\csname result@main/H2_glmer@high952\endcsname{3.02}
\expandafter\def\csname result@main/H3_brms@value1\endcsname{-1.9}
\expandafter\def\csname result@main/H3_brms@low1\endcsname{-4.9}
\expandafter\def\csname result@main/H3_brms@high1\endcsname{0.9}
\expandafter\def\csname result@main/H3_brms@low951\endcsname{-5.5}
\expandafter\def\csname result@main/H3_brms@high951\endcsname{1.5}
\expandafter\def\csname result@main/H3_brms@value2\endcsname{-1.90}
\expandafter\def\csname result@main/H3_brms@low2\endcsname{-4.95}
\expandafter\def\csname result@main/H3_brms@high2\endcsname{0.95}
\expandafter\def\csname result@main/H3_brms@low952\endcsname{-5.54}
\expandafter\def\csname result@main/H3_brms@high952\endcsname{1.50}
\expandafter\def\csname result@main/H3_glmer@value1\endcsname{-1.9}
\expandafter\def\csname result@main/H3_glmer@low1\endcsname{-5.0}
\expandafter\def\csname result@main/H3_glmer@high1\endcsname{1.0}
\expandafter\def\csname result@main/H3_glmer@low951\endcsname{-5.5}
\expandafter\def\csname result@main/H3_glmer@high951\endcsname{1.7}
\expandafter\def\csname result@main/H3_glmer@value2\endcsname{-1.90}
\expandafter\def\csname result@main/H3_glmer@low2\endcsname{-4.97}
\expandafter\def\csname result@main/H3_glmer@high2\endcsname{0.99}
\expandafter\def\csname result@main/H3_glmer@low952\endcsname{-5.45}
\expandafter\def\csname result@main/H3_glmer@high952\endcsname{1.67}
\expandafter\def\csname result@main/H4_brms@value1\endcsname{-1.0}
\expandafter\def\csname result@main/H4_brms@low1\endcsname{-4.6}
\expandafter\def\csname result@main/H4_brms@high1\endcsname{2.7}
\expandafter\def\csname result@main/H4_brms@low951\endcsname{-5.3}
\expandafter\def\csname result@main/H4_brms@high951\endcsname{3.3}
\expandafter\def\csname result@main/H4_brms@value2\endcsname{-1.00}
\expandafter\def\csname result@main/H4_brms@low2\endcsname{-4.63}
\expandafter\def\csname result@main/H4_brms@high2\endcsname{2.65}
\expandafter\def\csname result@main/H4_brms@low952\endcsname{-5.30}
\expandafter\def\csname result@main/H4_brms@high952\endcsname{3.32}
\expandafter\def\csname result@main/H4_glmer@value1\endcsname{-1.2}
\expandafter\def\csname result@main/H4_glmer@low1\endcsname{-5.0}
\expandafter\def\csname result@main/H4_glmer@high1\endcsname{2.4}
\expandafter\def\csname result@main/H4_glmer@low951\endcsname{-5.8}
\expandafter\def\csname result@main/H4_glmer@high951\endcsname{3.1}
\expandafter\def\csname result@main/H4_glmer@value2\endcsname{-1.23}
\expandafter\def\csname result@main/H4_glmer@low2\endcsname{-5.04}
\expandafter\def\csname result@main/H4_glmer@high2\endcsname{2.37}
\expandafter\def\csname result@main/H4_glmer@low952\endcsname{-5.78}
\expandafter\def\csname result@main/H4_glmer@high952\endcsname{3.13}
\expandafter\def\csname result@main/H5_brms@value1\endcsname{-1.9}
\expandafter\def\csname result@main/H5_brms@low1\endcsname{-4.6}
\expandafter\def\csname result@main/H5_brms@high1\endcsname{0.7}
\expandafter\def\csname result@main/H5_brms@low951\endcsname{-5.3}
\expandafter\def\csname result@main/H5_brms@high951\endcsname{1.2}
\expandafter\def\csname result@main/H5_brms@value2\endcsname{-1.85}
\expandafter\def\csname result@main/H5_brms@low2\endcsname{-4.63}
\expandafter\def\csname result@main/H5_brms@high2\endcsname{0.69}
\expandafter\def\csname result@main/H5_brms@low952\endcsname{-5.29}
\expandafter\def\csname result@main/H5_brms@high952\endcsname{1.16}
\expandafter\def\csname result@main/H5_glmer@value1\endcsname{-1.9}
\expandafter\def\csname result@main/H5_glmer@low1\endcsname{-4.4}
\expandafter\def\csname result@main/H5_glmer@high1\endcsname{0.8}
\expandafter\def\csname result@main/H5_glmer@low951\endcsname{-5.0}
\expandafter\def\csname result@main/H5_glmer@high951\endcsname{1.0}
\expandafter\def\csname result@main/H5_glmer@value2\endcsname{-1.85}
\expandafter\def\csname result@main/H5_glmer@low2\endcsname{-4.44}
\expandafter\def\csname result@main/H5_glmer@high2\endcsname{0.80}
\expandafter\def\csname result@main/H5_glmer@low952\endcsname{-4.97}
\expandafter\def\csname result@main/H5_glmer@high952\endcsname{1.02}
\expandafter\def\csname result@main/H6b_brms@value1\endcsname{2.2}
\expandafter\def\csname result@main/H6b_brms@low1\endcsname{-0.5}
\expandafter\def\csname result@main/H6b_brms@high1\endcsname{5.3}
\expandafter\def\csname result@main/H6b_brms@low951\endcsname{-1.1}
\expandafter\def\csname result@main/H6b_brms@high951\endcsname{6.0}
\expandafter\def\csname result@main/H6b_brms@value2\endcsname{2.19}
\expandafter\def\csname result@main/H6b_brms@low2\endcsname{-0.52}
\expandafter\def\csname result@main/H6b_brms@high2\endcsname{5.28}
\expandafter\def\csname result@main/H6b_brms@low952\endcsname{-1.08}
\expandafter\def\csname result@main/H6b_brms@high952\endcsname{5.95}
\expandafter\def\csname result@main/H6b_glmer@value1\endcsname{2.3}
\expandafter\def\csname result@main/H6b_glmer@low1\endcsname{-0.6}
\expandafter\def\csname result@main/H6b_glmer@high1\endcsname{5.4}
\expandafter\def\csname result@main/H6b_glmer@low951\endcsname{-1.2}
\expandafter\def\csname result@main/H6b_glmer@high951\endcsname{5.8}
\expandafter\def\csname result@main/H6b_glmer@value2\endcsname{2.27}
\expandafter\def\csname result@main/H6b_glmer@low2\endcsname{-0.60}
\expandafter\def\csname result@main/H6b_glmer@high2\endcsname{5.39}
\expandafter\def\csname result@main/H6b_glmer@low952\endcsname{-1.16}
\expandafter\def\csname result@main/H6b_glmer@high952\endcsname{5.83}
\expandafter\def\csname result@main/H6c_brms@value1\endcsname{4.1}
\expandafter\def\csname result@main/H6c_brms@low1\endcsname{0.4}
\expandafter\def\csname result@main/H6c_brms@high1\endcsname{8.5}
\expandafter\def\csname result@main/H6c_brms@low951\endcsname{-0.4}
\expandafter\def\csname result@main/H6c_brms@high951\endcsname{9.5}
\expandafter\def\csname result@main/H6c_brms@value2\endcsname{4.10}
\expandafter\def\csname result@main/H6c_brms@low2\endcsname{0.36}
\expandafter\def\csname result@main/H6c_brms@high2\endcsname{8.52}
\expandafter\def\csname result@main/H6c_brms@low952\endcsname{-0.36}
\expandafter\def\csname result@main/H6c_brms@high952\endcsname{9.49}
\expandafter\def\csname result@main/H6c_glmer@value1\endcsname{4.2}
\expandafter\def\csname result@main/H6c_glmer@low1\endcsname{0.2}
\expandafter\def\csname result@main/H6c_glmer@high1\endcsname{8.8}
\expandafter\def\csname result@main/H6c_glmer@low951\endcsname{-0.6}
\expandafter\def\csname result@main/H6c_glmer@high951\endcsname{9.7}
\expandafter\def\csname result@main/H6c_glmer@value2\endcsname{4.18}
\expandafter\def\csname result@main/H6c_glmer@low2\endcsname{0.17}
\expandafter\def\csname result@main/H6c_glmer@high2\endcsname{8.78}
\expandafter\def\csname result@main/H6c_glmer@low952\endcsname{-0.56}
\expandafter\def\csname result@main/H6c_glmer@high952\endcsname{9.67}
\expandafter\def\csname result@photometric_coupling/C0_brms@value1\endcsname{-0.3}
\expandafter\def\csname result@photometric_coupling/C0_brms@low1\endcsname{-3.1}
\expandafter\def\csname result@photometric_coupling/C0_brms@high1\endcsname{2.5}
\expandafter\def\csname result@photometric_coupling/C0_brms@low951\endcsname{-3.7}
\expandafter\def\csname result@photometric_coupling/C0_brms@high951\endcsname{3.1}
\expandafter\def\csname result@photometric_coupling/C0_brms@value2\endcsname{-0.28}
\expandafter\def\csname result@photometric_coupling/C0_brms@low2\endcsname{-3.08}
\expandafter\def\csname result@photometric_coupling/C0_brms@high2\endcsname{2.46}
\expandafter\def\csname result@photometric_coupling/C0_brms@low952\endcsname{-3.67}
\expandafter\def\csname result@photometric_coupling/C0_brms@high952\endcsname{3.08}
\expandafter\def\csname result@photometric_coupling/C1_brms@value1\endcsname{-1.1}
\expandafter\def\csname result@photometric_coupling/C1_brms@low1\endcsname{-3.6}
\expandafter\def\csname result@photometric_coupling/C1_brms@high1\endcsname{1.4}
\expandafter\def\csname result@photometric_coupling/C1_brms@low951\endcsname{-4.3}
\expandafter\def\csname result@photometric_coupling/C1_brms@high951\endcsname{1.8}
\expandafter\def\csname result@photometric_coupling/C1_brms@value2\endcsname{-1.05}
\expandafter\def\csname result@photometric_coupling/C1_brms@low2\endcsname{-3.64}
\expandafter\def\csname result@photometric_coupling/C1_brms@high2\endcsname{1.38}
\expandafter\def\csname result@photometric_coupling/C1_brms@low952\endcsname{-4.25}
\expandafter\def\csname result@photometric_coupling/C1_brms@high952\endcsname{1.82}
\expandafter\def\csname result@photometric_coupling/C5_brms@value1\endcsname{-2.0}
\expandafter\def\csname result@photometric_coupling/C5_brms@low1\endcsname{-4.8}
\expandafter\def\csname result@photometric_coupling/C5_brms@high1\endcsname{0.4}
\expandafter\def\csname result@photometric_coupling/C5_brms@low951\endcsname{-5.4}
\expandafter\def\csname result@photometric_coupling/C5_brms@high951\endcsname{0.8}
\expandafter\def\csname result@photometric_coupling/C5_brms@value2\endcsname{-2.05}
\expandafter\def\csname result@photometric_coupling/C5_brms@low2\endcsname{-4.80}
\expandafter\def\csname result@photometric_coupling/C5_brms@high2\endcsname{0.37}
\expandafter\def\csname result@photometric_coupling/C5_brms@low952\endcsname{-5.40}
\expandafter\def\csname result@photometric_coupling/C5_brms@high952\endcsname{0.85}
\expandafter\def\csname result@photometric_coupling/C8_brms@value1\endcsname{-1.0}
\expandafter\def\csname result@photometric_coupling/C8_brms@low1\endcsname{-3.8}
\expandafter\def\csname result@photometric_coupling/C8_brms@high1\endcsname{1.6}
\expandafter\def\csname result@photometric_coupling/C8_brms@low951\endcsname{-4.5}
\expandafter\def\csname result@photometric_coupling/C8_brms@high951\endcsname{2.2}
\expandafter\def\csname result@photometric_coupling/C8_brms@value2\endcsname{-0.99}
\expandafter\def\csname result@photometric_coupling/C8_brms@low2\endcsname{-3.79}
\expandafter\def\csname result@photometric_coupling/C8_brms@high2\endcsname{1.64}
\expandafter\def\csname result@photometric_coupling/C8_brms@low952\endcsname{-4.47}
\expandafter\def\csname result@photometric_coupling/C8_brms@high952\endcsname{2.18}

\author{Zixuan Jiang\textsuperscript{1,2},
Bo Liu\textsuperscript{2},
Xing Shi\textsuperscript{2},
Xudong Yang\textsuperscript{2},\\
Mengli Cheng\textsuperscript{2},
Jiaqi Xu\textsuperscript{2},
Kunzhe Huang\textsuperscript{2}\\[4pt]
\textsuperscript{1}College of Engineering, Peking University\\
\textsuperscript{2}Alibaba Cloud}

\title{When Does Cross-View Consistency Matter in Multi-View Appearance Augmentation?\\[4pt]
\large A Controlled Study for Visuomotor Policies}
\date{}
\begin{document}
\maketitle
\begin{abstract}
Multi-view visual augmentation often seeks consistency across cameras, but it is unclear which forms of consistency matter for downstream control. We use simulation-state replay to manipulate cross-view coupling while preserving the per-camera perturbation distributions, physical trajectories, and action labels. Across seven LIBERO manipulation tasks, marginal-matched changes in global appearance, ghost injection, and target color have small average costs for a late-fusion Diffusion Policy: point estimates are near zero or negative, and the largest upper endpoint of the 90\% posterior credible intervals is 2.55 percentage points. In contrast, geometrically anti-coupled observations reduce success by \res{geometry_coupling/R1b_c_brms}{value2} percentage points relative to coupled observations (90\% CrI [\res{geometry_coupling/R1b_c_brms}{low2}, \res{geometry_coupling/R1b_c_brms}{high2}]), with displacement magnitude, direction marginals, and the number of affected views matched. Camera interventions reveal substantial dependence on wrist-view geometry, while feature analysis shows appearance sensitivity patterns associated with the factors randomized during training. ACT shows a higher target-color cost estimate than DP, with uncertainty that includes zero and sensitivity to task composition. Within the tested tasks and policy configurations, these findings motivate selective consistency constraints: appearance coupling can sometimes be relaxed, whereas task-relevant geometry and cross-view spatial correspondence warrant protection.
\end{abstract}

\section{Introduction}
Generative visual augmentation can change the backgrounds, lighting, and object appearances in recorded robot demonstrations while reusing their action labels~\cite{cosmos,robotransfer,reshoot}. This offers a route to broader visual coverage without collecting new physical trajectories. For policies that observe both a third-person camera and a wrist camera, however, augmentation must also decide how the generated views relate to one another. Independently editing each camera is straightforward but may yield incompatible observations; generating views jointly can enforce consistency but adds modeling and computational requirements.

Does stronger cross-view consistency necessarily improve downstream control? A direct comparison between independent and joint generators does not isolate this question. Such generators can differ simultaneously in single-view quality, visual diversity, and cross-view coupling. Moreover, ``inconsistency'' is not a single intervention. Conflicting background textures, different target colors, and contradictory target positions need not have the same consequences for action selection.

We study the control value of coupling with a controlled simulation design. Recorded LIBERO states are replayed and rendered separately for each camera. We construct joint perturbation distributions with the same per-camera marginals, so that a camera receives the same distribution of perturbations regardless of how its perturbation is paired with the other view. This yields an estimate of the net control effect of changing cross-view coupling, at a fixed distribution of perturbations for each camera.

Our main experiments use a late-fusion Diffusion Policy (DP) on seven manipulation tasks. Marginal-matched global-appearance, ghost, and target-color contrasts have small average costs. At the same time, masking the wrist camera or introducing an unseen target displacement substantially reduces success. A further experiment matches the displacement magnitude, direction marginals, and number of affected cameras, and finds that geometrically anti-coupled observations are worse than coupled observations. Feature and co-visibility measurements provide descriptive context, and ACT supplies an exploratory comparison of policy configurations.

The contributions are threefold: (1) a state-replay design that manipulates cross-view joint distributions while preserving per-camera perturbation marginals; (2) downstream control estimates for several appearance contrasts and a marginal-matched geometry contrast; and (3) camera interventions and descriptive measurements that delineate the scope of the behavioral findings. Together, these contributions motivate allocating consistency constraints according to the visual information needed for control.

\begin{figure}[t]
\centering
\includegraphics[width=\linewidth]{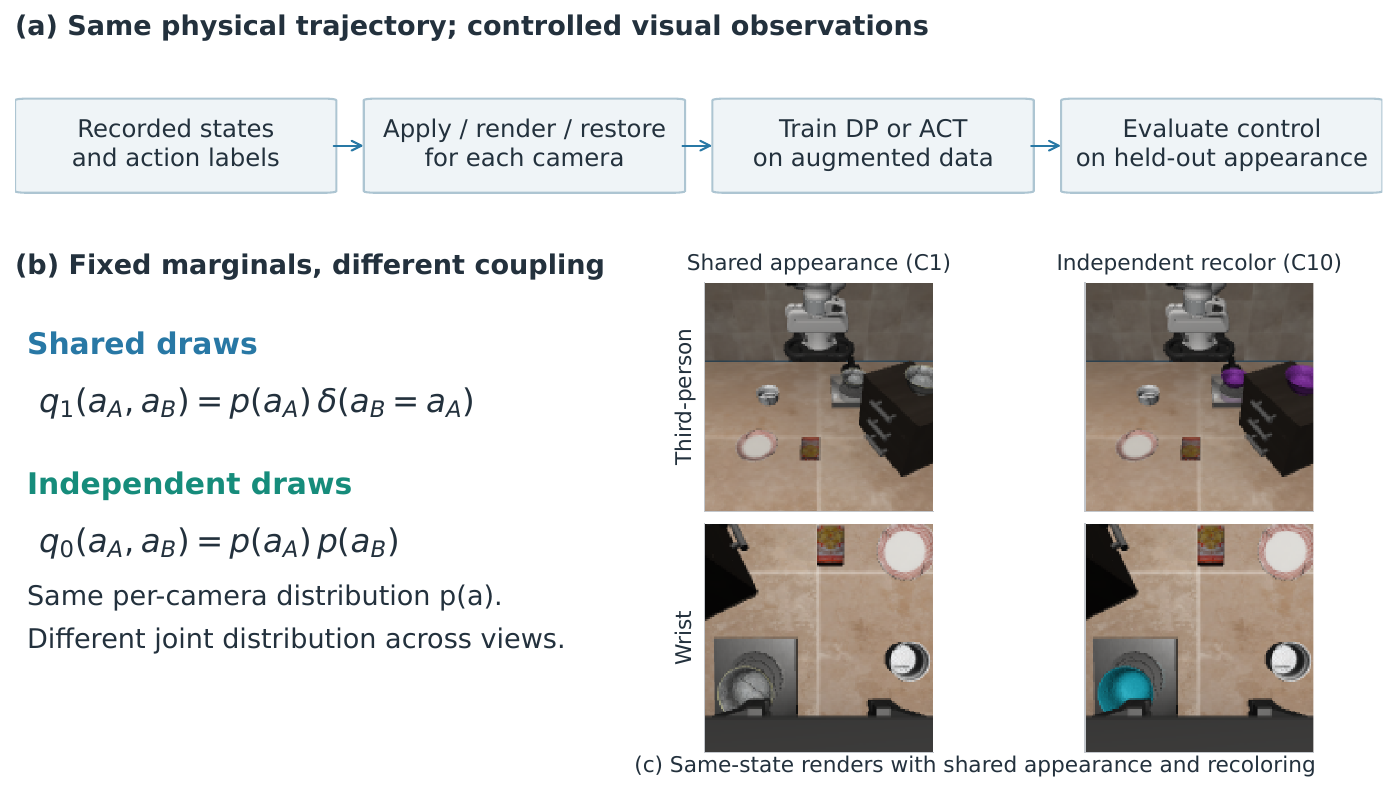}
\caption{\textbf{Controlled visual augmentation.} (a) State replay preserves physical trajectories and action labels while rendering each camera under its assigned recipe. (b) Shared and independent joint distributions have the same per-camera marginals. (c) Same-state examples of shared appearance (C1) and independent recoloring (C10), from \texttt{spatial\_8}, demo 0, frame 0, seed 1, variant 1. The historical renders are rotated into policy-input orientation. The matched recoloring experiment compares shared recoloring (C9) with C10.}
\label{fig:design}
\end{figure}

\section{Related Work}
\paragraph{Visual augmentation for robot learning.}
ROSIE uses generative inpainting to augment robot experience~\cite{rosie}; GenAug retargets demonstrated behavior to novel visual situations~\cite{genaug}. RoboEngine combines robot segmentation and background generation~\cite{roboengine}, while GreenAug changes backgrounds through green-screen compositing~\cite{greenaug}. Video-based approaches extend these ideas to recorded sequences: Cosmos-Transfer2.5 supports controlled world translation~\cite{cosmos}, and ReShoot re-renders recorded demonstrations under altered appearances while retaining action and proprioceptive labels~\cite{reshoot}. Our interventions are motivated by this progression from image editing to demonstration-level visual augmentation.

\paragraph{Multi-view generation.}
RoboTransfer explicitly uses cross-view interactions and geometric structure to improve consistency~\cite{robotransfer}. RoboVIP emphasizes multi-view, temporally coherent generation with visual identity prompting~\cite{robovip}; EMMA introduces generative manipulation-video augmentation that targets geometric and multi-view consistency~\cite{emma}. Such methods seek consistent synthetic observations and evaluate downstream policies. Our complementary question is narrower: what changes when coupling is manipulated while the single-view perturbation marginals are held fixed? The state-replay design makes that contrast directly accessible.

\paragraph{Randomization, robustness, and viewpoint choice.}
Domain randomization varies rendered appearances to support transfer beyond a training environment~\cite{domainrandomization}. RAD and DrQ demonstrate the value of image augmentation for pixel-based reinforcement learning~\cite{rad,drq}. LIBERO provides diverse manipulation tasks and demonstrations~\cite{libero}; LIBERO-Plus and THE COLOSSEUM expose vulnerabilities to visual and other environmental changes~\cite{liberoplus,colosseum}. RoboTwin 2.0 scales simulation data generation with structured domain randomization~\cite{robotwin}. Separately, Hsu et al. show that hand-centric observations can improve learning and generalization when their observability is sufficient~\cite{hands}. We distinguish the value of a viewpoint from the value of keeping every visual factor consistent across viewpoints.

\section{Controlled Design}\label{sec:method}
\subsection{Coupling with fixed per-camera marginals}
Let $A$ denote the third-person camera and $B$ the wrist camera. An appearance recipe $a$ is drawn from $p(a)$. Shared and independent recipe sampling define
\begin{align}
q_1(a_A,a_B)&=p(a_A)\,\delta(a_B=a_A), &
q_0(a_A,a_B)&=p(a_A)p(a_B),\\
q_\rho&=\rho q_1+(1-\rho)q_0, & \rho&\in[0,1].
\end{align}
Both marginals equal $p$ for every $\rho$. Recipes remain fixed within a demonstration variant. Each training set also retains an unmodified copy of every demonstration, identically across the augmented conditions. Thus the common original-data component does not change the marginal-matching property.

For policies trained under these distributions, the primary appearance contrast is
\begin{equation}
\Delta=\SR(q_1)-\SR(q_\rho),
\end{equation}
where $\SR$ is success on a fixed, held-out appearance set whose two views are consistent. Positive $\Delta$ means lower success after reducing coupling. This mixture specifically defines the global-appearance experiment; ghost, color, and geometry experiments use their own joint distributions below.

We distinguish \emph{single-view fidelity}, whether each observation is correct, from \emph{cross-view consistency}, whether observations are mutually compatible, and \emph{cross-view combination diversity}, how many combinations of perturbations are encountered. Reducing coupling can increase both visual conflict and combination diversity. Our estimates therefore measure the net effect of changing coupling, not a pure latent penalty for contradictory features.

\subsection{State replay and visual injection}
We restore each recorded simulator state rather than re-executing actions. For each camera in turn, we apply a visual recipe, render that camera, and restore the scene before processing the next camera. The physical trajectory, collision geometry, and recorded action labels are unchanged. Appearance manipulations alter visual content only. Target-offset manipulations move the rendered target but leave its physical location and collisions unchanged; the resulting observation is intentionally inconsistent with the original target geometry, even though the physical action label remains unchanged.

A small Cosmos-Transfer2.5 pilot informed the intervention types. Five task segments were generated under independent and shared prompt/seed schemes, producing ten paired-view units. Each clip contains 93 frames at 720p and 16 fps. Manual labels describe these examples, not population failure frequencies (Appendix~\ref{app:pilot}).

\paragraph{Marginal-matched appearance contrasts.}
Global appearance changes textures and effective light intensity/color, with $\rho\in\{1,0.5,0\}$. Coupled ghost injection enables a collision-free visual distractor in both cameras or neither, each with probability $1/2$; anti-coupled injection enables it in exactly one camera, selected uniformly. Each camera therefore has the same injection probability and recipe distribution. Being enabled does not guarantee image visibility, which we measure separately. Target recoloring applies to the grasped object's asset group with probability $1/2$ and draws from the same color pool. The paired conditions share or independently draw both the switch and color. Meshes, shapes, and object identities are retained. Conditions share keyed random streams to reduce comparison variance.

\paragraph{Photometric controls.}
C8 adds independent image-space white-balance and brightness jitter to the shared-appearance condition C1; C1 has no such jitter. The training contrast C1--C8 therefore changes the single-view perturbation distribution and cannot isolate coupling. We retain it as an additional-perturbation control. A separate test-time experiment compares T-shared and T-indep using identical per-camera photometric marginals, alongside an unperturbed T0 reference. Thus the two experiments characterize training-time robustness and test-time coupling, respectively.

\subsection{Geometry: fidelity and coupling}
Offset-Wrist (C11) is a positive control that shifts the rendered target by 10 cm in the wrist camera only, along a fixed task-wise direction $d$. The direction is selected from eight candidates to maximize separation from other objects. An exploratory Offset-Both condition (C12) shifts both views in direction $d$. Because C11 and C12 differ in the number of affected views as well as their relationship, their comparison does not identify coupling alone.

For the marginal-matched geometry experiment, we choose $d'$ from $d\pm90^\circ$, selecting the candidate with greater clearance. Both cameras are offset by 10 cm, but directions are paired differently:
\begin{align}
\text{Offset-Coupled: }&(d,d)\text{ or }(d',d'),\quad\text{each with probability }1/2,\\
\text{Offset-Anti: }&(d,d')\text{ or }(d',d),\quad\text{each with probability }1/2.
\end{align}
Each camera has the same $50\%/50\%$ direction marginal in both conditions. The $90^\circ$ separation retains a nonzero average displacement, rather than turning a systematic error into zero-mean noise. The primary endpoint is $c=\SR(\mathrm{Coupled})-\SR(\mathrm{Anti})$. The supporting endpoint $L=\SR(\mathrm{C1})-\SR(\mathrm{Coupled})$ concerns geometric fidelity but uses an older reference batch. Fourteen retrained C1 policies (seeds 1--2) describe batch sensitivity; they neither replace nor correct the main C1 reference. Appendix~\ref{app:design} gives all directions and the model encoding.

Finally, Offset-Test (E1) applies a 10 cm wrist-view displacement only at test time to C1 policies that were never trained on offsets. Zero- and 10-cm episodes are paired by initial state. This measures sensitivity to unexpected wrist-view geometry.

\subsection{Co-visibility and feature measurements}
Co-visibility uses demonstrations 0--4 of each task, sampling every ten frames at $128\times128$. Direct visibility requires at least 20 pixels in each view. Projected co-visibility additionally back-projects source pixels using ground-truth depth and camera calibration and checks occlusion in the destination view with a 0.02 m depth tolerance. We report direct visibility and projected co-visibility separately, with element- and direction-specific denominators.

Feature analysis uses four equally spaced states from demonstration 0 of each task, rendering the original appearance and three variants at each state. The invariance ratio is the mean feature RMS distance between appearances of the same state divided by the mean distance between states at the same appearance. Lower ratios indicate less appearance sensitivity relative to state sensitivity. We extract spatial-softmax keypoints from DP and visual tokens after ACT's $1\times1$ projection but before cross-view attention. Only seed-1 checkpoints are used; camera ratios are averaged and proprioception is zeroed to examine the visual pathway. This protocol measures appearance sensitivity relative to state sensitivity within each representation.

\section{Experimental Setup}\label{sec:setup}
\subsection{Tasks, data, and policies}
We fixed a pool of 16 candidate LIBERO tasks and trained an unaugmented policy using 15 demonstrations per task. Seven tasks were selected using a single-seed, 20-episode screening success range of 40--90\%. Table~\ref{tab:tasks} lists the tasks; their formal baseline success rates appear in Table~\ref{tab:rates}. Each augmented demonstration contributes four appearance variants and one original copy. Training seeds correspond to separately rendered augmented datasets; C0 reuses a single unaugmented dataset per task.

\begin{table}[t]
\centering\small
\caption{Seven LIBERO tasks. Short identifiers omit the \texttt{libero\_} prefix.}\label{tab:tasks}
\begin{tabularx}{\linewidth}{lX}
\toprule Task & Instruction \\\midrule
\texttt{10\_5} & Put both the alphabet soup and the tomato sauce in the basket.\\
\texttt{10\_9} & Pick up the book and place it in the back compartment of the caddy.\\
\texttt{goal\_1} & Open the top drawer and put the bowl inside.\\
\texttt{goal\_6} & Put the cream cheese in the bowl.\\
\texttt{goal\_7} & Put the wine bottle on the rack.\\
\texttt{spatial\_2} & Pick up the black bowl in the top drawer of the wooden cabinet and place it on the plate.\\
\texttt{spatial\_8} & Pick up the black bowl on the stove and place it on the plate.\\
\bottomrule
\end{tabularx}
\end{table}

Our main policy is Diffusion Policy~\cite{diffusion}, implemented in LeRobot~\cite{lerobot}, with separate ResNet18 encoders and spatial softmax for the two cameras. Camera features are concatenated before a diffusion action head, giving late fusion. Training runs for 20,000 steps with batch size 64, two observation frames, a 16-step prediction horizon, and eight executed actions per prediction. Inference uses ten DDIM steps. ACT~\cite{act} is trained for 25,000 steps and combines visual tokens within a transformer. Both configurations use $128\times128$ policy inputs and the final checkpoint. DP evaluation renders at $256\times256$ before policy resizing; ACT evaluation renders at $128\times128$. Most data and evaluations use EGL; geometry-coupling data and E1 use OSMesa.

DP and ACT differ in their visual feature representation, fusion location, and action head. Their comparison is therefore a comparison of policy configurations, not an isolated intervention on attention or fusion. A three-camera DP configuration is reported only as supplementary evidence because its added-view dependence prerequisite failed (Appendix~\ref{app:suppresults}).

\subsection{Sample accounting and evaluation}
The base experiments contain 784 policies and 74,060 evaluation episodes. Geometry coupling adds 98 policies and 14 C1 batch anchors, yielding 896 distinct policies in total. The main geometry analysis reuses 49 base-batch C1 policies, so it comprises 147 policies and 2,940 episodes; the reused policies are not counted twice in the total. Most conditions use five to seven training seeds. Formal ACT C1 uses seeds 2--5; its pilot models are excluded. Appendix~\ref{app:design} provides the condition grid.

The primary evaluation uses held-out, mutually consistent appearances (set B), with 20 episodes per policy and fixed initial states. Where conditions can be paired, episodes use matching initial states. Additional evaluations cover the original scene, camera masking, unseen colors, test-time target offsets, and photometric coupling. The latter uses 168 existing policies across T0, T-shared, and T-indep, totaling 10,080 episodes. Camera masking zeros the selected raw camera image after every reset and step, while leaving other cameras and proprioception unchanged; it is an out-of-distribution intervention.

\subsection{Statistical analysis}\label{sec:statistics}
The main analyses use binomial-logit mixed models of the form
\begin{equation}
\operatorname{logit}P(Y_{t,s,e}=1)=\alpha+X_{t,s,e}\beta+u_t+v_{t,s},
\qquad u_t\sim\mathcal{N}(0,\sigma_t^2),\quad v_{t,s}\sim\mathcal{N}(0,\sigma_{ts}^2).
\end{equation}
Here $t$, $s$, and $e$ index task, training seed, and evaluation episode. The design matrix contains the relevant condition, dose, or interaction terms. For the DP--ACT interaction, task--seed identifiers are additionally nested within policy configuration. For each posterior draw, success probabilities are obtained by logistic-normal integration over the random intercepts using 32-node Gauss--Hermite quadrature; probability differences are expressed in percentage points (pp). These are model-marginalized average effects estimated from the seven-task dataset, not raw equal-task sample means. A separate fixed condition-by-task model integrates only the task--seed intercept and averages its seven task-specific effects equally. We distinguish these estimands in the tables and do not interpret the main interval as a guarantee for each task or a new-task prediction interval.

We report posterior means and 90\% credible intervals (CrI), with 95\% intervals for the primary geometry comparison and selected supplements. Fixed slopes have $\mathcal{N}(0,2.5^2)$ priors, the intercept has a $\mathcal{N}(0,5^2)$ prior, and random-effect standard deviations have exponential priors with rate 1. The brms fits use four chains of 4,000 iterations, including 1,000 warm-up iterations per chain. Final fits have no divergent transitions and maximum $\hat R\leq1.003$. Leave-one-task-out fits and alternative estimators assess sensitivity to task composition and modeling choices.

After initial results had been observed, presentation shifted from the early planned tests to full-MCMC effect estimation. The geometry endpoints and decision rules were specified before that experiment's formal training. Earlier tests, subsequent changes, and limitations of the records remain documented in Appendices~\ref{app:sensitivity} and~\ref{app:transparency}.

\section{Results}\label{sec:results}
\subsection{Small average costs for the matched appearance contrasts}
\begin{figure}[t]
\centering\includegraphics[width=\linewidth]{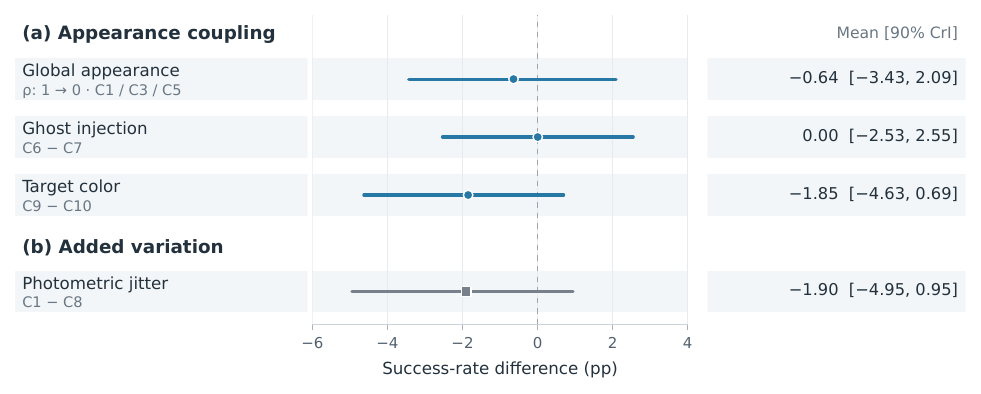}
\caption{\textbf{Average effects of appearance coupling and added photometric variation.} Points and bars show posterior means and 90\% CrI. (a) Coupling contrasts with matched per-camera perturbation distributions. (b) The effect of adding independent photometric jitter to C1. Positive values indicate lower success in the target condition; negative values favor it.}
\label{fig:appearance}
\end{figure}

The three marginal-matched DP appearance contrasts have point estimates near zero or below it (Figure~\ref{fig:appearance}; Table~\ref{tab:main_effects}). Global-appearance decorrelation yields \eff{main/H1_brms} pp, ghost anti-coupling yields \eff{main/H2_brms} pp, and independent target recoloring yields \eff{main/H5_brms} pp. Their largest 90\% upper endpoint is 2.55 pp. The estimates place the average coupling costs close to zero in this task set.

The descriptive global-appearance success rates are 82.9\%, 83.9\%, and 83.5\% at $\rho=1$, $0.5$, and $0$, respectively, with no clear dose-response pattern (Figure~\ref{fig:dose}). Across leave-one-task-out fits, the maximum 90\% upper endpoints are 3.30 pp for global appearance, 3.25 pp for ghosts, and 1.89 pp for target color. Removing \texttt{10\_9} eliminates the negative global-appearance estimate. The direction of the small average effects therefore depends partly on task composition. New-task predictive intervals from a separate random-slope analysis are wider than average-effect intervals (Appendix~\ref{app:sensitivity}).

Adding independent photometric jitter in C8 gives a C1--C8 contrast of \eff{main/H3_brms} pp. This additional-perturbation control also has limited average impact (Figure~\ref{fig:appearance}b).

\subsection{Geometric anti-coupling reduces success under matched marginals}
\begin{figure}[t]
\centering\includegraphics[width=\linewidth]{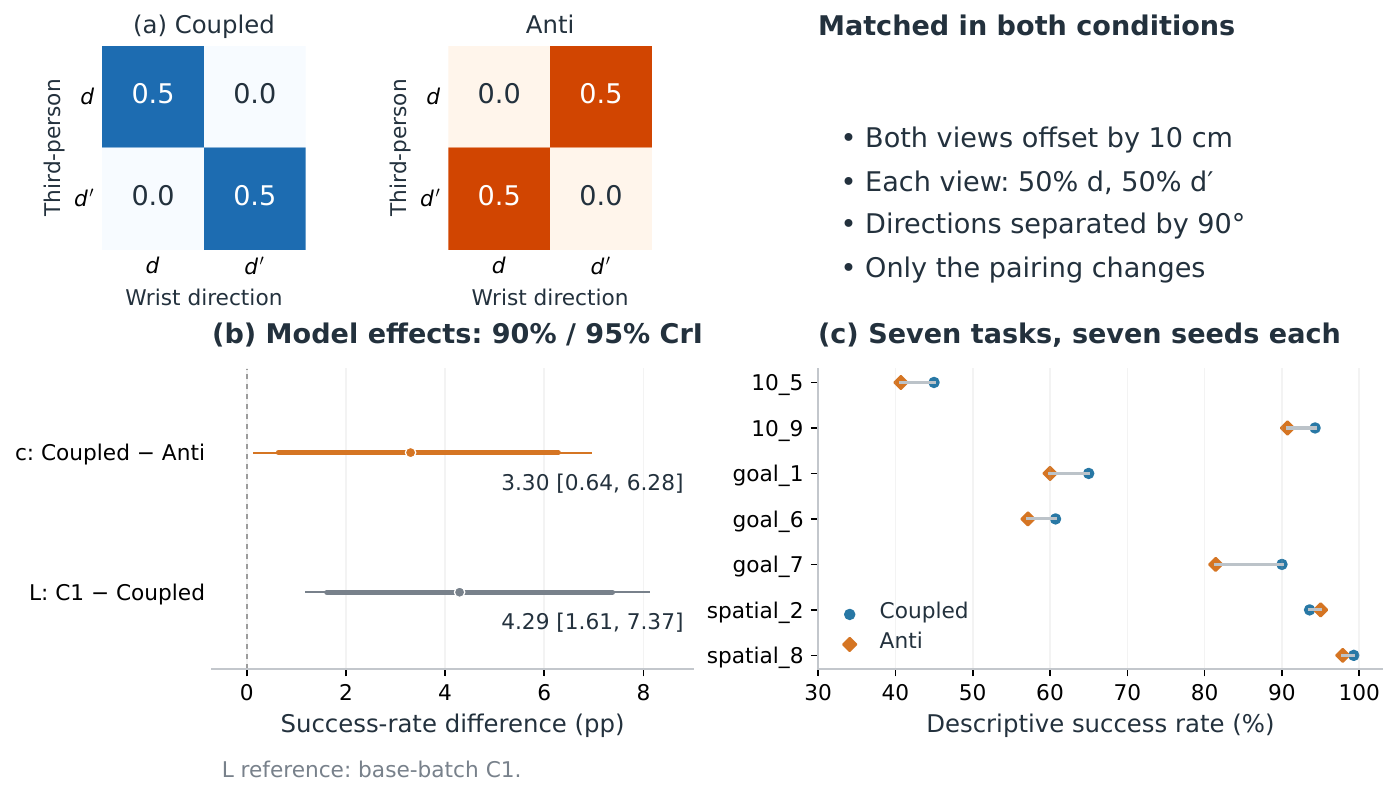}
\caption{\textbf{Marginal-matched geometry experiment.} (a) Joint direction probabilities; every row and column sums to 0.5. (b) Posterior means, thick 90\% CrI, and thin 95\% CrI. The primary within-batch contrast $c$ differs from the supporting cross-batch contrast $L$. (c) Descriptive success by task, with segments linking the two conditions. The two new conditions use 49 policies each.}
\label{fig:geometry}
\end{figure}

With both views displaced by 10 cm and the per-camera direction marginals matched, Offset-Anti reduces success relative to Offset-Coupled by \eff{geometry_coupling/R1b_c_brms} pp (Figure~\ref{fig:geometry}). The 95\% CrI is [\res{geometry_coupling/R1b_c_brms}{low952}, \res{geometry_coupling/R1b_c_brms}{high952}] pp. Six of seven tasks have positive descriptive Coupled--Anti differences; the task-equal descriptive mean is 3.6 pp (Table~\ref{tab:geo_tasks}). The main difference between these conditions is how displacement directions are paired across cameras, rather than the amount of error in either camera individually.

The supporting fidelity-related contrast is $L=$\eff{geometry_coupling/R1b_L_brms} pp. Unlike $c$, this contrast uses the older C1 batch. Retrained C1 anchors differ from base-batch C1 by an average of +5.7 pp, with +32.5 pp on \texttt{10\_9} and smaller, mixed-sign changes elsewhere. The anchor variation identifies batch sensitivity in the reference. The same-batch contrast $c$ therefore provides the main evidence for the geometric coupling effect.

\subsection{Policies retain strong dependence on wrist-view geometry}
\begin{figure}[t]
\centering\includegraphics[width=\linewidth]{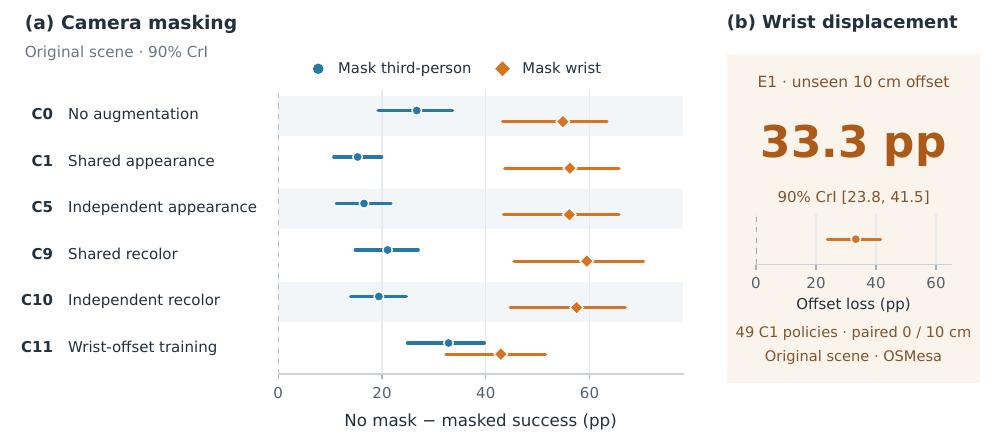}
\caption{\textbf{Sensitivity to camera masking and wrist-view displacement.} (a) Within-condition effects of masking one camera on the original scene. (b) E1 measures first exposure to a wrist target offset using paired 0/10 cm episodes and OSMesa rendering. Points and bars show posterior means and 90\% CrI.}
\label{fig:interventions}
\end{figure}

The 49 shared-appearance C1 policies lose \effone{camera_mask/E1_brms} pp when first exposed to a 10 cm wrist-view target offset at test time. Masking the wrist camera yields a loss of \effone{camera_mask/dp_C1_wrist_brms} pp (Figure~\ref{fig:interventions}). Both interventions reveal substantial reliance on wrist observations, including their geometric content.

Shared- and independent-appearance training yield similar wrist-masking losses. The modeled difference in masking losses, independent minus shared, is \eff{main/H4_brms} pp. Wrist-view dependence is thus retained under independent appearance augmentation. Training with Offset-Wrist also reduces success: the mean costs are \res{main/H0a_brms}{value2} pp for DP, \res{main/H0b_brms}{value2} pp for three-camera DP, and \res{main/H0c_brms}{value2} pp for ACT. The DP positive control is sensitive to removing \texttt{10\_5}, and the ACT comparison is confounded with training machine; these qualifications are retained in the supplementary analyses.

\subsection{Co-visibility and factor-specific feature sensitivity}
\begin{figure}[t]
\centering\includegraphics[width=\linewidth]{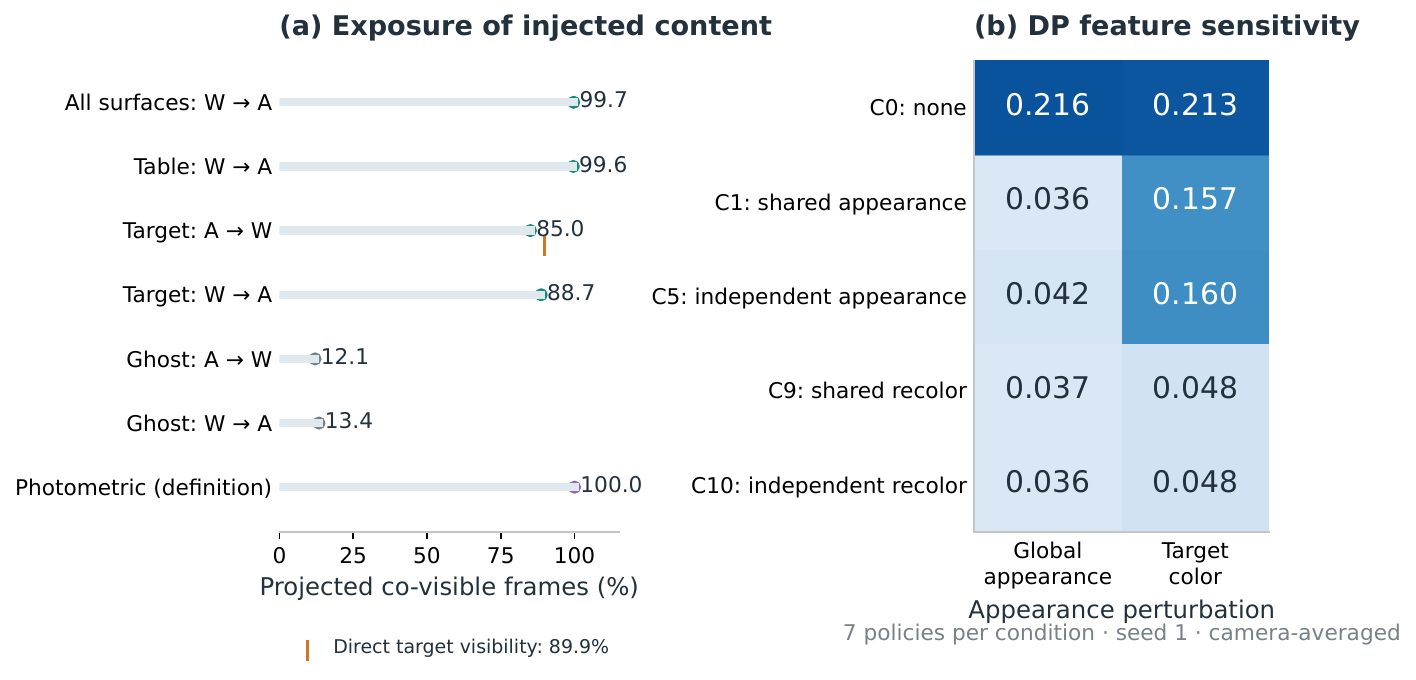}
\caption{\textbf{Co-visibility and factor-specific feature sensitivity.} (a) Projected co-visibility by element and direction; A denotes third-person and W wrist. The orange mark gives pooled direct target visibility. Photometric coverage is 100\% by definition. (b) DP feature-invariance ratios for two perturbation types, averaged over both cameras and seven seed-1 policies per condition. Lower values indicate smaller appearance-induced changes relative to state-induced changes.}
\label{fig:mechanisms}
\end{figure}

The low appearance costs occur across markedly different levels of cross-view exposure (Figure~\ref{fig:mechanisms}). The altered surfaces are projected co-visible in 99.7\% of their measurement frames. The table occupies about 48\% of the wrist image and is projected co-visible in 99.6\% of frames. Target objects are directly visible in both views in 89.9\% of the pooled target records, with directional projected co-visibility of 85.0--88.7\%. Ghosts have much lower projected co-visibility, at 12.1--13.4\%. Thus the surface and color results concern frequently co-visible content, whereas the ghost result concerns sparse cross-view exposure.

For DP, the global-appearance invariance ratio falls from 0.216 without augmentation to 0.036/0.042 with shared/independent appearance training. Its target-color ratio remains 0.157/0.160 under those conditions, but decreases to 0.048 with either shared or independent target-color augmentation. The reduction in feature sensitivity tracks the factors randomized during training. Shared and independent augmentation give similar ratios within each factor, complementing their similar behavioral effects. The full DP and ACT measurements are reported in Table~\ref{tab:features}.

\subsection{Matched test-time photometric coupling}
The test-time experiment compares T-shared and T-indep on the original scene, with white-balance factors in $[0.9,1.1]$ and brightness factors in $[0.85,1.15]$. Its four training conditions have coupling-cost estimates from -2.05 to -0.28 pp, all with 90\% intervals including zero (Figure~\ref{fig:photo}; Table~\ref{tab:photo}). C0, C1, and C5 were not trained with independent image-space photometric jitter; C8 was trained with a wider range. Even C0 changes little between T0 and T-shared (86.6\% versus 86.3\% descriptive success), so this is a mild perturbation. The result cannot by itself distinguish augmentation-induced invariance from low sensitivity already present without augmentation.

\subsection{Task heterogeneity and exploratory policy differences}
\begin{figure}[t]
\centering\includegraphics[width=\linewidth]{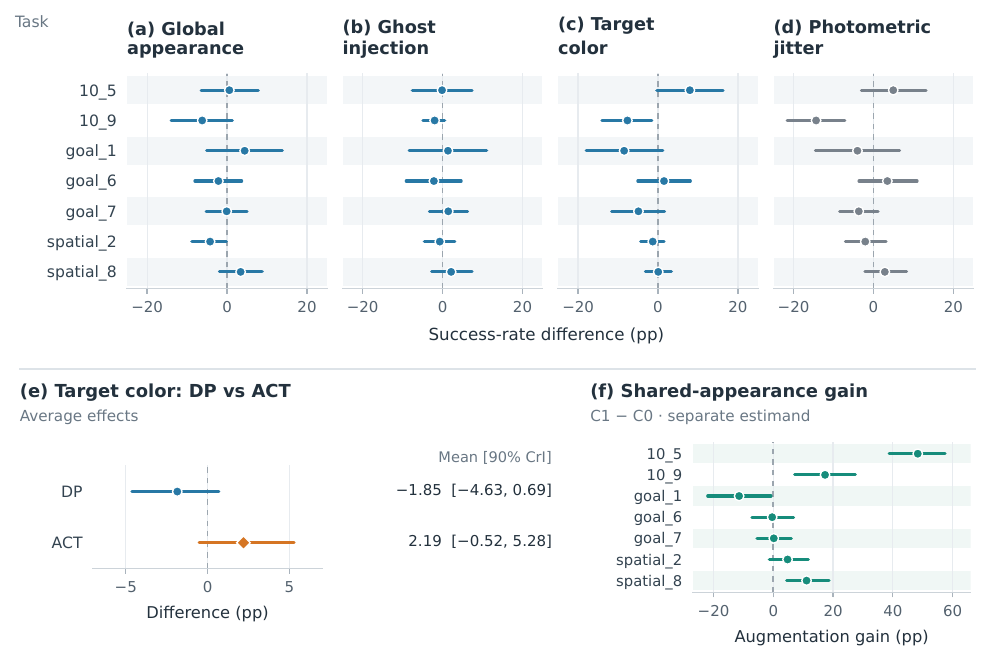}
\caption{\textbf{Task heterogeneity and policy configuration.} (a--d) Task-specific estimates from fixed task-interaction models. Panels (a--c) measure coupling effects; (d) measures added photometric variation. (e) Average target-color effects for DP and ACT. (f) Shared-appearance augmentation gain relative to C0. Bars denote 90\% CrI. Panel (f) uses a wider horizontal range to accommodate the large gain on \texttt{10\_5}.}
\label{fig:heterogeneity}
\end{figure}

ACT's target-color cost is \eff{main/H6b_brms} pp, compared with DP's \eff{main/H5_brms} pp (Figure~\ref{fig:heterogeneity}). The ACT--DP interaction is \eff{main/H6c_brms} pp, but its interval includes zero after removing any of \texttt{10\_9}, \texttt{goal\_1}, \texttt{goal\_7}, or \texttt{spatial\_2}. The individual ACT color-cost interval also includes zero. The higher ACT point estimate is accompanied by uncertainty and task-composition sensitivity. Training-machine sensitivity is examined in Appendix~\ref{app:transparency}.

Augmentation benefit and coupling cost are separate questions. The fixed task-interaction model estimates a 10.0 pp [6.8, 13.1] average benefit for shared appearance relative to no augmentation, ranging from -11.4 pp on \texttt{goal\_1} to +48.3 pp on \texttt{10\_5}. The appearance contrasts are also heterogeneous. Of 28 exploratory task-by-contrast cells, three 90\% intervals exclude zero, all favoring the less-coupled or additionally perturbed condition: global appearance on \texttt{spatial\_2}, and photometric jitter and target color on \texttt{10\_9}. Conversely, the color cost on \texttt{10\_5} is 8.0 pp [-0.3, 16.3], leaving considerable room for task-specific harm. The task-level analysis reveals variation around the small average effect.

\section{Discussion and Limitations}\label{sec:discussion}
\paragraph{Consistency should be selective.}
The matched experiments distinguish cross-view coupling from single-view fidelity. They show that changing the pairing of geometric displacements can reduce success even when the amount and marginal distribution of error in each view are unchanged. Meanwhile, several appearance-coupling changes have small average costs for the tested DP. We do not infer a universal ranking between geometry and appearance: the intervention types were not dose-matched, and no formal cross-type effect-difference test was performed.

For generative augmentation, the findings motivate exploring looser coupling for some background and appearance factors while protecting task-object identity, shape, and position, including spatial correspondence between cameras. Promising directions include preserving task-object pixels, sharing structural references, and imposing geometry constraints on target regions. Target color may need consistency when color is task-relevant or under different policy configurations. Our recoloring experiments keep identity and shape fixed and do not justify identity remapping or object deletion.

\paragraph{Appearance invariance and geometric dependence can coexist.}
A policy may be insensitive to a randomized appearance factor while retaining dependence on geometric information in the same camera. Camera interventions and factor-specific feature ratios are compatible with this interpretation. They do not prove it: masking is an out-of-distribution manipulation, co-visibility does not show that a policy compares corresponding regions, and a feature-distance ratio can change through its numerator, denominator, or both. More seeds, broader state sampling, and representation interventions would be needed for a causal mechanism claim.

\paragraph{Scope of generalization.}
The study uses seven selected simulation tasks, $128\times128$ policy inputs, and ResNet18 visual encoders trained from scratch. The task screen was noisy: \texttt{10\_5}'s unaugmented success fell from 70\% in screening to 27\% in formal held-out-appearance evaluation. The task set is not a representative random sample of robotic manipulation. Current-task averages do not establish low costs for unseen tasks, pretrained visual backbones, vision-language-action models, or real robots. Geometry coupling was tested only for DP, 10 cm displacements, and a $90^\circ$ direction separation. High-co-visibility ghosts, third-person-only offsets, and ACT geometry coupling were not tested.

\paragraph{Design and reproducibility limitations.}
Coupling contrasts combine any penalty from conflict with any benefit from combination diversity. Training photometric jitter is an unmatched control; its matched test-time counterpart is mild even for unaugmented policies. ACT differs from DP in several components and has training-machine confounding. Historical and current analyses have different estimands and uncertainty procedures, and some early recovery and public-registration records remain incomplete. Rendering backend changes can alter pixels and descriptive feature/co-visibility measurements; saved paper-run measurements are retained rather than silently replaced. Appendices~\ref{app:sensitivity} and~\ref{app:transparency} document these limitations and the analytical changes.

\section{Conclusion}
By replaying fixed physical trajectories and changing the joint distribution of per-camera perturbations, we study when cross-view coupling matters for visuomotor control. Global-appearance, ghost, and target-color coupling changes have small average costs for the tested DP configuration. A marginal-matched geometric anti-coupling intervention nevertheless reduces success by \res{geometry_coupling/R1b_c_brms}{value2} pp. Wrist-camera interventions and descriptive feature analysis show that geometric dependence and appearance insensitivity need not be contradictory. These results motivate selective consistency in multi-view augmentation, with priority given to task-relevant geometric fidelity and correspondence.

\FloatBarrier
\bibliographystyle{unsrtnat}
\bibliography{references}
\clearpage
\appendix
\setcounter{figure}{0}
\renewcommand{\thefigure}{S\arabic{figure}}
\counterwithin{table}{section}
\section{Implementation and Experimental Inventory}\label{app:design}
\subsection{Conditions and sample accounting}
Table~\ref{tab:conditions} maps descriptive names to historical identifiers. These identifiers locate configurations; they are not ordinal levels of a common perturbation scale.
\begin{table}[htbp]
\centering\small
\caption{Condition inventory.}\label{tab:conditions}
\begin{tabularx}{\linewidth}{lX}
\toprule Identifier & Definition \\\midrule
C0 & No augmentation; original observations only.\\
C1 / C3 / C5 & Global-appearance sharing with $\rho=1/0.5/0$.\\
C6 / C7 & Ghost injection: both-or-neither versus exactly one camera.\\
C8 & C1 plus independent image-space photometric jitter.\\
C9 / C10 & Shared versus independent target-color switch and draw.\\
C11 & Offset-Wrist: visual target shifted 10 cm in the wrist view.\\
C12 & Offset-Both: same-direction shift in both views; exploratory.\\
E1 & Test-only 0/10 cm wrist offset on C1 policies; exploratory.\\
Offset-Coupled & Two directions paired identically across views; 49 new policies.\\
Offset-Anti & Two directions paired oppositely across views; 49 new policies.\\
C1 batch anchor & Fourteen retrained C1 policies, seeds 1--2; descriptive only.\\
\bottomrule
\end{tabularx}
\end{table}

The base inventory comprises 434 two-camera DP policies excluding C12, 182 three-camera DP policies, 133 ACT policies, and 35 C12 policies (784 total). The geometry experiment adds 98 policies plus 14 anchors (896 total). ACT C1 uses seeds 2--5 (28 policies); C9, C10, and C11 use seeds 1--5 (35 each), all at 25k steps. Seven seed-1 50k ACT pilots, two 25k half-budget checks, and separate machine checks are outside that inventory. The 49 original C1 policies in the geometry analysis are already counted in the base inventory.

The frozen analysis package contains 93,660 evaluation rows: 74,060 base rows, 5,320 additional camera-mask rows, 2,240 geometry/anchor rows, 1,960 E1 rows, and 10,080 test-time photometric rows. These rows cover distinct evaluation units, not 93,660 independent policy trainings. The main geometry models combine 980 existing C1 rows with 1,960 Coupled/Anti rows and exclude the 280 new-anchor rows. The base episode table also contains evaluations not used for the principal endpoint; its total is not the main-model sample size.

\subsection{Injection parameters and actual rendering behavior}
\paragraph{Global appearance.}
The texture pools contain 16 table textures (13 train/3 test), 22 wall textures (18/4), and 25 floor textures (20/5). Recipes sample light intensity factors from 0.5--1.2 and color temperatures from 4000--7500 K. Light azimuth (0--360 degrees) and elevation (30--80 degrees) are also sampled in configuration, but did \emph{not} alter the rendered light direction in the historical pipeline: the updated model positions did not propagate to runtime light positions without a forward call. Injectors requiring a forward call restore the baseline light position, preserving this behavior across conditions. Accordingly, effective appearance variation comprises texture, light intensity, and light color, not the configured direction range. Held-out B appearances use test textures with baseline lighting.

\paragraph{Photometric jitter.}
C8 uses per-camera white-balance factors $U[0.8,1.2]$ and brightness $U[0.7,1.3]$, fixed within each episode and camera. T-shared/T-indep use the narrower ranges $U[0.9,1.1]$ and $U[0.85,1.15]$. A shared image-space photometric training counterpart to C8 was not run; shared global appearance in C1 is not such a counterpart.

\paragraph{Ghosts and target color.}
Ten collision-free visual assets are sampled within $x,y\in[-0.3,0.3]$ m and $z\in[0.92,1.1]$ m. Placement requires at least 8 cm clearance from every movable task-object center, with up to 2,000 attempts. It does not guarantee visibility from the other camera. Target-color augmentation uses nine training colors; brown, pink, and gray are held out. Instances in the target's asset group are recolored together with probability 0.5. Shared and independent conditions differ in whether the switches and draws are shared. This does not reproduce semantic identity swaps, shape changes, or missing objects.

\paragraph{Orientation and resolution.}
Raw simulator images are rotated by 180 degrees before LeRobot storage and policy processing. The qualitative historical examples in Figure~\ref{fig:design} receive the same rotation, with no super-resolution or image synthesis. Policy inputs are $128\times128$; DP evaluation first renders $256\times256$. Main datasets use EGL, whereas the geometry-coupling datasets and their C1 anchors, and E1, use OSMesa. Those backend differences are another reason not to subtract train- and test-time offset effects.

\subsection{Geometry directions and model encoding}
Table~\ref{tab:directions} records $d$ and $d'$ recovered from the frozen configuration. All angles are world-coordinate directions. For each task, $d'$ is selected from the two orthogonal candidates by the larger minimum clearance after a 10 cm shift.
\begin{table}[htbp]
\centering
\small
\caption{Fixed task-wise geometry directions in world coordinates (degrees).}\label{tab:directions}
\setlength{\tabcolsep}{4pt}
\begin{tabular}{lrrr}
\toprule
Task & d & d prime & Magnitude (cm) \\
\midrule
10\_5 & 180.0 & 270.0 & 10 \\
10\_9 & 135.0 & 45.0 & 10 \\
goal\_1 & 270.0 & 180.0 & 10 \\
goal\_6 & 45.0 & 135.0 & 10 \\
goal\_7 & 225.0 & 135.0 & 10 \\
spatial\_2 & 45.0 & 135.0 & 10 \\
spatial\_8 & 225.0 & 315.0 & 10 \\
\bottomrule
\end{tabular}
\end{table}

The geometry logit model uses two columns, \texttt{coupled} and $x$:
\[
\mathrm{C1}=(0,0),\qquad \mathrm{Coupled}=(1,0),\qquad \mathrm{Anti}=(0,1).
\]
It has formula \texttt{success \textasciitilde\ coupled + x + (1|task) + (1|task:seed)}. The $c$ contrast compares $(1,0)$ with $(0,1)$, and $L$ compares $(0,0)$ with $(1,0)$ after logistic-normal marginalization. Both fits use \texttt{adapt\_delta}=0.99 and have zero divergences. The complete direction and design-matrix specification fixes both the rendering intervention and the posterior probability contrasts.

\paragraph{Resource and measurement scope.}
Recorded median training times were 85.8 minutes for DP ($n=364$) and 123.4 minutes for three-camera DP ($n=12$), measured on the earlier eight-GPU machine. These are not an estimate of total study compute. Feature analysis covers 49 seed-1 checkpoints and 196 rows (two perturbation types by two cameras), with four states per task. No cross-seed feature uncertainty is estimated.
\FloatBarrier

\section{Complete Results and Supplementary Analyses}\label{app:suppresults}
\subsection{Primary contrasts and descriptive rates}
Tables~\ref{tab:main_effects} and~\ref{tab:geometry} give the numerical main effects; Table~\ref{tab:rates} reports descriptive success rates, including the unaugmented baseline. Model effects, common-seed paired differences, and raw task-equal rates are distinct and should not be interchanged.
\begin{table}[htbp]
\centering
\small
\caption{DP appearance contrasts. Positive values favor the reference; effects are in percentage points (pp).}\label{tab:main_effects}
\setlength{\tabcolsep}{4pt}
\begin{tabular}{lr}
\toprule
Contrast & Estimate [90\% CrI] \\
\midrule
Global appearance: linear trend, rho 1 to 0 & -0.64 [-3.43, 2.09] \\
Ghost: coupled minus single-view injection & 0.00 [-2.53, 2.55] \\
Target color: shared minus independent & -1.85 [-4.63, 0.69] \\
Additional photometric jitter: C1 minus C8 & -1.90 [-4.95, 0.95] \\
\bottomrule
\end{tabular}
\par\vspace{3pt}\begin{minipage}{\linewidth}\footnotesize The final row measures added image-space jitter; the preceding rows measure changes in cross-view coupling.\end{minipage}
\end{table}

\begin{table}[htbp]
\centering
\small
\caption{Geometry effects with marginal-matched direction distributions (pp).}\label{tab:geometry}
\setlength{\tabcolsep}{4pt}
\begin{tabular}{lrr}
\toprule
Estimand & Estimate [90\% CrI] & Estimate [95\% CrI] \\
\midrule
c: Coupled - Anti & 3.30 [0.64, 6.28] & 3.30 [0.14, 6.94] \\
L: C1 - Coupled & 4.29 [1.61, 7.37] & 4.29 [1.19, 8.11] \\
\bottomrule
\end{tabular}
\par\vspace{3pt}\begin{minipage}{\linewidth}\footnotesize Both models use 147 policies and 2,940 episodes. The 14 new C1 anchors are excluded. The c contrast is within the new batch, whereas L uses a base-batch reference.\end{minipage}
\end{table}

\begin{table}[htbp]
\centering
\scriptsize
\caption{Descriptive DP success on held-out appearance B (\%).}\label{tab:rates}
\setlength{\tabcolsep}{4pt}
\begin{tabular}{lrrrrrrrr}
\toprule
Condition & 10\_5 & 10\_9 & goal\_1 & goal\_6 & goal\_7 & spatial\_2 & spatial\_8 & Mean \\
\midrule
C0 & 27.0 & 60.0 & 65.0 & 90.0 & 93.0 & 89.0 & 84.0 & 72.6 \\
C1 & 82.9 & 77.1 & 51.4 & 88.6 & 92.1 & 92.9 & 95.0 & 82.9 \\
C3 & 80.0 & 85.0 & 54.0 & 92.0 & 94.0 & 95.0 & 87.0 & 83.9 \\
C5 & 82.1 & 83.6 & 47.1 & 90.7 & 92.1 & 97.1 & 91.4 & 83.5 \\
C6 & 80.7 & 97.1 & 57.9 & 85.0 & 94.3 & 95.7 & 94.3 & 86.4 \\
C7 & 80.7 & 99.3 & 56.4 & 87.1 & 92.9 & 96.4 & 92.1 & 86.4 \\
C8 & 77.0 & 92.0 & 56.0 & 85.0 & 96.0 & 95.0 & 92.0 & 84.7 \\
C9 & 79.3 & 84.3 & 47.1 & 87.9 & 85.0 & 97.1 & 97.1 & 82.6 \\
C10 & 70.7 & 92.1 & 55.7 & 86.4 & 90.0 & 98.6 & 97.1 & 84.4 \\
C11 & 53.0 & 95.0 & 53.0 & 74.0 & 81.0 & 92.0 & 93.0 & 77.3 \\
C12 & 52.0 & 97.0 & 60.0 & 68.0 & 82.0 & 95.0 & 92.0 & 78.0 \\
\bottomrule
\end{tabular}
\par\vspace{3pt}\begin{minipage}{\linewidth}\footnotesize Each task receives equal weight in the mean. Conditions have different numbers of seeds; these differences are not model-marginalized effects.\end{minipage}
\end{table}

\begin{figure}[htbp]
\centering\includegraphics[width=\linewidth]{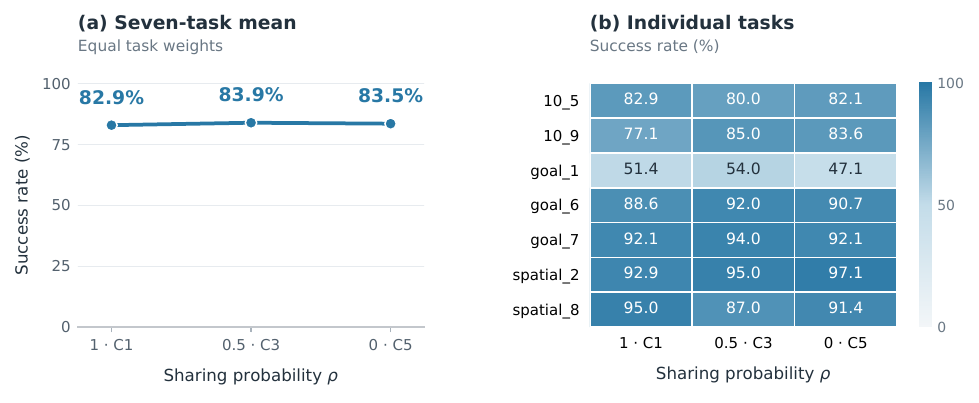}
\caption{\textbf{Global-appearance dose response.} (a) Equal-weight seven-task descriptive average. (b) Individual-task success rates, with cell values in percent and a common 0--100\% color scale. Columns follow C1, C3, and C5 as sharing probability $\rho$ decreases from 1 to 0. All rates use their condition's included seeds; no confidence or credible intervals are inferred.}\label{fig:dose}
\end{figure}
\FloatBarrier

\subsection{Geometry and camera interventions}
Tables~\ref{tab:geo_tasks} and~\ref{tab:anchors} separate the within-new-batch Coupled--Anti comparison from the small batch-anchor sample. The large anchor difference on \texttt{10\_9} warrants caution with the cross-batch $L$ contrast, not a numerical adjustment.
\begin{table}[htbp]
\centering
\small
\caption{Geometry-coupling success by task (\%, seven equally weighted seeds).}\label{tab:geo_tasks}
\setlength{\tabcolsep}{4pt}
\begin{tabular}{lrrrr}
\toprule
Task & Base C1 & Coupled & Anti & Coupled - Anti (pp) \\
\midrule
10\_5 & 82.9 & 45.0 & 40.7 & +4.3 \\
10\_9 & 77.1 & 94.3 & 90.7 & +3.6 \\
goal\_1 & 51.4 & 65.0 & 60.0 & +5.0 \\
goal\_6 & 88.6 & 60.7 & 57.1 & +3.6 \\
goal\_7 & 92.1 & 90.0 & 81.4 & +8.6 \\
spatial\_2 & 92.9 & 93.6 & 95.0 & -1.4 \\
spatial\_8 & 95.0 & 99.3 & 97.9 & +1.4 \\
\bottomrule
\end{tabular}
\par\vspace{3pt}\begin{minipage}{\linewidth}\footnotesize The descriptive mean difference is 3.6 pp; the main model estimate is 3.30 pp.\end{minipage}
\end{table}

\begin{table}[htbp]
\centering
\small
\caption{Retrained minus base-batch C1 success (pp; seeds 1--2 only).}\label{tab:anchors}
\setlength{\tabcolsep}{4pt}
\begin{tabular}{lr}
\toprule
Task & Batch difference \\
\midrule
10\_5 & +5.0 \\
10\_9 & +32.5 \\
goal\_1 & +5.0 \\
goal\_6 & -7.5 \\
goal\_7 & +5.0 \\
spatial\_2 & +2.5 \\
spatial\_8 & -2.5 \\
Mean & +5.7 \\
\bottomrule
\end{tabular}
\par\vspace{3pt}\begin{minipage}{\linewidth}\footnotesize Fourteen task--seed pairs, used to characterize batch sensitivity.\end{minipage}
\end{table}

\begin{table}[htbp]
\centering
\small
\caption{DP camera-mask effects on the original scene (pp; 90\% CrI).}\label{tab:mask}
\setlength{\tabcolsep}{4pt}
\begin{tabular}{lrrr}
\toprule
Training & Policies & Mask third-person & Mask wrist \\
\midrule
C0 & 35 & 26.7 [19.2, 33.6] & 54.9 [43.2, 63.4] \\
C1 & 49 & 15.3 [10.7, 19.9] & 56.2 [43.7, 65.7] \\
C5 & 49 & 16.5 [11.2, 21.7] & 56.1 [43.4, 65.7] \\
C9 & 49 & 21.1 [14.8, 27.0] & 59.5 [45.4, 70.4] \\
C10 & 49 & 19.4 [14.0, 24.7] & 57.5 [44.7, 66.9] \\
C11 & 35 & 32.8 [24.9, 39.7] & 42.9 [32.3, 51.5] \\
C12 & 35 & 30.8 [23.5, 37.1] & 45.7 [34.0, 55.1] \\
\bottomrule
\end{tabular}
\par\vspace{3pt}\begin{minipage}{\linewidth}\footnotesize C12 is exploratory. Masking sets raw pixels to zero before preprocessing; the normalized tensor need not be zero.\end{minipage}
\end{table}

\begin{table}[htbp]
\centering
\small
\caption{Test-only wrist offset: descriptive success and unrounded differences.}\label{tab:test_offset}
\setlength{\tabcolsep}{4pt}
\begin{tabular}{lrrr}
\toprule
Task & 0 cm success (\%) & 10 cm success (\%) & Loss (pp) \\
\midrule
10\_5 & 82.1 & 2.1 & 80.0 \\
10\_9 & 95.7 & 77.1 & 18.6 \\
goal\_1 & 57.1 & 20.7 & 36.4 \\
goal\_6 & 90.0 & 58.6 & 31.4 \\
goal\_7 & 94.3 & 33.6 & 60.7 \\
spatial\_2 & 96.4 & 80.0 & 16.4 \\
spatial\_8 & 92.1 & 95.0 & -2.9 \\
\bottomrule
\end{tabular}
\par\vspace{3pt}\begin{minipage}{\linewidth}\footnotesize Original scene and OSMesa; paired initial states. The descriptive task-equal mean is 34.4 pp; the model estimate is 33.3 pp.\end{minipage}
\end{table}

\FloatBarrier

E1 has a descriptive task-equal loss of 34.4 pp, versus the model-marginalized 33.3 pp. Its task-level sign-flip test gives $p=0.031$. On \texttt{spatial\_8}, the descriptive change reverses sign; Table~\ref{tab:test_offset} retains that observation. Training-time and test-time offsets use different evaluation scenes and rendering settings, and cannot be subtracted.

Offset-Both (C12) was initially restricted to descriptive analysis. Its original task-cluster 90\% intervals for camera masking were 32.9 pp [10.2, 55.5] for third-person and 50.4 pp [26.0, 74.9] for wrist; the available narrative record does not fully identify that interval algorithm. Later brms masking estimates are \effone{camera_mask/dp_C12_agentview_brms} and \effone{camera_mask/dp_C12_wrist_brms} pp, respectively, for the same 35 policies. They are exploratory reanalyses, not additional behavior evaluations. The directly reproducible C12--C11 and C1--C12 common-seed contrasts appear in Table~\ref{tab:offset_exploratory}; neither isolates coupling because the interventions also change which cameras are offset.
\begin{table}[htbp]
\centering
\footnotesize
\caption{Exploratory paired descriptive contrasts (pp; task-cluster t intervals, 6 degrees of freedom).}\label{tab:offset_exploratory}
\setlength{\tabcolsep}{4pt}
\begin{tabular}{lrrr}
\toprule
Contrast & Pairs & Estimate [90\% CI] & Estimate [95\% CI] \\
\midrule
C12 - C11 & 35 & 0.7 [-2.2, 3.7] & 0.7 [-3.0, 4.4] \\
C1 - C12 & 35 & 4.3 [-8.2, 16.8] & 4.3 [-11.5, 20.0] \\
\bottomrule
\end{tabular}
\par\vspace{3pt}\begin{minipage}{\linewidth}\footnotesize Only common seeds enter each comparison. These intervals do not use the unified MCMC procedure.\end{minipage}
\end{table}

Historical paired analyses after Offset-Wrist training reported a 19.7 pp [4.7, 34.7] increase in the third-person masking loss and a 12.4 pp [3.8, 21.0] decrease in the wrist masking loss. Those intervals are not from the unified MCMC procedure. They describe sensitivity to missing inputs and do not directly identify an internal reassignment of view dependence.
\FloatBarrier

\subsection{Co-visibility and feature ratios}
There are 5,104 measurement rows, not 5,104 independent frames. Global-surface measurements have 684 records per element. Ghost measurements have 356 A-to-W and 328 W-to-A records. Target-color measurements pool 12 element-by-direction groups, corresponding to 842 target-element/frame observations per direction. The pooled direct-visibility ratio is 89.9\%; projected co-visibility is 85.0--88.7\%, weighted by the corresponding frame counts. Table projection exceeds 98.2\% for every task. Direct target visibility is 93.1\%, 83.1\%, 90.1\%, and 92.1\% for approach, grasp, transport, and post-placement stages, respectively. These measures concern content exposure, not whether a policy compares the exposed regions.

Ghost projected co-visibility ranges from 0 to 36.1\% across task/direction groups. In \texttt{10\_5}, the source-view visible-frame proportion is zero for reasons not resolved by the current record. This uncertainty reinforces the low-dose qualification and is not repaired by relabeling injection eligibility as visibility.
\begin{table}[htbp]
\centering
\footnotesize
\caption{Element- and direction-specific co-visibility measurements (\%).}\label{tab:covisibility}
\setlength{\tabcolsep}{4pt}
\begin{tabular}{lrrrrr}
\toprule
Element & Direction & Records & Source visible & Both direct & Projected \\
\midrule
ghost & A->W & 356 & 69.9 & - & 12.1 \\
ghost & W->A & 328 & 16.5 & - & 13.4 \\
all\_surfaces & W->A & 684 & 100.0 & 100.0 & 99.7 \\
floor & W->A & 684 & 16.8 & 9.1 & 0.3 \\
table & W->A & 684 & 100.0 & 100.0 & 99.6 \\
wall & W->A & 684 & 9.5 & 9.5 & 4.8 \\
\bottomrule
\end{tabular}
\par\vspace{3pt}\begin{minipage}{\linewidth}\footnotesize A: third-person; W: wrist. Denominators are the displayed measurement records, not all unique frames. Target-object pooled results and their separate denominator are described in the text.\end{minipage}
\end{table}

\begin{table}[htbp]
\centering
\small
\caption{Encoder invariance ratios, averaged over cameras and seven seed-1 policies.}\label{tab:features}
\setlength{\tabcolsep}{4pt}
\begin{tabular}{lrrr}
\toprule
Policy & Training & Global appearance & Target color \\
\midrule
DP & C0 & 0.216 & 0.213 \\
DP & C1 & 0.036 & 0.157 \\
DP & C5 & 0.042 & 0.160 \\
DP & C9 & 0.037 & 0.048 \\
DP & C10 & 0.036 & 0.048 \\
ACT & C9 & 0.164 & 0.144 \\
ACT & C10 & 0.165 & 0.141 \\
\bottomrule
\end{tabular}
\par\vspace{3pt}\begin{minipage}{\linewidth}\footnotesize Ratios have no cross-seed uncertainty interval. DP and ACT feature spaces differ, so their absolute ratios are not a controlled comparison of architectures.\end{minipage}
\end{table}

The pooled feature ratio over the augmented DP groups and both perturbation types is 0.071, versus 0.215 for unaugmented DP. We emphasize the factor-specific, within-architecture pattern rather than these pooled values: the ratio depends on two distance terms, and DP keypoints and ACT tokens occupy different spaces. No seed-level uncertainty or causal mediation claim is attached to these descriptive measurements.
\FloatBarrier

\subsection{Test-time photometric coupling}
\begin{figure}[htbp]
\centering\includegraphics[width=\linewidth]{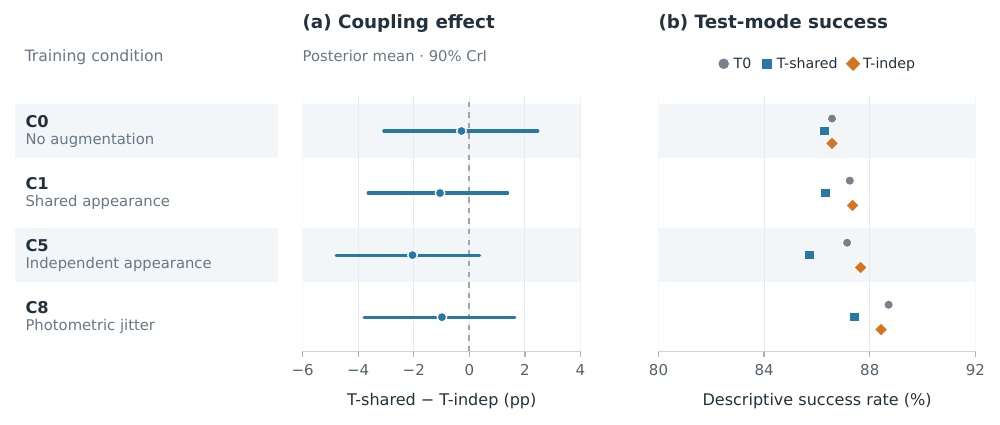}
\caption{\textbf{Matched photometric perturbations at test time.} (a) Posterior mean coupling costs with 90\% CrI. (b) Descriptive success for unperturbed (T0), shared (T-shared), and independent (T-indep) modes, vertically offset within each training-condition row for readability. The success-rate axis spans 80--92\%. Each policy has 20 episodes per mode on the original scene. Training C8 includes stronger independent photometric jitter; C0, C1, and C5 do not.}\label{fig:photo}
\end{figure}

Tables~\ref{tab:photo} and~\ref{tab:photo95} retain success rates and model contrasts separately. Figure~\ref{fig:photo} shows the matched T-shared--T-indep contrast and all three descriptive mode rates. The narrow test range has limited effect even on C0, constraining its ability to distinguish learned appearance invariance from pre-existing insensitivity.
\begin{table}[htbp]
\centering
\small
\caption{Test-time photometric coupling on the original scene.}\label{tab:photo}
\setlength{\tabcolsep}{4pt}
\begin{tabular}{lrrrr}
\toprule
Training & T0 (\%) & Shared (\%) & Indep. (\%) & Cost [90\% CrI] \\
\midrule
C0 & 86.6 & 86.3 & 86.6 & -0.28 [-3.08, 2.46] \\
C1 & 87.2 & 86.3 & 87.3 & -1.05 [-3.64, 1.38] \\
C5 & 87.1 & 85.7 & 87.7 & -2.05 [-4.80, 0.37] \\
C8 & 88.7 & 87.4 & 88.4 & -0.99 [-3.79, 1.64] \\
\bottomrule
\end{tabular}
\par\vspace{3pt}\begin{minipage}{\linewidth}\footnotesize Cost = SR(T-shared) minus SR(T-indep), in pp. There are 168 reused policies and 10,080 episodes across three modes. Each contrast fits only the two perturbed modes.\end{minipage}
\end{table}

\begin{table}[htbp]
\centering
\small
\caption{Additional 95\% posterior intervals for test-time photometric coupling (pp).}\label{tab:photo95}
\setlength{\tabcolsep}{4pt}
\begin{tabular}{lr}
\toprule
Training & Estimate [95\% CrI] \\
\midrule
C0 & -0.28 [-3.67, 3.08] \\
C1 & -1.05 [-4.25, 1.82] \\
C5 & -2.05 [-5.40, 0.85] \\
C8 & -0.99 [-4.47, 2.18] \\
\bottomrule
\end{tabular}
\end{table}

\FloatBarrier

\subsection{Task effects and policy configurations}
Tables~\ref{tab:fixed_gain} and~\ref{tab:fixed_other} report all fixed task-interaction estimates, including negative augmentation gains and the task-specific color cost that admits larger harm. The task-equal means in these models differ slightly from the main random-intercept model. Table~\ref{tab:architecture} gives the separate DP and ACT configuration estimates.
\begin{table}[htbp]
\centering
\scriptsize
\caption{Augmentation gain and global-appearance coupling: fixed task-interaction estimates (pp; 90\% CrI).}\label{tab:fixed_gain}
\setlength{\tabcolsep}{4pt}
\begin{tabular}{lrr}
\toprule
Task & Augmentation gain & Global appearance \\
\midrule
10\_5 & 48.3 [38.7, 57.5] & 0.6 [-6.5, 7.8] \\
10\_9 & 17.3 [7.1, 27.5] & -6.2 [-14.0, 1.3] \\
goal\_1 & -11.4 [-21.9, -0.8] & 4.4 [-5.1, 13.8] \\
goal\_6 & -0.4 [-7.1, 6.8] & -2.2 [-8.0, 3.6] \\
goal\_7 & 0.2 [-5.4, 6.1] & -0.1 [-5.2, 5.0] \\
spatial\_2 & 4.8 [-1.3, 11.7] & -4.2 [-8.8, -0.2] \\
spatial\_8 & 11.1 [4.5, 18.6] & 3.4 [-1.9, 8.8] \\
Task-equal mean & 10.0 [6.8, 13.1] & -0.6 [-3.2, 1.9] \\
\bottomrule
\end{tabular}
\par\vspace{3pt}\begin{minipage}{\linewidth}\footnotesize Cell-wise 90\% CrI. Augmentation gain is C1 minus C0; the other contrasts favor the reference when positive. Photometric jitter is added in C8.\end{minipage}
\end{table}

\begin{table}[htbp]
\centering
\scriptsize
\caption{Other appearance contrasts: fixed task-interaction estimates (pp; 90\% CrI).}\label{tab:fixed_other}
\setlength{\tabcolsep}{4pt}
\begin{tabular}{lrrr}
\toprule
Task & Ghost & Added photo. jitter & Target color \\
\midrule
10\_5 & -0.1 [-7.6, 7.4] & 4.9 [-3.0, 13.2] & 8.0 [-0.3, 16.3] \\
10\_9 & -2.0 [-4.9, 0.5] & -14.4 [-21.6, -7.2] & -7.7 [-14.0, -1.5] \\
goal\_1 & 1.4 [-8.3, 11.0] & -4.0 [-14.5, 6.5] & -8.5 [-17.9, 1.2] \\
goal\_6 & -2.2 [-9.1, 4.7] & 3.5 [-3.6, 10.9] & 1.5 [-5.0, 8.1] \\
goal\_7 & 1.4 [-3.3, 6.3] & -3.7 [-8.5, 1.2] & -4.9 [-11.6, 1.6] \\
spatial\_2 & -0.7 [-4.6, 3.1] & -2.1 [-7.0, 3.2] & -1.3 [-4.4, 1.5] \\
spatial\_8 & 2.2 [-2.7, 7.4] & 2.8 [-2.1, 8.3] & 0.1 [-3.2, 3.4] \\
Task-equal mean & 0.0 [-2.4, 2.4] & -1.8 [-4.6, 0.9] & -1.8 [-4.3, 0.6] \\
\bottomrule
\end{tabular}
\par\vspace{3pt}\begin{minipage}{\linewidth}\footnotesize Cell-wise 90\% CrI. Augmentation gain is C1 minus C0; the other contrasts favor the reference when positive. Photometric jitter is added in C8.\end{minipage}
\end{table}

\begin{table}[htbp]
\centering
\small
\caption{Positive controls and target-color contrasts across policy configurations (pp; 90\% CrI).}\label{tab:architecture}
\setlength{\tabcolsep}{4pt}
\begin{tabular}{lrr}
\toprule
Configuration & C1 - Offset-Wrist & Shared - independent color \\
\midrule
DP & 5.50 [2.28, 9.03] & -1.85 [-4.63, 0.69] \\
ACT & 8.50 [4.60, 12.93] & 2.19 [-0.52, 5.28] \\
\bottomrule
\end{tabular}
\par\vspace{3pt}\begin{minipage}{\linewidth}\footnotesize The ACT minus DP color-cost interaction is 4.10 [0.36, 8.52] pp. Machine sensitivity is examined in Appendix~\ref{app:transparency}.\end{minipage}
\end{table}

\FloatBarrier

\paragraph{The high-overlap three-camera experiment did not pass its prerequisite.}
A third camera was added 45 degrees from the original third-person camera. The target is simultaneously visible in the two third-person views in 99.37--100\% of frames. Recoloring conflicts are applied to that pair; the wrist reuses the original third-person color. However, masking the added view costs only 2.3 pp, below the pre-specified 5 pp view-dependence threshold. H7a's historical estimate, 1.4 pp [-1.1, 3.9], therefore lacks its planned interpretive prerequisite; H7b's 3.3 pp [1.6, 5.2] interaction is subject to the same limitation. These results do not establish harmlessness of high-overlap inconsistency. Content-level co-visibility measurements answer a different question and do not repair the failed dependence check.

\paragraph{Image metrics did not pass the calibration criterion.}
Seven image measures were calibrated against five inconsistency types: Lab color difference, RGB MAE, luminance difference, ORB match and unmatched counts, color difference at matched points, and median Sampson error. None of 35 cells reached both AUC $\geq0.75$ and effect size $\geq0.5$. The closest AUC was 0.712 for luminance difference under independent appearance. At $128\times128$, pairs usually contained only one or two good ORB matches, making geometric metrics unstable. High surface visibility does not imply abundant low-resolution keypoint matches; these results do not establish that the metrics are generally invalid.

\paragraph{Held-out target colors.}
The exploratory F-set evaluation recolors the target asset group with held-out colors. Table~\ref{tab:color_generalization} compares 49 common task--seed pairs using task-cluster t intervals (six degrees of freedom), not MCMC. C9--C1 is -2.0 pp [-3.4, -0.7], whereas C10--C1 is +0.5 pp [-2.6, 3.7]. These results do not establish that color augmentation necessarily improves color generalization or that independent color augmentation outperforms shared appearance.
\begin{table}[htbp]
\centering
\footnotesize
\caption{Exploratory paired descriptive contrasts (pp; task-cluster t intervals, 6 degrees of freedom).}\label{tab:color_generalization}
\setlength{\tabcolsep}{4pt}
\begin{tabular}{lrrr}
\toprule
Contrast & Pairs & Estimate [90\% CI] & Estimate [95\% CI] \\
\midrule
C9 - C1 & 49 & -2.0 [-3.4, -0.7] & -2.0 [-3.7, -0.4] \\
C10 - C1 & 49 & 0.5 [-2.6, 3.7] & 0.5 [-3.4, 4.5] \\
C9 - C10 & 49 & -2.6 [-5.7, 0.6] & -2.6 [-6.6, 1.5] \\
\bottomrule
\end{tabular}
\par\vspace{3pt}\begin{minipage}{\linewidth}\footnotesize Only common seeds enter each comparison. These intervals do not use the unified MCMC procedure.\end{minipage}
\end{table}

\FloatBarrier

\section{Statistical Sensitivity and Historical Tests}\label{app:sensitivity}
\subsection{Historical planned comparisons}
Table~\ref{tab:pretests} preserves the earlier tests, their original confidence-interval definitions, and p-values. It is not a table of p-values attached to the main Bayesian estimates. Identifiers map as follows: H0a/H0b/H0c are Offset-Wrist positive controls for DP/three-camera DP/ACT; H1 is global-appearance dose; H2 ghost coupling; H3 the C1--C8 additional-photometric contrast; H4 the independent-minus-shared wrist-masking loss; H5 DP target-color coupling; H6b ACT target-color coupling; H6c the ACT--DP color interaction; H7a the three-camera color contrast; and H7b its interaction with the two-camera DP contrast.

The historical procedure used separate Holm correction families for H1--H5 and H6b/H6c. Each positive control used $\alpha=0.05$. The historical ``drop below 8 pp'' decision required a nonsignificant corrected harm test, a 90\% interval wholly inside $(-8,+8)$ pp, and a passed corresponding positive control. Under that procedure H1, H2, H5, and H6b received the below-bound decision, H3 passed non-inferiority, H4 was not supported, and H6c indicated a difference between configurations. H0a's positive-control conclusion was not robust to the random-slope model. H7a failed its dependence prerequisite; H7b was reported only as an estimate under the same limitation. These decisions retain their historical meaning and do not turn the current MCMC intervals into equivalence tests. H6b/H6c's unrounded historical 90\% intervals were [-0.546, 5.289] and [2.149, 6.629] pp.
\begin{table}[htbp]
\centering
\footnotesize
\caption{Historical planned tests (pp).}\label{tab:pretests}
\setlength{\tabcolsep}{4pt}
\begin{tabular}{lrrrr}
\toprule
ID & Estimate & 90\% CI & 95\% CI & p \\
\midrule
H0a & 5.6 & [3.2, 8.2] & [2.8, 8.7] & 4.6e-05 \\
H0b & 9.3 & [6.9, 11.9] & [6.4, 12.4] & 6.6e-13 \\
H0c & 9.0 & [6.5, 11.6] & [6.1, 12.2] & 2.1e-11 \\
H1 & -0.7 & [-2.4, 1.1] & [-2.7, 1.4] & 1 \\
H2 & 0.0 & [-2.5, 2.5] & [-2.9, 2.9] & 1 \\
H3 & -2.0 & [-4.1, 0.3] & [-4.5, 0.7] & 1.1e-13$\dagger$ \\
H4 & -0.8 & [-2.6, 1.0] & [-3.0, 1.3] & 1 \\
H5 & -1.9 & [-4.5, 0.7] & [-5.0, 1.1] & 1 \\
H6b & 2.4 & [-0.5, 5.3] & [-1.1, 5.8] & 0.094 \\
H6c & 4.3 & [2.1, 6.6] & [1.8, 7.1] & 0.00065 \\
H7a & 1.4 & [-1.1, 3.9] & [-1.5, 4.3] & 0.18 \\
H7b & 3.3 & [1.6, 5.2] & [1.3, 5.5] & — \\
\bottomrule
\end{tabular}
\par\vspace{3pt}\begin{minipage}{\linewidth}\footnotesize H1--H5 and H6b/H6c use separate Holm families. H3 $\dagger$ denotes its non-inferiority p-value, not an exploratory experiment. Historical decisions are described in the text.\end{minipage}
\end{table}

\FloatBarrier

\subsection{Random slopes and permutation tests}
The historical random-slope analysis adds condition-by-task random slopes and separates a population-average contrast from prediction for a new task (Table~\ref{tab:random_slopes}). It is independent of the unified brms refits. The maximum average-effect upper endpoint among the four historical appearance analyses is 4.3 pp, but photometric and color new-task intervals are considerably wider. H0a's random-slope average interval includes zero. These findings illustrate why average effects should not be generalized into per-task guarantees.
\begin{table}[htbp]
\centering
\small
\caption{Historical random-slope sensitivity, separate from the unified brms fits.}\label{tab:random_slopes}
\setlength{\tabcolsep}{4pt}
\begin{tabular}{lrr}
\toprule
ID & Population mean [90\% interval] & New-task prediction [90\% interval] \\
\midrule
H0a & 6.4 [-2.9, 17.8] & 6.4 [-17.9, 41.1] \\
H1 & -0.5 [-4.1, 3.2] & -0.5 [-7.3, 7.1] \\
H2 & 0.1 [-2.9, 3.2] & 0.1 [-5.1, 5.8] \\
H3 & -1.4 [-6.6, 4.3] & -1.3 [-14.2, 12.1] \\
H5 & -1.7 [-5.9, 2.6] & -1.8 [-12.0, 7.5] \\
H6b & 2.3 [-1.1, 6.1] & 2.2 [-3.0, 10.0] \\
\bottomrule
\end{tabular}
\par\vspace{3pt}\begin{minipage}{\linewidth}\footnotesize Intervals retain the historical analysis definition. Settings were recorded as M=2,000, K=500, with 12,000 posterior draws; this is not the main random-intercept analysis.\end{minipage}
\end{table}

\begin{table}[htbp]
\centering
\small
\caption{Exact task-level sign-flip tests on common-seed paired differences.}\label{tab:permutation}
\setlength{\tabcolsep}{4pt}
\begin{tabular}{lrr}
\toprule
ID & Mean difference (pp) & Two-sided p \\
\midrule
H0a & 5.0 & 0.438 \\
H0b & 8.4 & 0.0625 \\
H0c & 8.0 & 0.0156 \\
H1 & -0.6 & 0.75 \\
H2 & 0.0 & 1 \\
H3 & -2.4 & 0.5 \\
H5 & -1.8 & 0.531 \\
\bottomrule
\end{tabular}
\par\vspace{3pt}\begin{minipage}{\linewidth}\footnotesize All 128 task-level sign patterns are enumerated. These are sensitivity analyses, not replacements for the main estimates.\end{minipage}
\end{table}

\FloatBarrier

\subsection{Leave-one-task-out fits and estimator sensitivity}
\begin{figure}[htbp]
\centering\includegraphics[width=\linewidth]{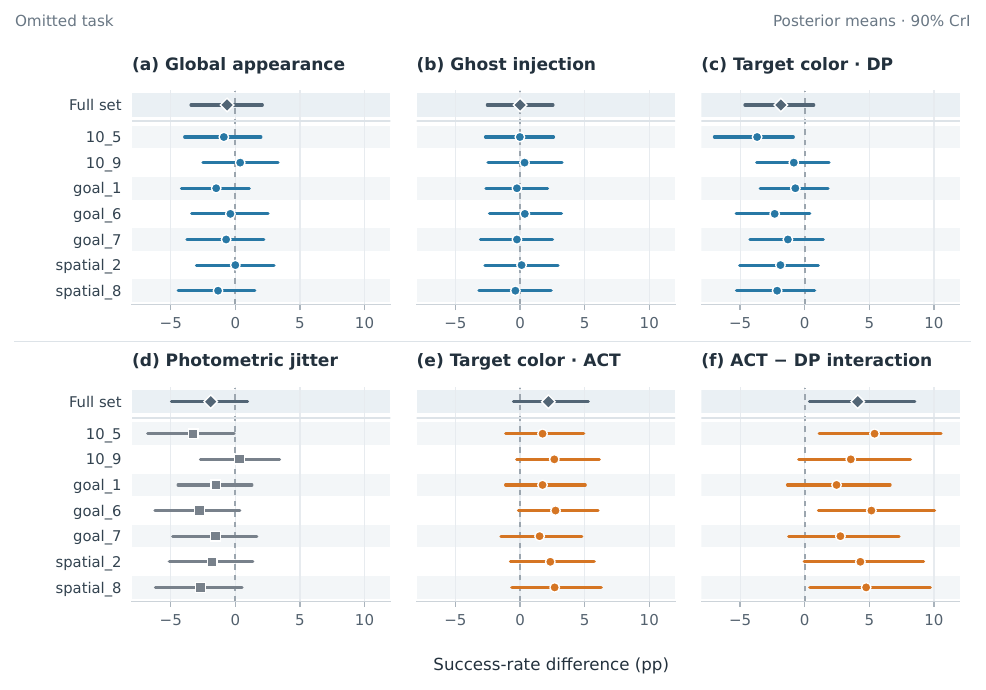}
\caption{\textbf{Leave-one-task-out sensitivity.} Each task row omits the named task. Points and bars are posterior means and 90\% CrI on a common horizontal scale. The shaded \emph{Full set} row shows the full-task posterior mean (diamond) and 90\% CrI; dashed lines mark zero. These are sample-composition checks, not independent replications. The photometric contrast remains an added-perturbation control.}\label{fig:loto}
\end{figure}

Table~\ref{tab:loto} and Figure~\ref{fig:loto} summarize the full leave-one-task-out analysis. The DP positive-control interval includes zero when \texttt{10\_5} is omitted: 1.58 pp [-1.65, 5.05]. H6c includes zero under four omissions. Removing \texttt{10\_9} gives H1/H2/H3 point estimates of +0.37/+0.35/+0.32 pp, weakening any interpretation of a general decorrelation benefit. Omitting \texttt{10\_5} yields negative H3/H5 intervals of -3.26 pp [-6.79, -0.10] and -3.69 pp [-6.99, -0.89], but these subset analyses do not establish a broadly beneficial independent-augmentation effect.

Nine initial fits had one to three divergences with \texttt{adapt\_delta}=0.95. Refitting used 0.99, or 0.995 for the H4 omission of \texttt{spatial\_8}; all final fits have zero divergences and point-estimate changes at most 0.06 pp. Final output diagnostics and model inputs are available, but recovery records for eight of those reruns are incomplete. The missing historical recovery records are not evidence of remaining divergences.
\begin{table}[htbp]
\centering
\footnotesize
\caption{Leave-one-task-out sensitivity (brms; pp).}\label{tab:loto}
\setlength{\tabcolsep}{4pt}
\begin{tabular}{lrrrr}
\toprule
ID & Estimate range & Max. 90\% upper & Omitted task & Covers zero \\
\midrule
H0a & 1.58 to 9.29 & 13.83 & 10\_9 & 1/7 \\
H0b & 6.79 to 11.48 & 16.59 & goal\_1 & 0/7 \\
H0c & 5.59 to 10.29 & 15.78 & goal\_1 & 0/7 \\
H1 & -1.48 to 0.37 & 3.30 & 10\_9 & 7/7 \\
H2 & -0.36 to 0.37 & 3.25 & 10\_9 & 7/7 \\
H3 & -3.26 to 0.32 & 3.42 & 10\_9 & 6/7 \\
H4 & -3.09 to 0.84 & 4.76 & 10\_9 & 7/7 \\
H5 & -3.69 to -0.73 & 1.89 & 10\_9 & 6/7 \\
H6b & 1.51 to 2.75 & 6.28 & spatial\_8 & 7/7 \\
H6c & 2.46 to 5.40 & 10.55 & 10\_5 & 4/7 \\
\bottomrule
\end{tabular}
\end{table}

Table~\ref{tab:fitters} compares historical variational Bayes (VB), glmer with parametric bootstrap, and full brms MCMC. Seven of ten brms intervals are at least 20\% wider than their earlier VB counterparts. The H3 glmer fit is singular. Each column retains its original interval definition; the main paper uses final brms effects.
\begin{table}[htbp]
\centering
\scriptsize
\caption{Fitter sensitivity. Each interval retains the corresponding method's interpretation.}\label{tab:fitters}
\setlength{\tabcolsep}{4pt}
\begin{tabular}{lrrr}
\toprule
ID & Historical VB [90\%] & glmer [90\% CI] & brms [90\% CrI] \\
\midrule
H0a & 5.63 [3.24, 8.17] & 5.64 [2.27, 9.25] & 5.50 [2.28, 9.03] \\
H0b & 9.31 [6.88, 11.92] & 9.19 [5.17, 13.23] & 9.07 [5.29, 13.36] \\
H0c & 9.00 [6.53, 11.63] & 8.88 [4.42, 13.16] & 8.50 [4.60, 12.93] \\
H1 & -0.70 [-2.38, 1.08] & -0.64 [-3.17, 1.98] & -0.64 [-3.43, 2.09] \\
H2 & 0.01 [-2.46, 2.45] & 0.00 [-2.46, 2.35] & 0.00 [-2.53, 2.55] \\
H3 & -1.99 [-4.08, 0.25] & -1.90 [-4.97, 0.99] & -1.90 [-4.95, 0.95] \\
H4 & -0.78 [-2.63, 1.00] & -1.23 [-5.04, 2.37] & -1.00 [-4.63, 2.65] \\
H5 & -1.90 [-4.50, 0.67] & -1.85 [-4.44, 0.80] & -1.85 [-4.63, 0.69] \\
H6b & 2.37 [-0.55, 5.29] & 2.27 [-0.60, 5.39] & 2.19 [-0.52, 5.28] \\
H6c & 4.32 [2.15, 6.63] & 4.18 [0.17, 8.78] & 4.10 [0.36, 8.52] \\
\bottomrule
\end{tabular}
\par\vspace{3pt}\begin{minipage}{\linewidth}\footnotesize H3 is singular under glmer. The three columns retain their original uncertainty definitions.\end{minipage}
\end{table}

\FloatBarrier

\section{Analysis Changes, Machine Sensitivity, and Reproducibility}\label{app:transparency}
\subsection{Pre-specified and post-observation decisions}
The geometry-coupling and test-time photometric supplement was specified at 14:42 China Standard Time on November 4, 2026; formal geometry training started at 13:16 on November 5. The geometry rule supported a coupling cost if the 90\% interval for $c$ lay above zero. A stronger statement that the cost was mainly attributable to geometric error additionally required evidence for $L>0$ and a 90\% upper endpoint for $c$ below 2.5 pp. The observed results satisfy the first rule but not the stronger one. The photometric experiment met its directional rule but not its strong-support rule, and its weak effect on C0 limits mechanism identification.

An earlier supplementary plan was recorded on October 23, 2026, explicitly after results had been observed. The public archival registration is incomplete. We therefore distinguish internal time-stamped decisions from externally verifiable prospective registration, rather than describing all analyses as prospectively registered. Co-visibility and feature analysis are descriptive; leave-one-task-out and estimator comparisons assess sensitivity.

\begin{table}[htbp]
\centering\footnotesize
\caption{Material changes and departures from earlier plans.}\label{tab:deviations}
\begin{tabularx}{\linewidth}{p{.22\linewidth}X}
\toprule Item & Implementation and interpretation \\\midrule
Task/data scale & Ten planned tasks became seven; demonstrations per task changed from 20 to 15. Generalization and data-scale claims are limited accordingly.\\
Policy inventory & About 1,080 planned base policies became 784. The later geometry experiment adds 112, yielding 896.\\
Wrist displacement & The positive-control displacement increased from 5 to 10 cm following the pilot rule; conclusions concern the implemented magnitude.\\
Main estimator & A planned glmer analysis was initially implemented with VB; the present main effects use full MCMC. Earlier intervals remain separate.\\
ACT training budget & The budget changed from 50k to 25k steps after a half-budget check met its threshold. Results concern the actual 25k configuration.\\
Augmentation gate & The early common-seed estimate was 9.7 pp, below the planned 10 pp threshold. The original disposition record for that case is incomplete; the later 10.0 pp fixed-task estimate does not retroactively pass the gate.\\
Interaction implementation & The architecture/three-camera contrast vectors were corrected. Historical records state that pp estimates and decisions were unchanged, but a complete code-backed reconstruction of that assertion is missing.\\
Random slopes & Before execution, the random-slope model changed from a possible replacement main analysis to a sensitivity analysis.\\
Task omission & An early script dropping \texttt{10\_5} omitted the shared-appearance condition. It was corrected; the current analysis uses all seven omission fits.\\
ACT machines & Training planned on one machine used two. Machine adjustment cannot remove all confounding.\\
E1 and C12 & E1 uses original scenes, OSMesa, and paired 0/10 cm episodes. C12's later masking models go beyond its original descriptive-only scope and are exploratory.\\
\bottomrule
\end{tabularx}
\end{table}
\FloatBarrier

\subsection{ACT training-machine sensitivity}
All formal ACT C1 policies were trained on the host machine, and all C11 policies on the other machine. Both recoloring conditions span the two machines. Three historical checks address different issues and do not share the main unified brms procedure:
\begin{itemize}
\item Re-evaluating 91 remotely trained policies on the host gives host-minus-original evaluation of -0.6 pp (95\% CI [-3.4, 2.1]); across all 133 policies the difference is -0.5 pp [-2.3, 1.3]. An earlier roughly 5 pp evaluation discrepancy did not recur.
\item Training shared-appearance ACT on both machines for seeds 2--3 gives 14 paired policies and a remote-minus-host training effect of +1.1 pp (95\% CI [-7.7, 9.8]) under common host evaluation. This is a small point estimate with wide uncertainty.
\item An exploratory color model adds a training-machine covariate to the independent-color indicator and random task/task--seed intercepts, using 1,400 episodes from 70 policies. Standardizing to the empirical remote-machine proportion of 0.8 gives a same-minus-independent color effect of 4.2 pp (95\% CI [1.6, 7.0]; 90\% CI [2.0, 6.5]). The corresponding marginal machine effect is +3.1 pp (95\% CI [1.0, 5.2]).
\end{itemize}
Both adjusted color upper endpoints are below the historical 8 pp threshold, but the sample and estimator differ from the main comparison. The ACT positive control cannot jointly identify a machine covariate because C1 and C11 are completely separated by training machine. Positive machine-effect point estimates would favor C11 and could attenuate its apparent cost, but the wide direct-check interval does not establish that this confounding is resolved.

\subsection{What can and cannot be reproduced}
The refactored analysis package preserves 93,660 evaluation rows and 896 policy metadata records, saved paper-run model outputs, and code for preparing the inputs. The manuscript's figure/table export reconstructs all 106 model inputs and verifies their hashes against the paper run; it does not rerun MCMC. Archived analysis records report that 106 brms and ten glmer outputs reproduced the paper-run values under R 4.1.2, brms 2.16.3, rstan 2.21.3, StanHeaders 2.21.0-7, lme4 1.1-28, posterior 1.2.0, and g++ 11.4. This is a recorded analysis-reproduction result, not a claim that every trained policy or rendered video can be regenerated bit-for-bit.

Saved co-visibility and feature summaries are the paper-run measurements. Archived verification checks on one task each found differences of up to 19 pixels and 0.014 in camera-averaged per-policy feature ratios relative to historical measurements; unchanged and refactored feature code agreed with one another on rerun, but the historical discrepancy's cause remains unresolved. Backend/driver sensitivity is therefore documented rather than hidden by replacing the saved measurements.

The archived analysis package does not include the large rendered training datasets or 896 trained checkpoints, nor code for every superseded VB, random-slope, machine-sensitivity, or image-metric analysis. Those historical analyses remain transparently separated. Four incomplete seed-7 geometry units were archived and regenerated. Recovery records for eight initial-divergence reruns remain incomplete, although final fit diagnostics pass. Task-wise geometry directions and model coding are specified in Appendix~\ref{app:design}.

High-co-visibility ghost testing was triggered by the planned low-dose criterion but was not conducted. Third-person-only offsets and ACT geometry coupling were also not run. These unperformed experiments are limitations, not implicit supporting results. The available evidence thus distinguishes reproducible final analyses from incomplete historical records and experiments left for future work.
\FloatBarrier

\section{Generator Pilot and Intervention Coverage}\label{app:pilot}
The pilot used five LIBERO task segments (demonstration 0, the first 93 frames) and two prompt/seed schemes for Cosmos-Transfer2.5~\cite{cosmos}. Scheme a sampled them independently for the two views; scheme b shared them. The resulting ten paired-view units were generated at 720p and 16 fps and annotated manually. A category is counted when its severity label is at least one; several categories can occur in the same unit.

\begin{figure}[htbp]
\centering\includegraphics[width=\linewidth]{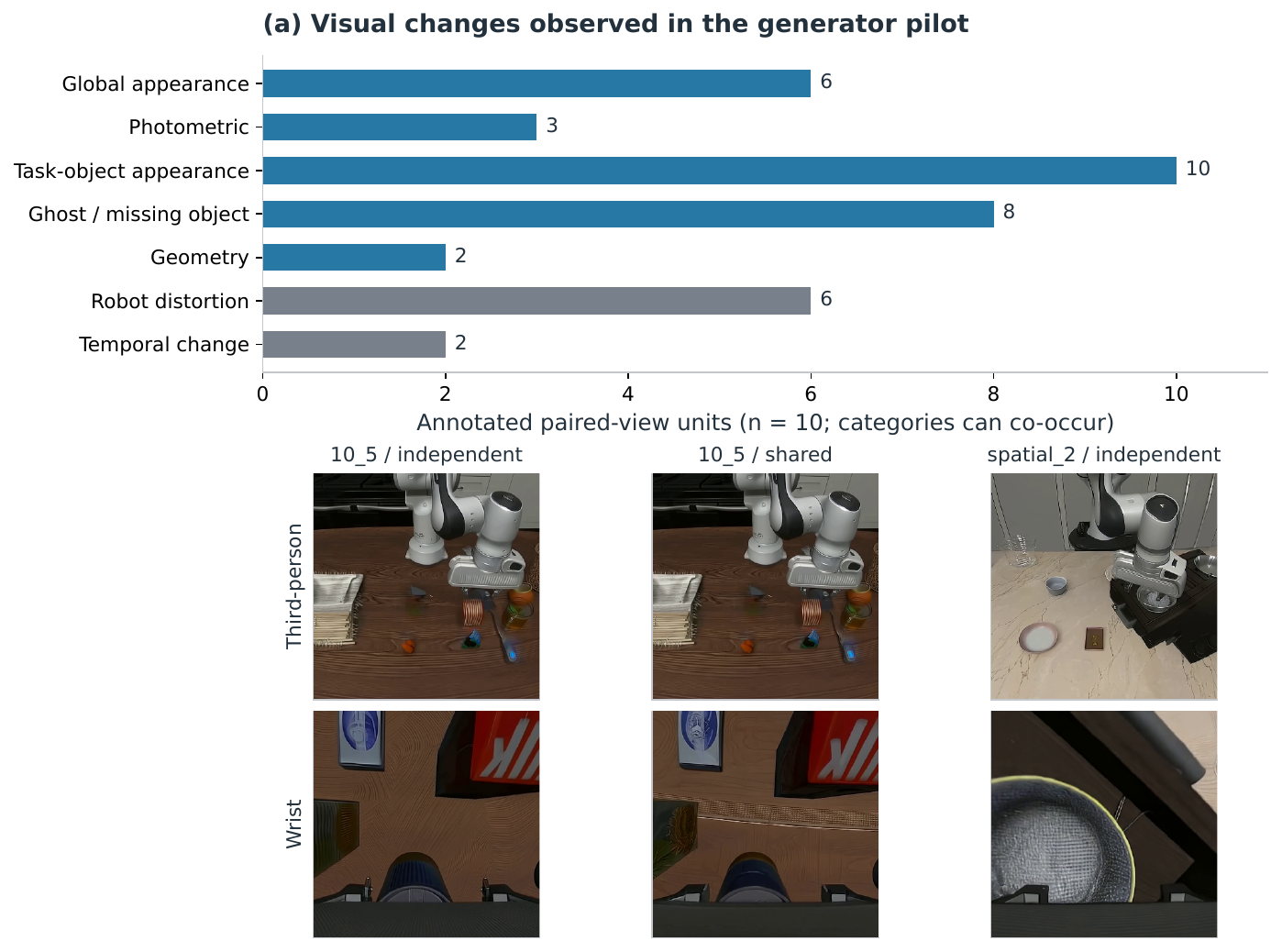}
\caption{\textbf{Cosmos-Transfer2.5 pilot observations.} Top: counts of annotated categories among ten paired-view units, comprising five tasks under two prompt/seed schemes. Bottom: third-person and wrist frames at frame 46 of demonstration 0 for two tasks. The labels identify independent or shared prompts and seeds. Images are taken from the saved generator outputs. The gray bars correspond to robot distortion and temporal changes, which lie outside the static injection experiments.}\label{fig:pilot}
\end{figure}

Task-object appearance differences occurred in all ten units, including all five with shared prompts and seeds. Ghosts or missing objects occurred in eight, global-appearance differences in six, photometric differences in three, and geometry changes in two. Robot distortion occurred in six and within-view temporal change in two (Table~\ref{tab:pilot}). Figure~\ref{fig:pilot} pairs the counts with representative generated frames. These observations guided the choice of controllable visual factors.

The controlled interventions separate these visual factors: recoloring preserves object mesh and identity, ghost injection adds a visual asset, and target offset shifts rendered position while preserving physical state. Both \texttt{10\_5} units also contain identity remapping, which is not separately counted; its overlap with geometry labels remains unresolved. Identity remapping, robot distortion, and temporal discontinuities remain directions for future experiments. The pilot describes selected clips rather than generator-wide failure rates.

\begin{table}[htbp]
\centering
\footnotesize
\caption{Generator pilot: counts out of ten paired-view units.}\label{tab:pilot}
\setlength{\tabcolsep}{4pt}
\begin{tabular}{lrl}
\toprule
Failure label & Units & Controlled counterpart \\
\midrule
Task-object appearance & 10 & Recoloring \\
Ghost / missing object & 8 & Ghost injection \\
Global appearance & 6 & Appearance recipes \\
Photometric & 3 & White balance / brightness \\
Geometry & 2 & Visual target offset \\
Robot distortion & 6 & Not covered \\
Temporal change & 2 & Not covered \\
\bottomrule
\end{tabular}
\par\vspace{3pt}\begin{minipage}{\linewidth}\footnotesize Labels can co-occur; counts cannot be summed into a failure rate.\end{minipage}
\end{table}

\FloatBarrier

\end{document}